%% file: iclr2026/iclr2026_conference.tex
\documentclass{article} 
\usepackage{iclr2026_conference,times}

\input{math_commands.tex}

\usepackage{hyperref}
\usepackage{url}

\usepackage[utf8]{inputenc} 
\usepackage[T1]{fontenc}    
\usepackage{hyperref}       
\usepackage{url}            
\usepackage{booktabs}       
\usepackage{amsfonts}       
\usepackage{nicefrac}       
\usepackage{microtype}      
\usepackage{xcolor}         

\usepackage{amsmath}
\usepackage{amssymb}
\usepackage{mathtools}
\usepackage{amsthm}
\usepackage{tcolorbox}

\usepackage{listings}
\usepackage{adjustbox}  
\usepackage{longtable}
\usepackage{multirow}
\usepackage{enumitem}
\usepackage[table]{xcolor}  
\definecolor{Gray}{gray}{0.9}  
\usepackage{float}  
\usepackage{adjustbox}
\usepackage[table]{xcolor}  
\usepackage{graphicx}
\usepackage{caption}
\usepackage[capitalize,noabbrev]{cleveref}

\usepackage{microtype}
\usepackage{graphicx}
\usepackage{subfigure}
\usepackage{booktabs} 

\newcommand{\xhdr}[1]{\noindent\textbf{#1.}}

\definecolor{darkgreen}{rgb}{0.0, 0.5, 0.0}

\title{The Limits of Inference Scaling Through\\ Resampling}

\author{Benedikt Stroebl \hspace{0.3cm}Sayash Kapoor\hspace{0.3cm}Arvind Narayanan \\
Princeton University \\
\texttt{\{stroebl,sayashk,arvindn\}@princeton.edu} 
}

\iclrfinalcopy 
\begin{document}

\maketitle

\begin{abstract}
Recent research has generated hope that inference scaling, such as resampling solutions until they pass verifiers like unit tests, could allow weaker models to match stronger ones. Beyond inference, this approach also enables training reasoning models, where data is curated using rejection sampling against a verifier. However, we show that this approach is fundamentally limited when verifiers are \textit{imperfect} and have a non-zero probability of producing false positives. Resampling cannot decrease this probability, so it imposes an upper bound to the accuracy of resampling-based inference scaling, \textit{regardless of compute budget}. Our analysis shows that there is a strong correlation between the model’s single-sample accuracy and its false positive rate on HumanEval and MBPP, whose unit tests have limited coverage. Therefore, no amount of inference scaling of weaker models can enable them to match the single-sample accuracy of a sufficiently strong model. Empirical results show that optimal sampling attempts are often fewer than 10, as the negative utility of false positives outweighs benefits, bending inference scaling curves downward. Finally, false positives may have other undesirable qualities, like poor adherence to coding style conventions.
\begin{figure*}[h!]
\centering
\begin{minipage}[b]{0.53\textwidth}
\includegraphics[width=1.03\columnwidth]{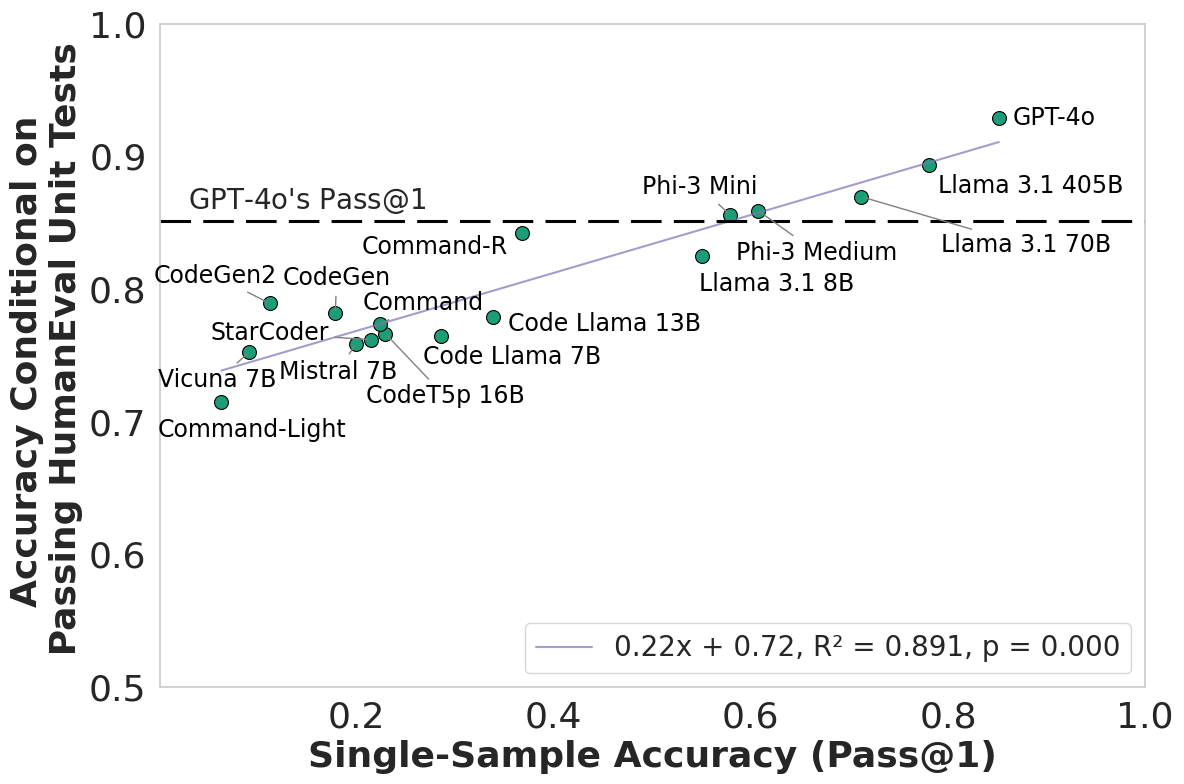}
\caption*{\textbf{Single-sample accuracy vs. resampling limits on HumanEval+.} The x-axis shows single-sample accuracy on HumanEval+ (which contains comprehensive unit tests), while the y-axis shows the highest achievable accuracy when resampling with an infinite compute budget, using HumanEval's more limited unit tests as verifiers. Weaker models (models with lower single-sample accuracy) produce false positive solutions at higher rates. Models below the cutoff line are unable to match GPT-4o through resampling, as GPT-4o's Pass@1 exceeds the accuracy of such a model even when conditioned on its solutions passing the unit tests. Results on MBPP+ follow a similar pattern (\cref{fig:scatterplot_sec2}).}
\label{fig:abstract_scatter}
\end{minipage}
\hfill
\begin{minipage}[b]{0.43\textwidth}
\centering

\includegraphics[width=0.88\columnwidth]{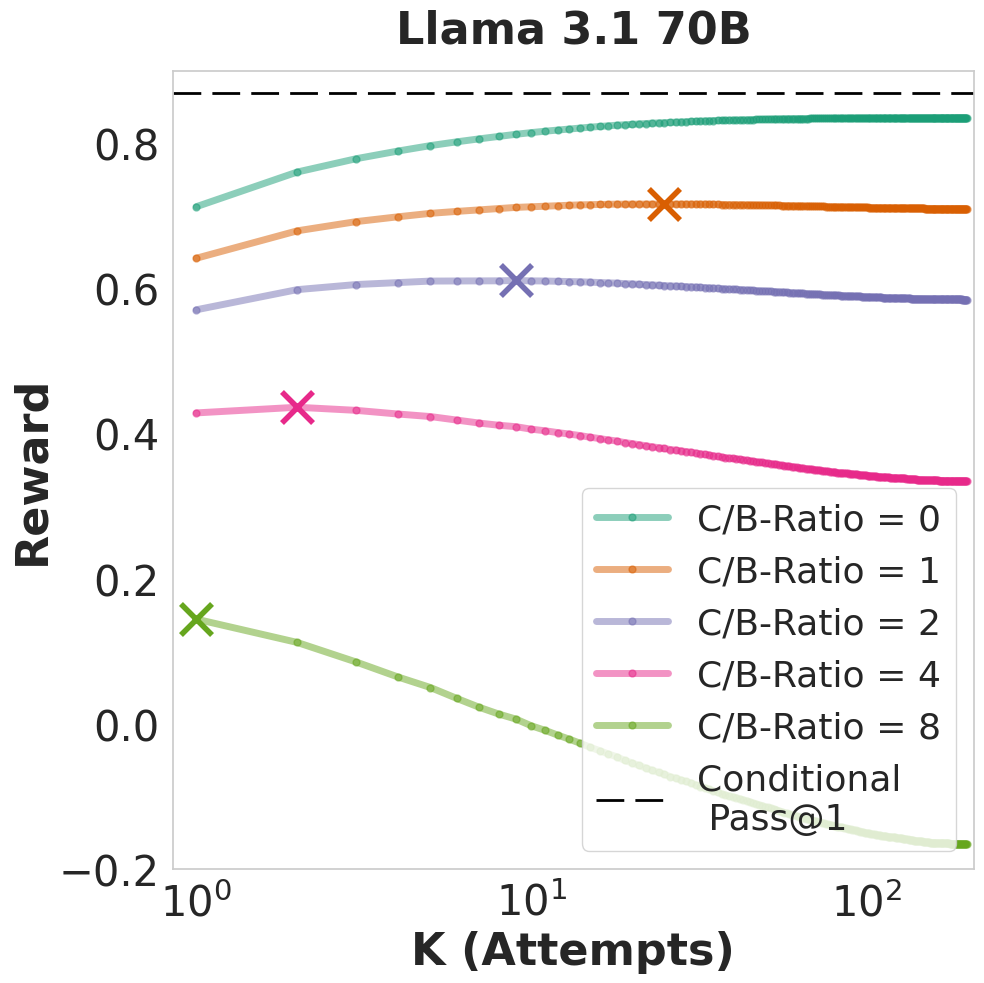}
\captionsetup{justification=raggedright, singlelinecheck=false}
\caption*{\textbf{The cost of false positives limits the reward of resampling.} False positives w.r.t. an imperfect verifier (HumanEval unit tests) incur a "cost" (e.g., subtle bugs in code), while correct answers provide a benefit. The reward (y-axis) depends on this cost-benefit ratio (C/B-Ratio). Problem instances that require more attempts tend to be harder, hence more susceptible to false positives. Thus, even with \textit{zero computational cost}, for realistic cost-benefit ratios, the optimal number of samples $K$ is finite and very low.\vspace{12pt}}
\label{fig:abstract_emp_llama_70b}
\end{minipage}
\caption{Overview of our main findings.\vspace{-0.4cm}}
\label{fig:empirical_scaling_curves}
\end{figure*}
\end{abstract}

\section{Introduction}

Scaling the amount of compute used during inference is a promising way to improve LLM performance. Techniques include reasoning \citep{wei_chain--thought_2023, wu_thinking_2024, setlur_rewarding_2024}, reflecting on model outputs to revise candidate solutions \citep{shinn_reflexion_2023, zhong_debug_2024}, and compositions of these and other atomic techniques \citep{saad-falcon_archon_2024, welleck_decoding_2024}.

Inference scaling through {\em resampling} stands out for its simplicity and broad applicability. It works by generating many candidate outputs until one is satisfactory, based on feedback from a {\em verifier}~\citep{song_good_2024, qin_large_2024, brown_large_2024, hassid_larger_2024, li_common_2024}. Unlike techniques such as majority voting where gains from inference scaling quickly plateau (\cref{table:inference-scaling-hierarchy}), resampling has given rise to the hope of usefully scaling inference compute by many orders of magnitude.

We provide evidence that tempers this assumption. Our key concern is that the generalization gap --- where a model performs well on benchmarks but fails to generalize to the real-world --- is amplified when using repeated sampling to lift the performance of weaker models. 

Specifically, we study the use of unit tests as verifiers for coding benchmarks, to see if inference scaling for less capable models allows us to match the accuracy of more capable models. We make the following contributions.

\xhdr{Review of inference scaling techniques and their limitations (\cref{sec:scaledinferencecomputewithverifiers})} We review papers on inference scaling, categorizing the primary techniques and listing their domain-specific applications and known limitations.

\xhdr{Demonstration of generalization gap (\cref{sec:worsegeneralization})} We provide empirical evidence on two benchmarks, HumanEval+ and MBPP+ \citep{liu_is_2023}, showing that the apparent gains from resampling with imperfect verifiers are unlikely to translate into real-world performance. Despite achieving comparable results to stronger models on standard unit tests, less capable models suffer from a larger generalization gap—producing incorrect solutions that fail the extended test suite (false positives) at higher rates than stronger models. 

In particular, we observe that even if given an {\em infinite} inference budget, in many cases a weaker model cannot match the performance of a single invocation of a sufficiently strong model.

\xhdr{Empirical analysis to understand the limitations of inference scaling with imperfect verifiers (\cref{sec:theorysection})} We examine how introducing a cost (negative utility) for returning false positives impacts the optimal number of resampling attempts on HumanEval+. We find that even with an infinite inference budget, the optimal number of samples is often finite and very low (e.g., $K \leq 5$ in \cref{fig:scaling_four_models}). Hence, resampling quickly reaches a point of diminishing returns without bridging the performance gap for smaller models.  If the cost of an incorrect solution is higher than the benefit of a correct solution, the optimal $K$ can be {\em zero} --- the risk of a false positive for a weak model is high enough that it is effectively useless (\cref{fig:scaling_four_models}). In \cref{app:theoretical-model} we present a theoretical model that complements the findings in this section.

\xhdr{Evidence that this affects code quality beyond correctness (\cref{sec:beyondcorrectness})} 
We show that the reliance on imperfect verifiers not only affects the functional correctness but also overall quality of the generated code. We evaluate candidate solutions on HumanEval+ based on various readability metrics such as adherence to naming conventions (that we specify in the prompt) like \texttt{snake\_case} and \texttt{camelCase}, line-level commenting, and guidelines regarding the maximum line length and number of lines in function implementations. We find that false positive solutions are lower quality across all models and metrics when compared to true positive solutions. While other aspects of code quality such as simplicity and modularity are harder to test automatically, we speculate that the same pattern holds for those properties as well. 

We also conduct a qualitative analysis to identify recurring error types causing a larger generalization gap for weaker models.

\begin{figure*}[ht]
    \centering
    \includegraphics[width=0.9\textwidth]{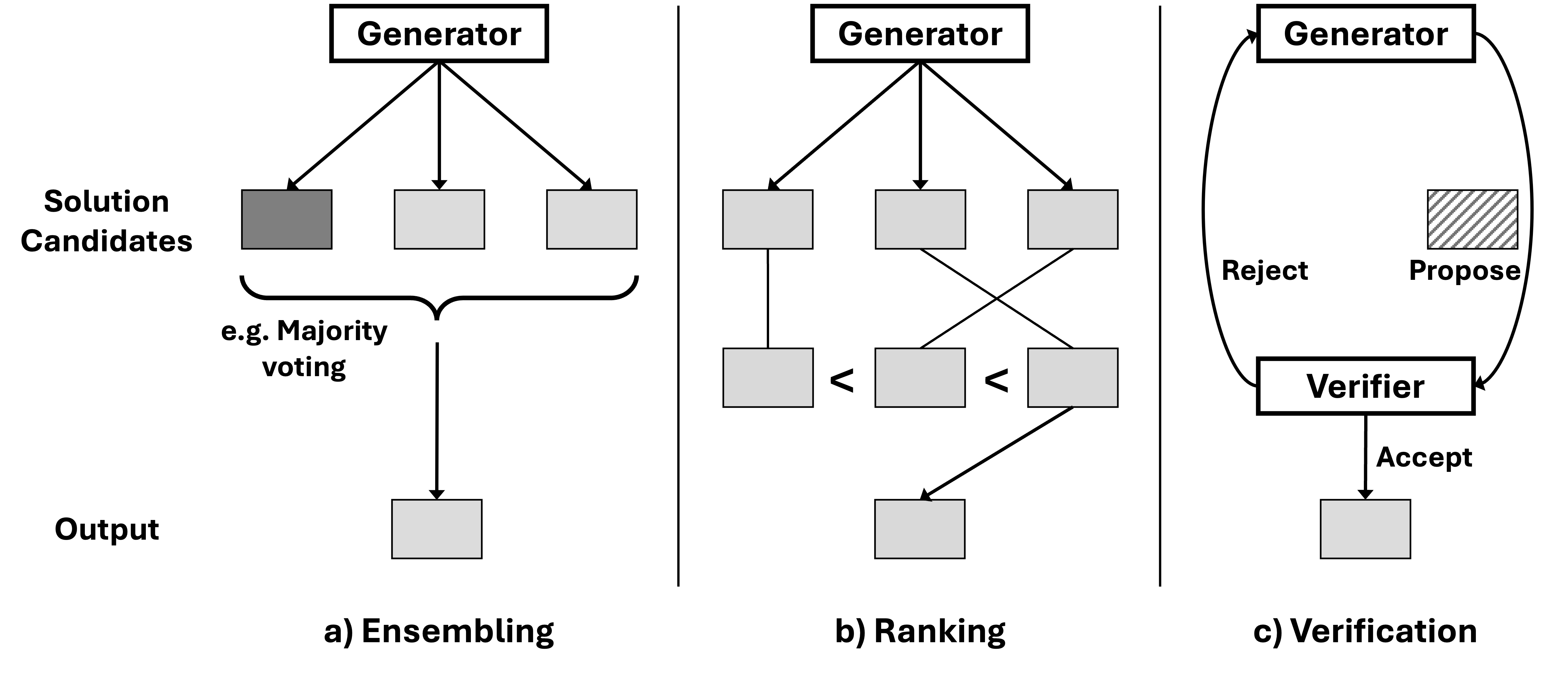}
    \caption{Schematic illustration of various resampling techniques for inference scaling.  \vspace{-0.4cm}}
    \label{fig:techniques}
\end{figure*}

Our findings have three additional implications. First, they show the importance of building highly accurate verifiers. This goal might benefit from treating verification technology as a specialized subfield with its own metrics and benchmarks. This is especially true for training-time uses: models trained with feedback from imperfect verifiers may learn to exploit weaknesses in the verifier rather than solve tasks robustly, potentially leading to safety concerns \citep{krakovna_avoiding_2020, amodei_concrete_2016}. 

Second, the use of imperfect verifiers as the ground truth for evaluation is flawed. We used HumanEval+ and MBPP+ for evaluation; the gaps we identify would have been invisible if we had used HumanEval and MBPP both as verifiers and as benchmarks. While the limitations of these benchmarks for measuring absolute performance are well known \citep{zhang_naturalcodebench_2024, liu_is_2023}, our results show that they might result in misleading comparisons between models as well.

Third, our findings highlight limitations in resampling-based data curation for reasoning models. Reasoning models rely on datasets curated through rejection sampling against verifiers. When these verifiers are imperfect, the curated datasets risk including mislabeled examples, which incurs a cost on model performance. This introduces a bottleneck: without stronger base models or highly accurate verifiers, the gains from resampling-based data curation to train reasoning models are likely limited.

While we do not claim that resampling is the predominant scaling technique, our findings suggest that the persisting gap between oracle and imperfect verifiers should be taken seriously and could pose limitations across inference scaling strategies. We invite research on ways to mitigate the issues identified in this work.

\vspace{-0.1cm}
\section{Scaling inference compute with verifiers}
\label{sec:scaledinferencecomputewithverifiers}

\begin{table*}[h!]
  \centering
  \caption{\textbf{Overview of inference scaling techniques.} This table shows the main categories of techniques for inference scaling along with their descriptions and known limitations. Note that rankers, often implemented with reward models, are sometimes referred to as verifiers and the boundary can be unclear.}
  \begin{adjustbox}{width=\textwidth, center}
    \begin{tabular}{p{2.3cm}p{4.4cm}p{7.6cm}}
      \textbf{Technique} & \textbf{Description} & \textbf{Limitations} \\
      \toprule
       Ranking & \raggedright Scores and ranks the best samples from multiple candidates \citep{cobbe_training_2021, hassid_larger_2024, liu_tinygsm_2023, lightman_lets_2023, liu_improving_2023, hosseini_v-star_2024, kirchner_prover-verifier_2024, setlur_rewarding_2024, snell_scaling_2024, vacareanu_general_2024, chen_when_2024} & \vspace{-0.4cm}\begin{itemize}[leftmargin=*, itemsep=0pt]
        \item Doesn't scale with sample budget \citep{brown_large_2024} 
        \item Underperform compared to other methods \citep{zhang_generative_2024}
      \end{itemize} \\
      \midrule
       Majority Voting & \raggedright Using consensus among multiple samples to determine final output\citep{wang_self-consistency_2023,li_more_2024, wang_soft_2024} & \vspace{-0.4cm}\begin{itemize}[leftmargin=*, itemsep=0pt]
        \item Hurts performance on hard tasks and non-monotonous scaling under task heterogeneity \citep{chen_are_2024}
        \item Sample inefficient for queries with many answer possibilities \citep{wang_soft_2024}
        \item Limited applicability for tasks with non-discrete answers
      \end{itemize} \\
      \midrule
       Oracle \newline Verification & \raggedright Leverages ground-truth evaluator for free until correct solution is found \citep{xin_deepseek-prover_2024, first_baldur_2023} & \vspace{-0.4cm}\begin{itemize}[leftmargin=*, itemsep=0pt]
        \item Not available for most domains
        \vspace{-0.3cm}
      \end{itemize} \\
      \midrule
      \rowcolor{Gray}
       Imperfect \newline Verification \newline \textbf{(This paper)} & \raggedright Scores and accepts or rejects candidate solutions \citep{zhang_generative_2024, davis_networks_2024, yao_tree_2023, gundawar_robust_2024, kambhampati_llms_2024} & \vspace{-0.4cm}\begin{itemize}[leftmargin=*, itemsep=0pt]
        \item Bigger generalization gap for weaker models (\cref{sec:worsegeneralization})
        \item Optimal number of samples is finite and low (\cref{sec:theorysection})
        \item Low code quality of false positives (\cref{sec:beyondcorrectness})
      \end{itemize} \\
      \bottomrule
    \end{tabular}
  \end{adjustbox}
  \label{table:inference-scaling-hierarchy}
\end{table*}

\cref{table:inference-scaling-hierarchy} provides an overview of the main techniques for scaling inference compute with LLMs. Some methods such as majority voting \citep{wang_self-consistency_2023, chen_are_2024} or resampling using verifiers \citep{brown_large_2024, xin_deepseek-prover_2024} generate many candidate solutions and then select one. Other methods such as reasoning \citep{wei_chain--thought_2023} and critique  \citep{shinn_reflexion_2023, madaan_self-refine_2023}  refine a single solution. In practice, these methods can be combined in flexible ways and the distinction between them is not always clear (\cref{app:edgecases}). Note that our notion of inference scaling excludes methods such as those used to train OpenAI's o1 series of models, since we are only looking at improvements during inference time to available language models, rather than training improvements.

All these methods except verifier-based resampling are known to have important limitations that cast doubt on how much scaling is truly possible, as summarized in \cref{table:inference-scaling-hierarchy}. Resampling using verifiers has a different control flow than other methods, which gives it an intuitive appeal (\cref{fig:techniques}): we can potentially regenerate solutions {\em indefinitely} until one is correct. This enthusiasm around resampling is partly driven by the empirically observed  \textit{inference scaling laws}, which suggest that the fraction of tasks for which we find at least one correct solution scales predictably with the number of samples over multiple orders of magnitude \citep{brown_large_2024}.

However, the usefulness of this depends on the availability of a capable verifier \citep{davis_networks_2024}. In some settings, we may have an {\em oracle verifier}, such as a proof checker, that does not suffer from false positives --- that is, if the proof checker verifies the proof, it is guaranteed to be correct. False negatives of the verifier (including nontermination under a fixed compute budget) are less of a problem, as one can simply generate more samples until a true positive is found. It is possible that {\em every} correct solution is a false negative of the verifier, but it is unclear if this is a problem that arises in practice.

\begin{table*}[h!]
\centering
\begin{adjustbox}{width=\textwidth,center}
\begin{tabular}{p{3.5cm}p{1.3cm}p{2.2cm}p{2.7cm}p{5.2cm}}
Paper & \parbox[t]{1.3cm}{Verifier\\ Category} & Verifier Type & Domain & Verifier Implementation \\
\midrule
\citet{chen_codet_2022} & Imperfect & Unit tests & Coding  & Checks agreement of tests and samples \\
\rowcolor{Gray}
\citet{shinn_reflexion_2023} & Imperfect & LM-as-judge,\newline Unit tests & Coding, QA & LLM evaluator generates decision rewards \\
\citet{yao_tree_2023} & Imperfect & LM-as-judge & Planning & LLM evaluates reasoning steps \\ 
\rowcolor{Gray}
\citet{first_baldur_2023} & Oracle & Proof checker & Math & Proof checker \\ 
\citet{thakur_-context_2024} & Oracle & Proof checker & Math & Proof checker \\ 
\rowcolor{Gray}
\citet{yang_leandojo_2023} & Oracle & Proof checker & Math & Proof checker \\
\citet{wang_lego-prover_2023} & Oracle & Proof checker & Math & Proof checker \\
\rowcolor{Gray}
\citet{azerbayev_llemma_2024} & Oracle & Proof checker & Math & Proof checker, Majority voting \\
\citet{huang_mustard_2024} & Oracle & Proof checker & Math & Proof checker \\
\rowcolor{Gray}
\citet{xin_deepseek-prover_2024} & Oracle & Proof checker & Math & Proof checker \\ 
\citet{brown_large_2024} & Oracle & Proof checker & Math & Proof checker\\
\rowcolor{Gray}
\citet{davis_networks_2024} & Imperfect & LM-as-judge & QA, Math & LLM judges correctness of generations \\
\citet{hassid_larger_2024} & Imperfect & Unit tests & Coding & Unit tests \\ 
\rowcolor{Gray}
\citet{zhang_generative_2024} & Imperfect & Generative RM & Math & Verification as part of the model output \\
\citet{zhuge_agent-as--judge_2024} & Imperfect & Agent-as-judge & Agents, Coding & Agents evaluate outputs of other agents \\
\rowcolor{Gray}
\citet{kapoor_ai_2024} & Imperfect & Unit tests & Coding & Unit tests \\
\citet{saad-falcon_archon_2024} & Imperfect & LM-as-judge & Coding, Reasoning & LLM judges correctness of generations \\
\rowcolor{Gray}
\citet{liang_improving_2024} & Imperfect & Program-of-thought & Coding, Math  & Checks CoT against generated PoT \\
\citet{cook_ticking_2024} & Imperfect & LM-as-judge & Instruction-following & LLM checks answer against generated checklists \\
\rowcolor{Gray}
\citet{gundawar_robust_2024} & Imperfect & Agent-as-judge & Travel planning & Pre-defined constraints verified by critic agents \\
\end{tabular}
\end{adjustbox}
\caption{Survey of papers on LLM verification methods, their approaches, and specific verifier implementations.\vspace{-0.4cm}}
\label{table:verification-survey}
\end{table*}

But in other settings such as coding and reasoning, we only have {\em imperfect verifiers} such as unit tests or LM judges, which suffer from false positives: incorrect solutions that nonetheless pass the verifier. In these settings, we don't have easy methods to guarantee the correctness of generated solutions at inference time. As a result, we cannot distinguish between false positives and true positives simply by increasing the compute budget. We survey papers that use verifiers in \cref{table:verification-survey}.

In this paper, we investigate the effect of scaling inference compute with access to imperfect verifiers. We distinguish ranking from verification: ranking selects the best among multiple candidates, while verification accepts or rejects each candidate independently against a correctness criterion.

\vspace{-0.1cm}
\section{Repeated sampling with weaker models leads to worse generalizability}
\label{sec:worsegeneralization}

In computer programming tasks, unit tests are commonly employed as verifiers to assess the correctness of candidate solutions generated by language models. While unit tests are practical and efficient, they often suffer from imperfect test coverage, leading to false positives where incorrect solutions pass the tests \citep{gulwani_program_2017}. This affects many benchmarks such as HumanEval \citep{chen_evaluating_2021}, APPS \citep{hendrycks_measuring_2021}, or MBPP \citep{austin_program_2021}. This imperfection raises the question: Do less capable models produce false positives—implementations that pass the standard unit tests but fail the comprehensive ones—at a higher rate than stronger models? 

\xhdr{Experimental setup} To investigate this, we conducted experiments on two widely used coding benchmarks: HumanEval+ and MBPP+. 
MBPP consists of simple programming tasks designed to evaluate the basic coding abilities of models \citep{austin_program_2021}. 
HumanEval+ and MBPP+ are extensions of the original HumanEval and MBPP benchmarks \citep{liu_is_2023} and contain additional hidden test cases to assess correctness beyond the unit tests included in the original benchmarks.

We evaluated models of varying capabilities, including weak and stronger models, generating at least 200 samples per task and model on HumanEval+ (1,000 for Command Light, to minimize tasks without any passing solutions) and 50 samples per task on MBPP+. To assess the generalization gap, we then evaluated solutions that passed the original benchmark test sets on the more comprehensive hidden test cases (see \cref{app:beyondcorrectness} for details). These tests are extensive, and we assume that solutions that pass the full set of tests are correct. (If this assumption is not true, the generalization gaps that we reveal only grow bigger.)

\begin{figure*}[ht!]
    \centering
    \begin{minipage}[b]{0.48\textwidth}
        \centering
        \includegraphics[height=4.8cm]{humaneval_no_fns_no_lines.png}
        \caption*{HumanEval+}
    \end{minipage}
    \hfill
    \begin{minipage}[b]{0.48\textwidth}
        \centering
        \includegraphics[height=4.8cm]{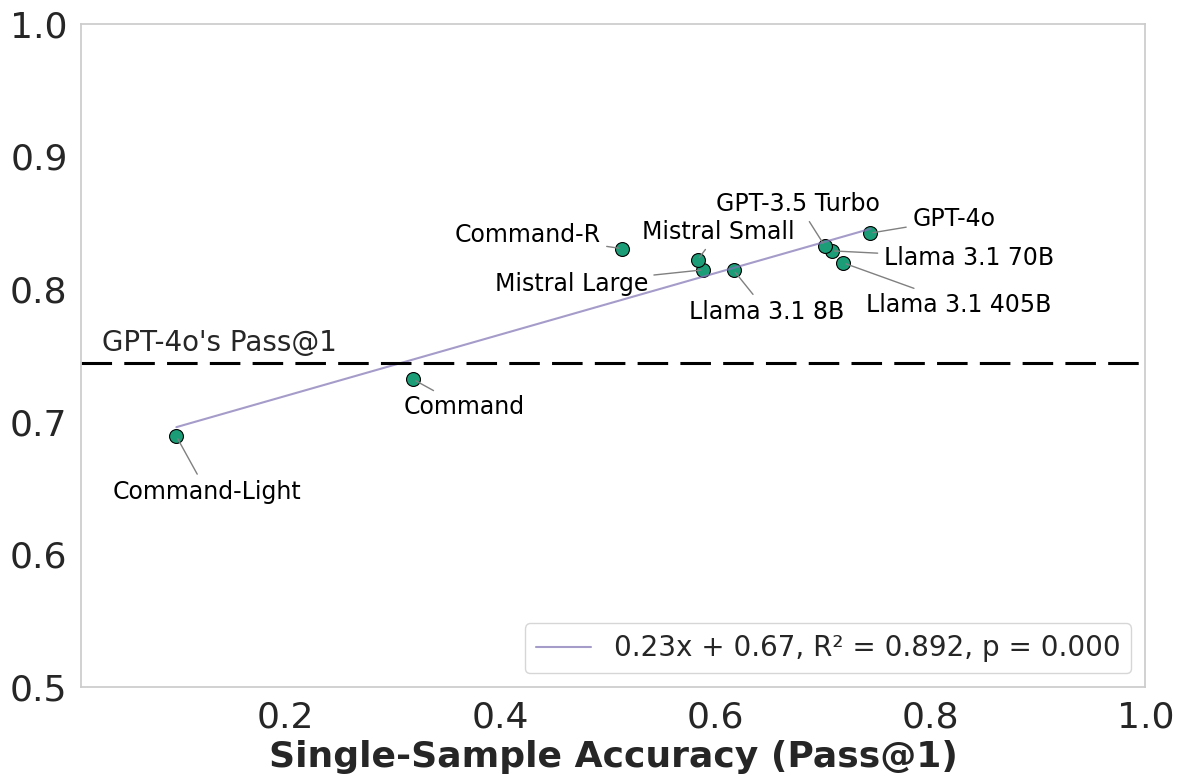}
        \caption*{MBPP+}
    \end{minipage}
    \caption{\textbf{Generalization gap with infinite compute budget.} We show the relationship between the accuracy of individual samples (x-axis) and the achievable accuracy given an infinite compute budget and limited unit tests (y-axis; note that it starts at 0.5). We evaluate performance on the extended test suites of HumanEval+ and MBPP+, using the unit tests from the original benchmarks as verifiers. For both benchmarks, the trend is that less capable models are more likely to generate false positives than stronger models. In \cref{app:worsegeneralizaton}, we show our results with the full y-axis as well as upper (lower) bounds on the conditional accuracy accounting for tasks for which we did not observe \textit{any} solutions passing the unit tests.\vspace{-0.2cm}}
    \label{fig:scatterplot_sec2}
\end{figure*}

\xhdr{Findings} Weaker models exhibit a higher probability of producing false positives compared to stronger models (\cref{fig:scatterplot_sec2}). This probability scales inversely with the true capability. This linear relationship holds with remarkable consistency across models of various families, including Cohere's Command models, GPT-4o, and the Llama-3.1 family. This suggests that while weaker models appear to perform well on standard benchmarks through increased sampling, they fail to generalize effectively and, importantly, they generalize worse than more capable models. They tend to generate fragile solutions that exploit the limitations of the unit tests. We speculate that this is because weaker models' ``true understanding'' of the programming tasks is worse.

The empirical results reinforce a core insight. Suppose $\mathbb{P}_{strong}(\text{Correct}) > \mathbb{P}_{weak}(\text{Correct} | \text{Pass Verifier})$. That is, the single-sample accuracy of a strong model exceeds that of a weaker model, even conditioned on the weaker model passing the base unit tests. Then the weaker model cannot match the performance of a single invocation of the stronger model, no matter how big the compute budget for the weaker model. In \cref{fig:scatterplot_sec2}, this is shown by a horizontal line. No model below the line can match the performance of GPT-4o through resampling. 

The effect is largely driven by a subset of the tasks where the unit tests are poor. When we limit the analysis to these tasks, the relationship is even more pronounced \cref{app:worsegeneralizaton}. 

Note that our results rely on human-generated unit tests as verifiers. In practice, we might expect to use language-model-generated unit tests for inference-time verification. It is an open question as to how the findings change when the unit-test verifiers are LLM-generated.

\vspace{-0.1cm}
\section{How many samples are optimal?}
\label{sec:theorysection}

In the previous section we looked at the behavior of resampling in the limit as the number of samples grows large. Now we look at inference scaling curves, which allow us to study how accuracy varies as a function of the number of samples. 

We add one important detail: we model the cost of false positives, such as code that passes unit tests but has subtle bugs. The cost of bugs (which might result in buggy software being deployed) is not easily comparable to the labor-saving benefit of correct solutions, and this cost-benefit ratio can vary greatly depending on the application. So we consider many possible values for the cost-benefit ratio, including zero, which is the setting considered in previous work on inference scaling. The ratio can potentially be much higher than 1 in some applications, such as security sensitive ones, since bugs might translate to exploitable vulnerabilities.

\begin{figure*}[!t]
\centering
\begin{minipage}{\textwidth}
    \centering
    \includegraphics[height=0.67cm]{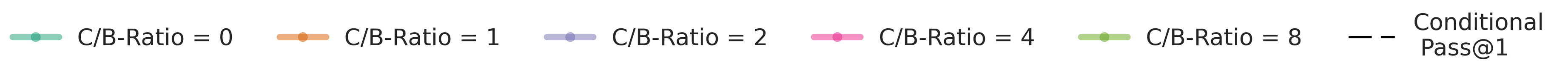}
\end{minipage}

\begin{minipage}{0.25\textwidth}
    \centering
    \includegraphics[height=3.6cm]{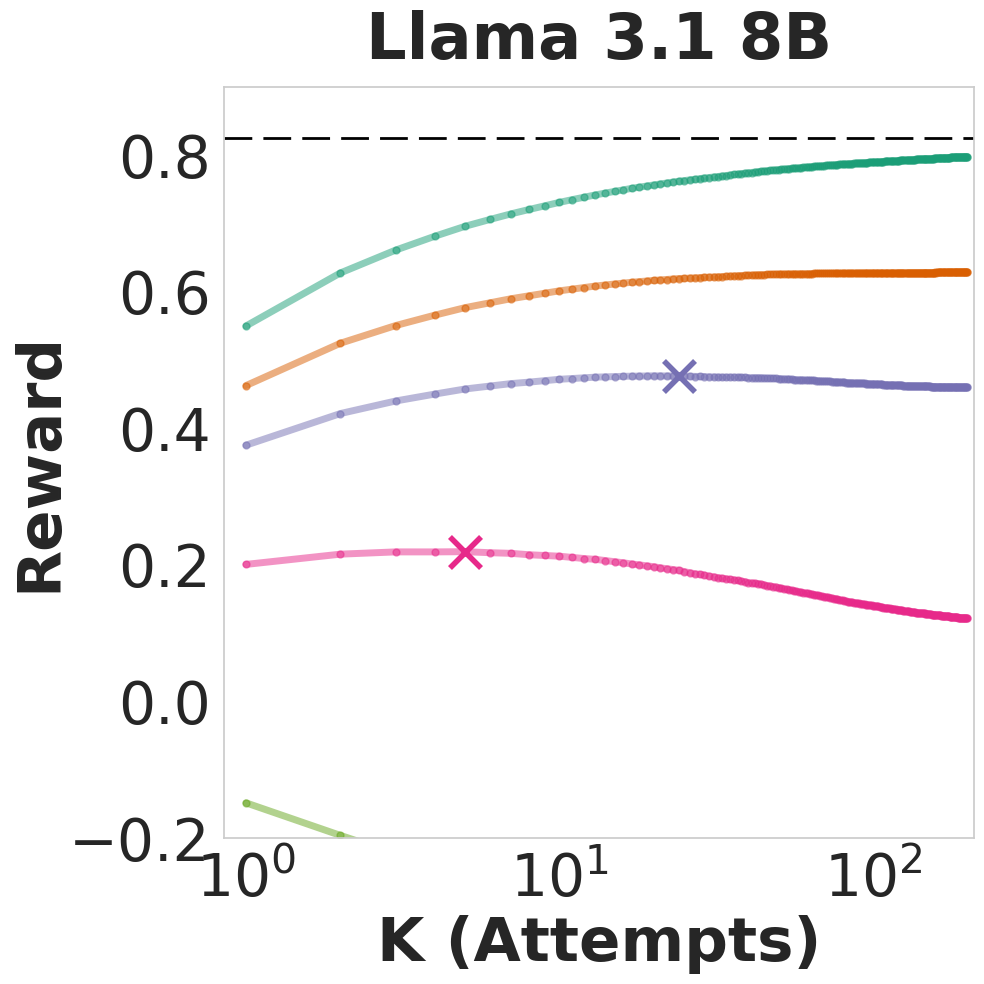}
\end{minipage}%
\hfill%
\begin{minipage}{0.25\textwidth}
    \centering
    \includegraphics[height=3.6cm]{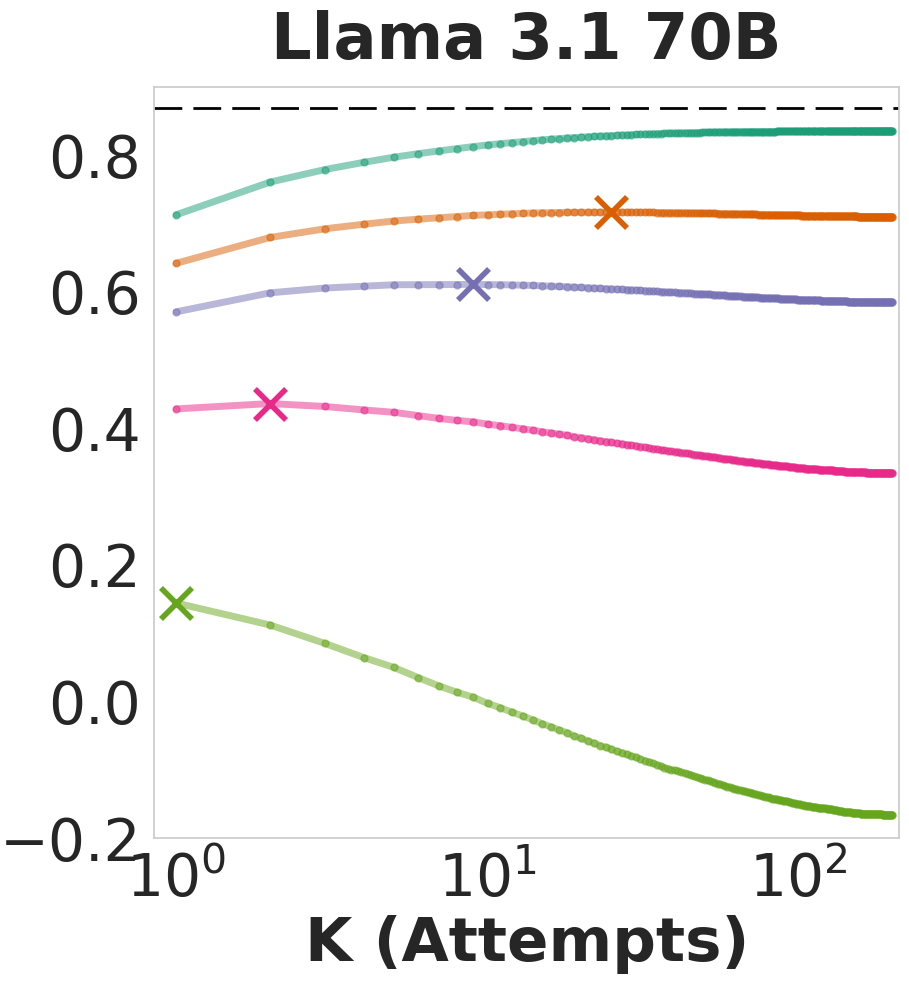}
\end{minipage}%
\hfill%
\begin{minipage}{0.25\textwidth}
    \centering
    \includegraphics[height=3.6cm]{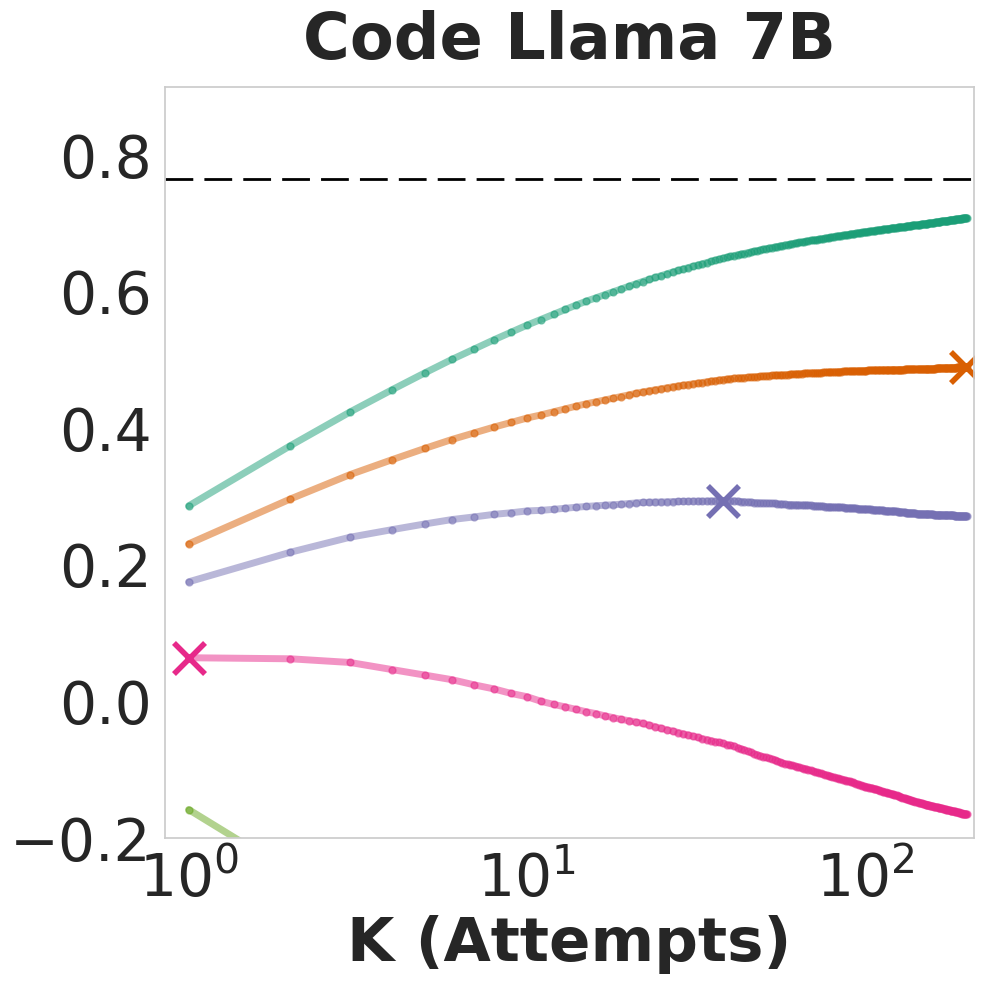}
\end{minipage}%
\hfill%
\begin{minipage}{0.25\textwidth}
    \centering
    \includegraphics[height=3.6cm]{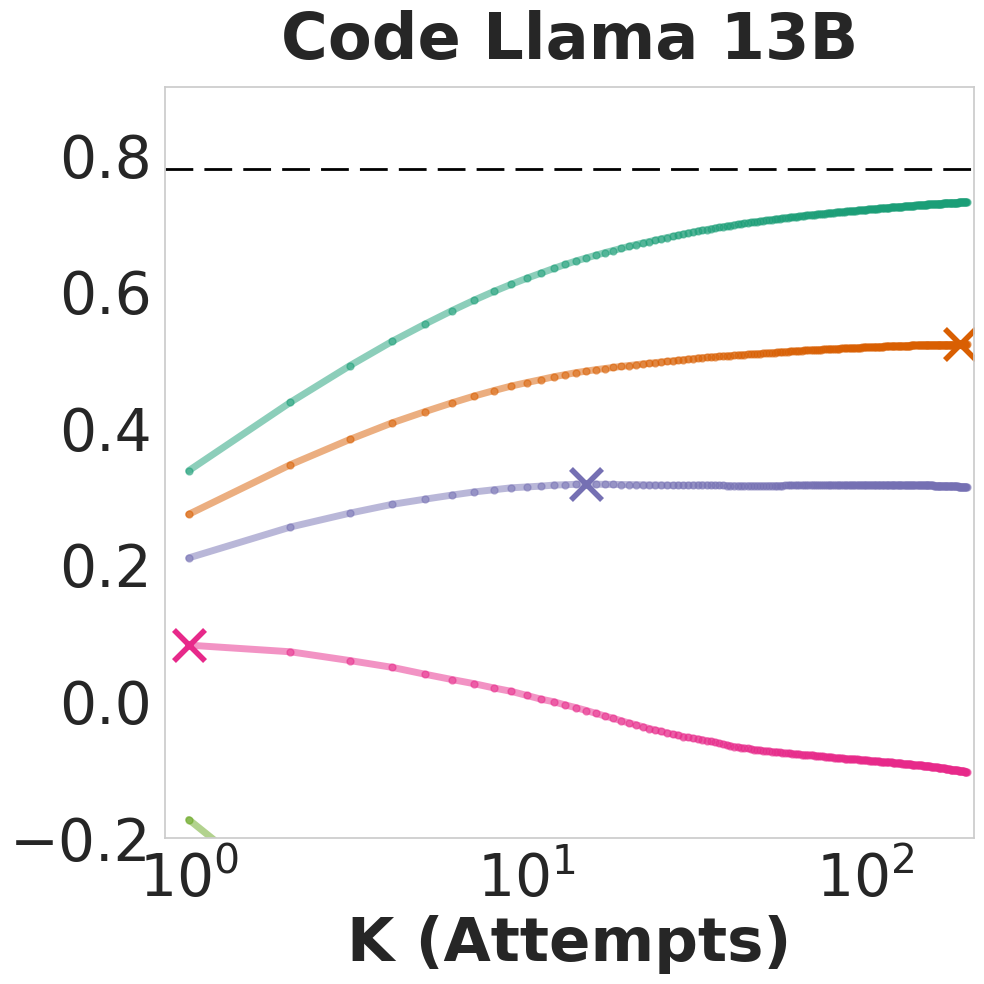}
\end{minipage}
\caption{\textbf{Inference scaling curves in the presence of a cost for false positives.} We show the reward as a function of the number of attempts $K$ across various cost-benefit ratios for the Llama-3.1 and Code Llama model families. Crosses mark the optimal number of samples for each setting. Standard inference scaling curves with no cost (i.e., cost-benefit ratio is 0) are provided for reference. We find that, even at \textit{zero computational cost}, there is a finite optimal number of samples $K$ that is often very low.\vspace{-0.4cm}}
\label{fig:scaling_four_models}
\end{figure*}

\xhdr{Experimental setup} For each model of interest, we generated 200 samples for each task in the HumanEval benchmark. For each K $\leq$ 200, If a passing solution was found within $K$ samples, we assigned rewards based on the outcome: a true positive yielded a benefit of 1, while a false positive incurred a cost, with values set according to different cost-benefit ratios: 0, 1, 2, 4, or 8 (\cref{fig:scaling_four_models}). If no passing solution was found within $K$ samples, we assigned a reward of 0 (both, cost and benefit are 0). We repeated this whole process 1,000 times and computed the mean reward for each $K$. The set of samples was the same in all 1,000 runs, but the order of samples was randomly permuted. This setup allows us to empirically observe the relationship between the number of sampling attempts $K$ and the reward for various cost-benefit ratios. The results are illustrated in \cref{fig:scaling_four_models} for the Llama-3.1 \citep{dubey_llama_2024} and Code Llama \citep{roziere_code_2024} model families. The scaling curves for GPT-4o are included in \cref{app:scaling-curve-gpt4o}.

\begin{figure}[t!]
\centering
\begin{minipage}{0.48\columnwidth}
    \centering
    \includegraphics[height=3.8cm]{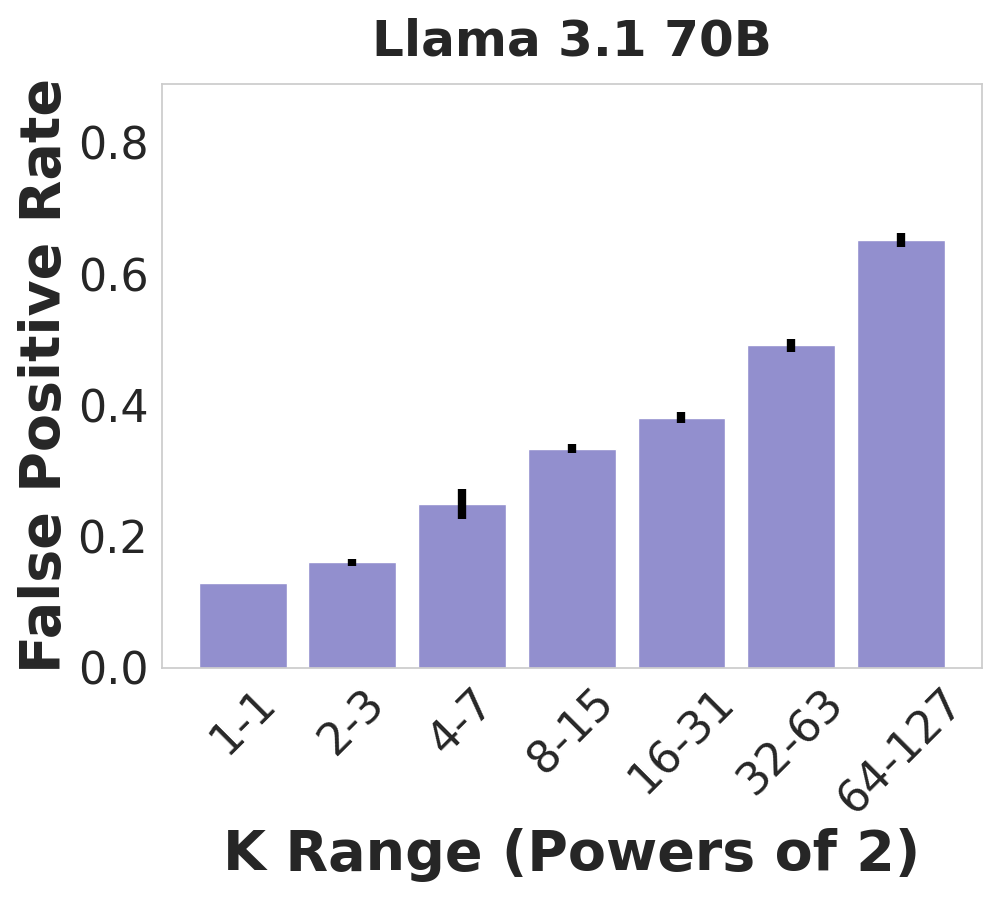}
    \label{fig:fpr-llama}
\end{minipage}%
\begin{minipage}{0.48\columnwidth}
    \centering
    \includegraphics[height=3.8cm]{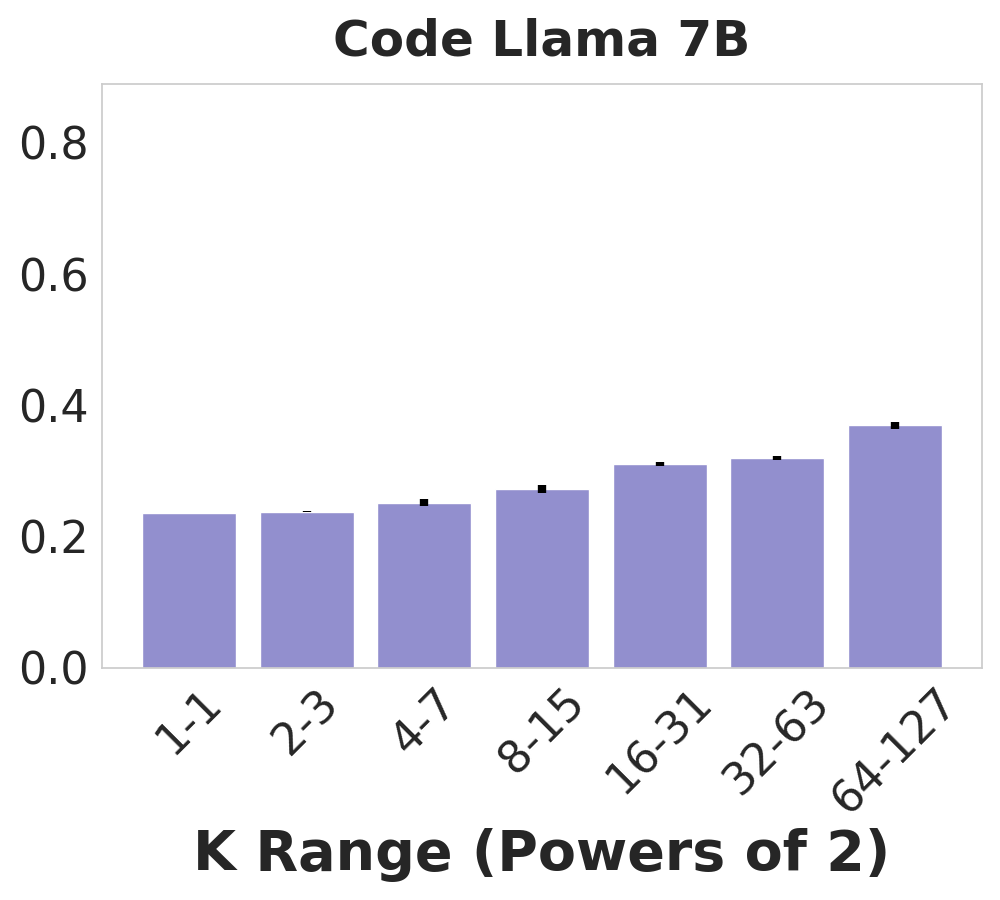}
    \label{fig:fpr-codellama}
\end{minipage}
\caption{False positive rate as a function of the number of attempts $K$ for Llama 3.1 70B and Code Llama 7B on HumanEval+. We include plots for additional models in \cref{app:theoretical-model}.\vspace{-0.3cm}}
\label{fig:false_positive_rate}
\end{figure}

\begin{figure*}[ht]
\centering
\begin{minipage}{0.48\textwidth}
    \centering
    \includegraphics[height=3cm]{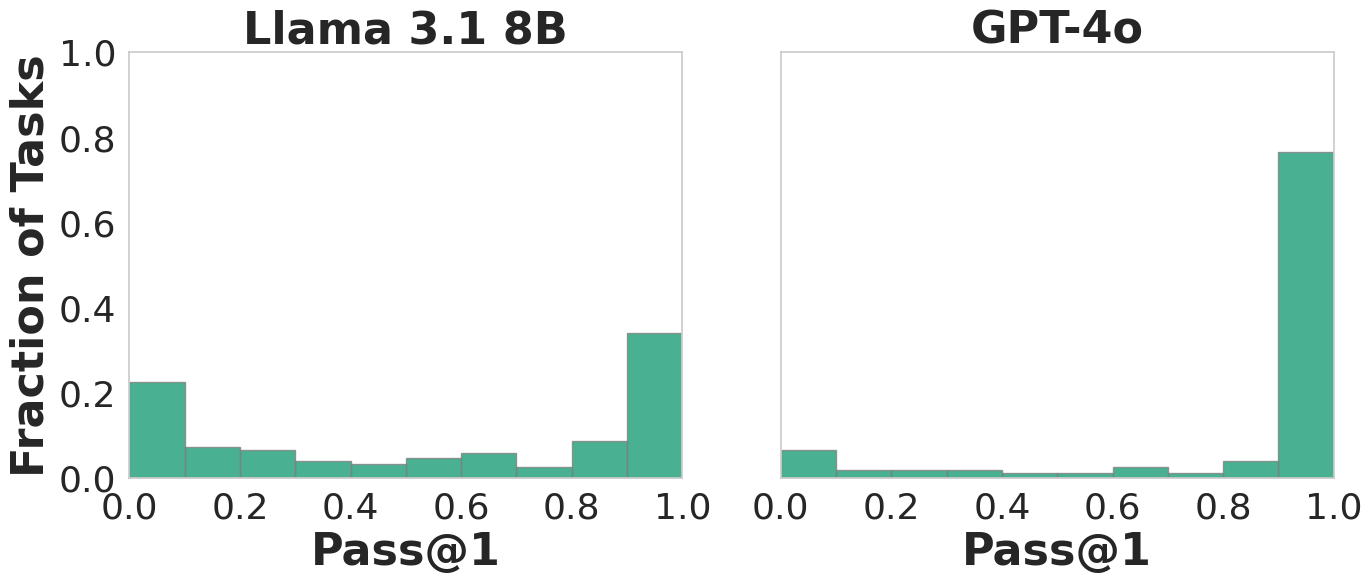}
    \caption*{HumanEval+}
\end{minipage}%
\hfill%
\begin{minipage}{0.48\textwidth}
    \centering
    \includegraphics[height=3cm]{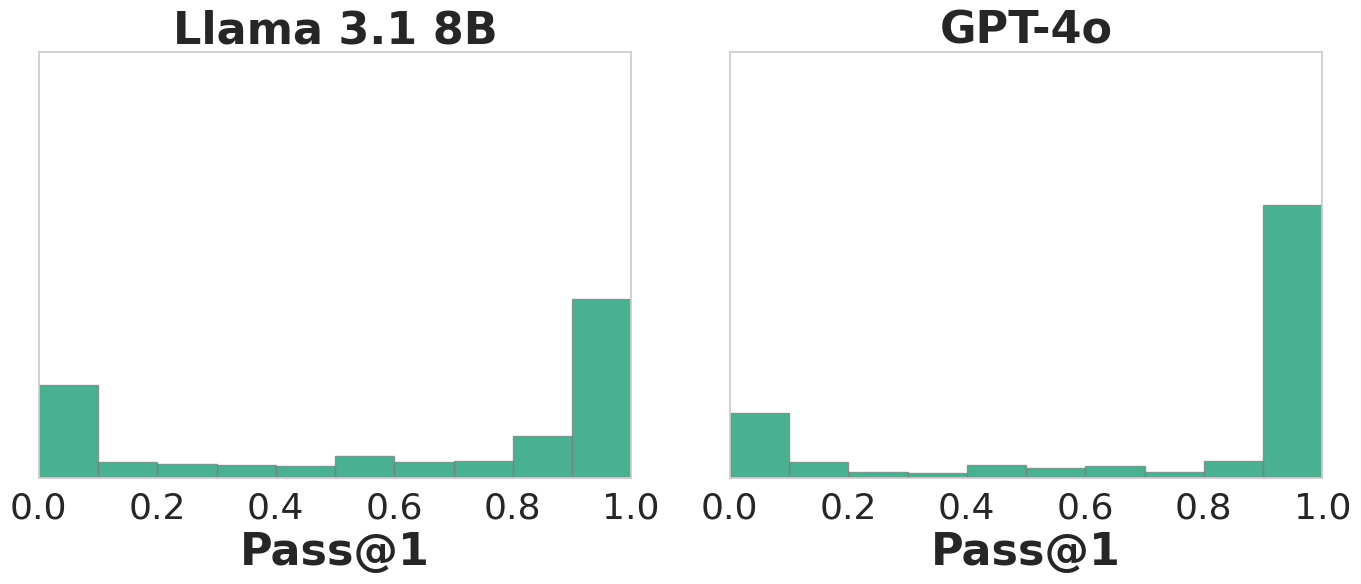}
    \caption*{MBPP+}
\end{minipage}
\caption{Distribution of task difficulties for Llama 3.1 8B and GPT-4o on HumanEval+ and MBPP+. We include barplots for all models on both benchmarks in \cref{app:worsegeneralizaton}.\vspace{-0.2cm}}
\label{fig:diff_dist}
\end{figure*}

\xhdr{Findings} The results show that the effectiveness of repeated sampling quickly reaches a point of diminishing and even negative returns. Each additional attempt brings a trade-off: although it might yield a correct solution, it might instead yield a false positive, and the false positive rate increases with $K$ (\cref{fig:false_positive_rate}). At first this is surprising, since sampling is a memoryless process. To understand why it happens, we need to look at the distribution of task difficulty (Figure \ref{fig:diff_dist}), which turns out to be strongly bimodal. The easy tasks get solved within a few attempts, and for the remaining harder tasks, false positives are more likely. This aligns with findings by \citet{chen_are_2024}, who observed a similar inverse U-shaped accuracy curves explained by the heterogeneity in task difficulties.

Thus, even with \textit{zero computational cost}, the optimal number of samples is finite and low (\cref{fig:scaling_four_models}). For example, at a cost-benefit ratio of 4, the optimal number of samples is $K \leq 5$ for all four models. If the ratio is high enough, the optimal number of samples is zero --- the expected cost of a false positive outweighs the expected benefit of a correct solution, so the reward is always negative and it is best not to attempt a solution at all.

We note one important caveat: for some models such as Llama-3.1-70B, the false positive rate increases dramatically with $K$, whereas for others such as the Code Llama and Command families, the increase is much more gradual, resulting in much higher values of the optimal $K$, especially for low cost-benefit ratios. We have not been able to identify any intuitive reason for this difference.

To summarize, weaker models cannot “sample their way” to top-level performance if the verifier cannot reliably filter out false positives because the risks quickly outweigh the benefits. Our findings in this section align with our theoretical model in \cref{app:theoretical-model}, which generalizes these findings to other benchmarks.

\vspace{-0.1cm}
\section{False positive solutions are low-quality even beyond correctness}
\label{sec:beyondcorrectness}

While correctness is a fundamental criterion for evaluating code generated by LLMs, it is not the only determinant of code quality. High-quality code possesses attributes beyond mere functionality, such as readability, maintainability, and efficiency. Readability simplifies error-checking and is considered one of the most useful properties of high-quality code \citep{borstler_developers_2023}. It can be measured using various metrics, including code length guidelines (e.g., PEP8), adherence to naming conventions like \texttt{snake\_case} or \texttt{camelCase}, and consistent commenting \citep{zheng_beyond_2024}. Intuitively, shorter code with clear variable names and informative comments is generally easier to read and maintain. 

To understand the relationship between imperfect verifiers and code quality, we evaluated the readability of code generated by various models in our setup. 

\xhdr{Experimental setup} On HumanEval, we evaluate the readability scores of candidate solutions that pass standard unit tests and the more comprehensive test suite. For each measure of code readability, we use a different prompt instructing the model to adhere to the desired guidelines (see \cref{app:beyondcorrectness} for detailed prompts). We rely on the prompts and implementation from \citet{zheng_beyond_2024}.

\begin{figure*}[h]
\begin{center}
\includegraphics[height=7cm]{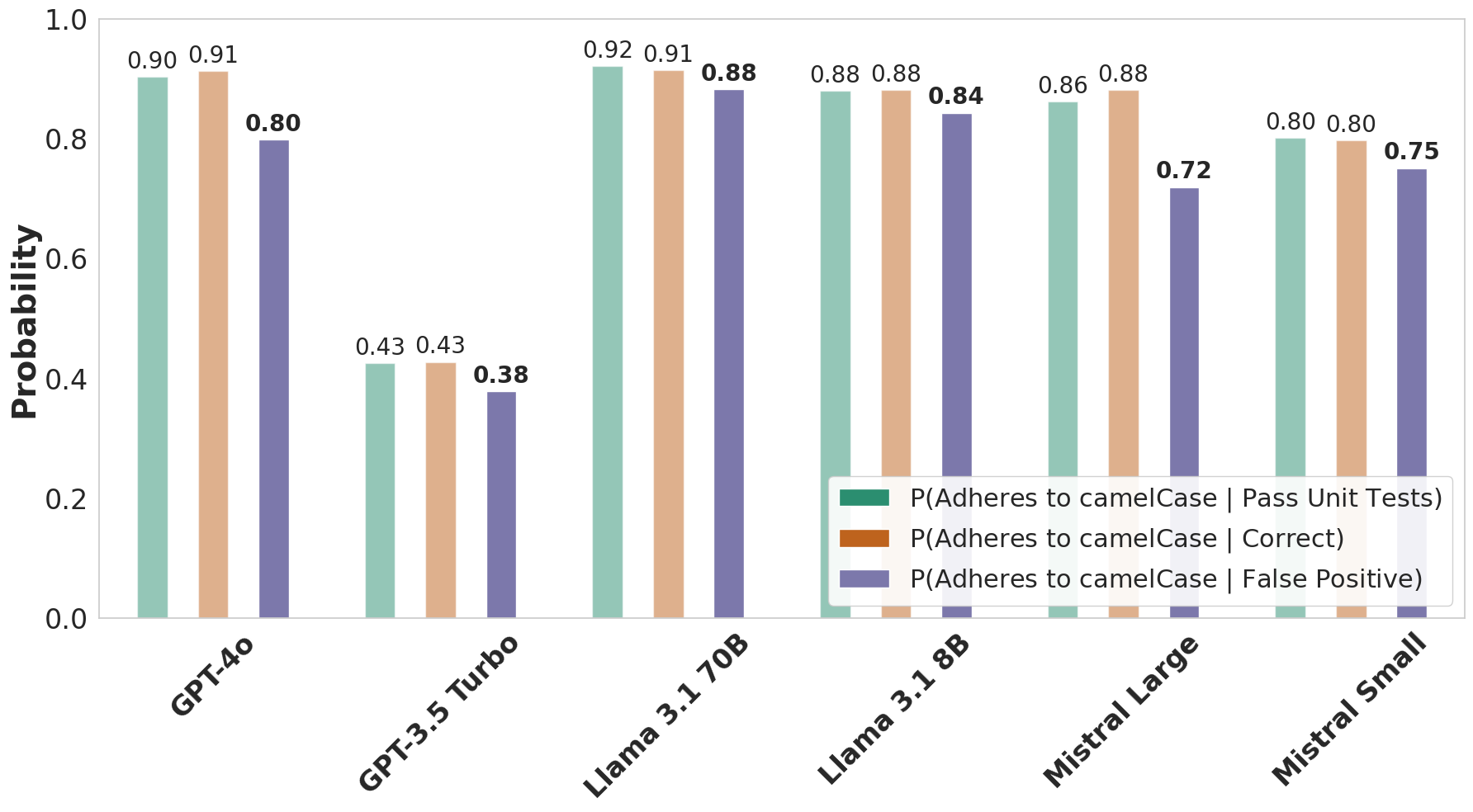}
\end{center}
\caption{\textbf{False positives tend to be lower-quality code than correct implementations.} For example, false positive solutions fail to adhere to the \texttt{camelCase} naming convention more often than robust implementations. \cref{app:fig:codequality} shows that this holds consistently across models and for all four code quality metrics we test. GPT-3.5 exhibits a low relative performance in using \texttt{camelCase} but performs comparably to other models in following \texttt{snake\_case} (\cref{app:beyondcorrectness}). This has also been found in previous work \citep{zheng_beyond_2024}.}
\label{fig:codequality}
\end{figure*}

The results show notable differences in code quality between false positives and robust implementations. False positives, passing only the standard but not the extended unit tests, tend to have worse code quality across all metrics (see \cref{fig:codequality}). This trend is consistent across models of varying capabilities. This suggests that the limitations of imperfect verifiers for coding tasks extend beyond correctness issues but also affect other code characteristics important for software development. This affects weaker models more, given that they are more prone to generate false positives. 

An open question arising from our findings is whether fine-tuning LLMs on code quality metrics could improve not only the quality of generated code but also robustness \citep{jain_llm-assisted_2023}, potentially mitigating the prevalence of false positives.

\xhdr{Qualitative analysis of false positives} To better understand the characteristics of false positive implementations, we randomly sample 10 implementations across all models that pass the standard tests but fail the additional unit tests, 5 for each benchmark. Through manual analysis, we identified several recurring error types. All examples mentioned in the following are included in \cref{app:beyondcorrectness_qualexamples}.

\vspace{-0.3cm}
\begin{enumerate}[leftmargin=11pt, itemsep=-1pt]
    \item \textbf{Logical errors:} Such errors were common. For instance, in the HumanEval/30 task, the model is tasked with returning only positive numbers from a list. The solution shown in \cref{fig:flukeexample} incorrectly converts floats to integers, passing the basic tests that included only integers but failing extended tests that introduce float values.

    \item \textbf{Edge case handling:} Sometimes solutions failed to account for atypical inputs, which happened to not be covered by the standard unit tests. For example, in HumanEval/6, the solution failed to handle an empty list input. It is important to note that tasks in HumanEval and MBPP often are ambiguous as to how edge cases should be handled. For example, for HumanEval/149 some solutions fail because they return an assertion error instead of an empty list for the edge case of getting an empty list as input. We expect that ambiguity should affect weaker and stronger models similarly, but have not tested this.
    \item \textbf{Inefficient implementations:} While most false positives result from logic errors or edge case mishandling, some were also caused by inefficient implementations. For instance, in HumanEval/15, the solution involved a for-loop that became inefficient when handling larger inputs, causing a timeout on the extended tests. Following \citet{liu_is_2023}, we set the timeout such that each candidate solution must compute in less than one second or four times the time it takes to run each test on the ground truth implementation, whichever is greater.
\end{enumerate}

\vspace{-0.2cm}
\section{Discussion}
\label{sec:discussion}

We study a setting where all generators are paired with the same verifier. The verifier has imperfect coverage, but no false negatives. In real-world deployment settings, human-written unit tests are rarely available and we would need to rely on the use of automated test generation techniques. These approaches include symbolic execution \citep{lukasczyk_pynguin_2022}, specialized transformers \citep{tufano_unit_2021}, and LLMs \citep{chen_chatunitest_2024, chen_codet_2022, siddiq_using_2024}. Model-generated tests introduce new challenges including a disparity between verifiers and a risk of false negatives. This could widen the generalization gap. 
We leave an investigation of the impact of model-generated unit tests as a next step.

Resampling is used not only to scale inference but also to train large reasoning models. Many state-of-the-art reasoning models are trained on datasets curated through rejection sampling \citep{sky_t1_2025, deepseek-ai_deepseek-r1_2025, bespoke_stratos}, where verifiers filter out incorrect outputs. However, imperfect verifiers can introduce mislabeled examples, implicitly incurring a cost of false positives. We hypothesize that weaker models paired with imperfect verifiers fail to produce datasets of sufficient quality to train competitive reasoning models, creating a bottleneck: without stronger base models or more accurate verifiers, gains from resampling-based data curation to train reasoning models are limited. 

Although our experiments focus on coding tasks, we want to emphasize that our theoretical results are domain-agnostic: resampling relying on imperfect verifiers with non-zero false positive rates will face the same fundamental ceiling.

Related work has also studied the risks of over-optimizing against imperfect rewards. For example, \citet{gao2022scalinglawsrewardmodel} analyzes how optimizing against proxy reward models can lead to degraded true performance. Our setting differs in that we focus on inference-time resampling, where the ceiling on achievable accuracy arises directly from false positives rather than reward misalignment. 

Finally, our findings weaken support for previous papers' claims that resampling is an effective strategy to increase accuracy by trading off inference time compute \citep{kapoor_ai_2024, chen_are_2024}; here resampling with imperfect verifiers is inherently limited.

\xhdr{Limitations} Our experiments focus on repeated sampling in the context of coding tasks. Coding offers a clear example of the challenges posed by imperfect verifiers, other domains might exhibit different behavior. Future work could extend these findings to tasks such as reasoning \citep{hosseini_not_2024}, web agents \citep{bai_digirl_2024, he_webvoyager_2024}, or agent-user interaction \citep{yao_tau-bench_2024}. Another limitation is prompt sensitivity, which affects LLM evaluations \citep{biderman_lessons_2024, liang_holistic_2023}. While we followed the original authors' implementation provided with the HumanEval+ and MBPP+ benchmarks \citep{liu_is_2023}, prompt engineering could influence false positive generation. Additionally, we did not investigate how benchmark contamination affects our findings, as models could be overly optimized for passing the standard test cases. We did not explore mitigation strategies such refining solutions after they passed the verifier \citep{saad-falcon_archon_2024}. Similarly, we did not test alternative strategies to inference scaling that, e.g., induce more diversity during sampling \citep{wang_planning_2024}. Finally, the optimal number of samples K that we identify depends on the cost-benefit ratio, which varies by application context. While we show that K is finite and often very low across a range of ratios, we do not provide empirical evidence for what specific ratio applies in any given real-world deployment. In security-critical applications, the cost of a false positive may far exceed the benefit of a correct solution, whereas in lower-stakes settings a higher K may remain practical.

\xhdr{Acknowledgments} We thank Boris Hanin, Peter Henderson, Alexander Wettig, Jon Saad-Falcon, and Zachary Siegel for providing valuable feedback on earlier drafts of this work. We thank Cohere for their support of this work with API credits. We also thank Schmidt Sciences for their funding and support of this work.

\xhdr{Reproducibility} We release code to reproduce all experimental results of this paper in a GitHub repository\footnote{URL: \texttt{\url{https://github.com/benediktstroebl/inference-scaling-limits}}}. This repository also contains all code samples for all models used in our experiments. Additionally, we provide an implementation of the theoretical model in \cref{app:theoretical-model} as a Python notebook.

\bibliography{iclr2026/iclr2026_conference}
\bibliographystyle{iclr2026_conference}

\newpage
\appendix
\section{Additional details on \cref{sec:scaledinferencecomputewithverifiers}}
\label{app:scaledinferencecomputewithverifiers}

\subsection{Edge cases in our generator-verifier setting}
\label{app:edgecases}

The setting described in \cref{fig:techniques} considers verifiers and generators as distinct components, where verifiers score and accept or reject individual samples from the generator's output to enable accuracy improvements through resampling. 

This creates interesting edge cases with methods like Chain-of-Verification (CoVe) \citep{dhuliawala_chain--verification_2023}, Tree of Thoughts (ToT) \citep{yao_tree_2023}, and Reward Models (RMs) \citep{snell_scaling_2024} where verification and generation are more tightly coupled. While ToT fits within our framework by producing aggregate scores and a decision on whether the problem is solvable from a given state or not (i.e. potentially rejecting solutions), CoVe differs fundamentally in its verification approach. Instead of producing numeric scores and accepting or rejecting solution candidates, CoVe uses verification of intermediate facts used for answering a question to improve a single response through iterative refinement. This makes CoVe less suitable for inference scaling through resampling because there is no way to distinguish between the quality of multiple samples and using a verifier's verdict for resampling.
\newpage

\section{Additional details on \cref{sec:worsegeneralization}}
\label{app:worsegeneralizaton}

\begin{figure}[H]
\centering
\includegraphics[width=0.9\textwidth]{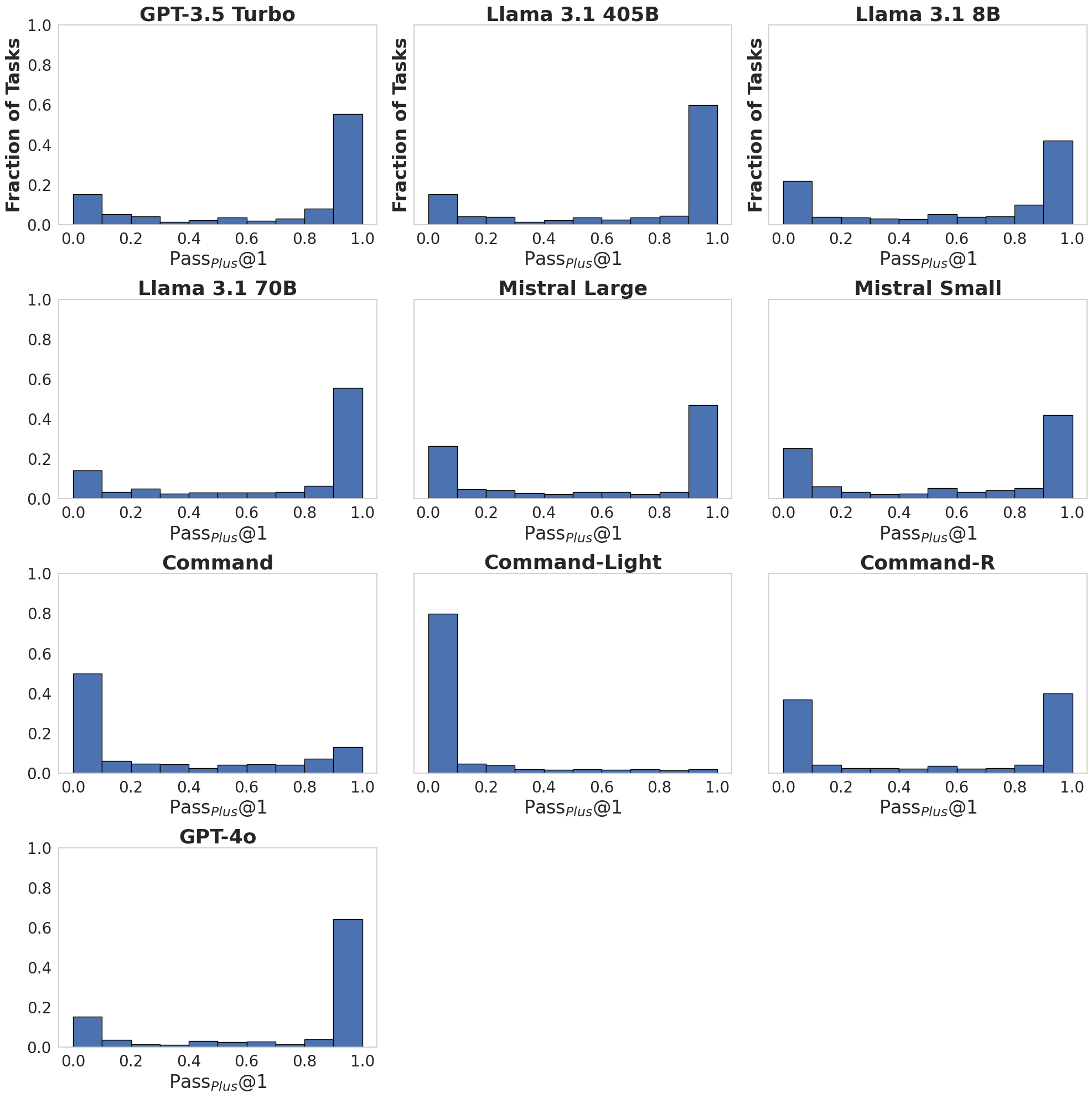}
\caption{Distribution of task difficulties across models on MBPP+.}
\label{fig:task_diff_mbpp}
\end{figure}
\newpage

\begin{figure}[H]
\centering
\includegraphics[width=0.9\textwidth]{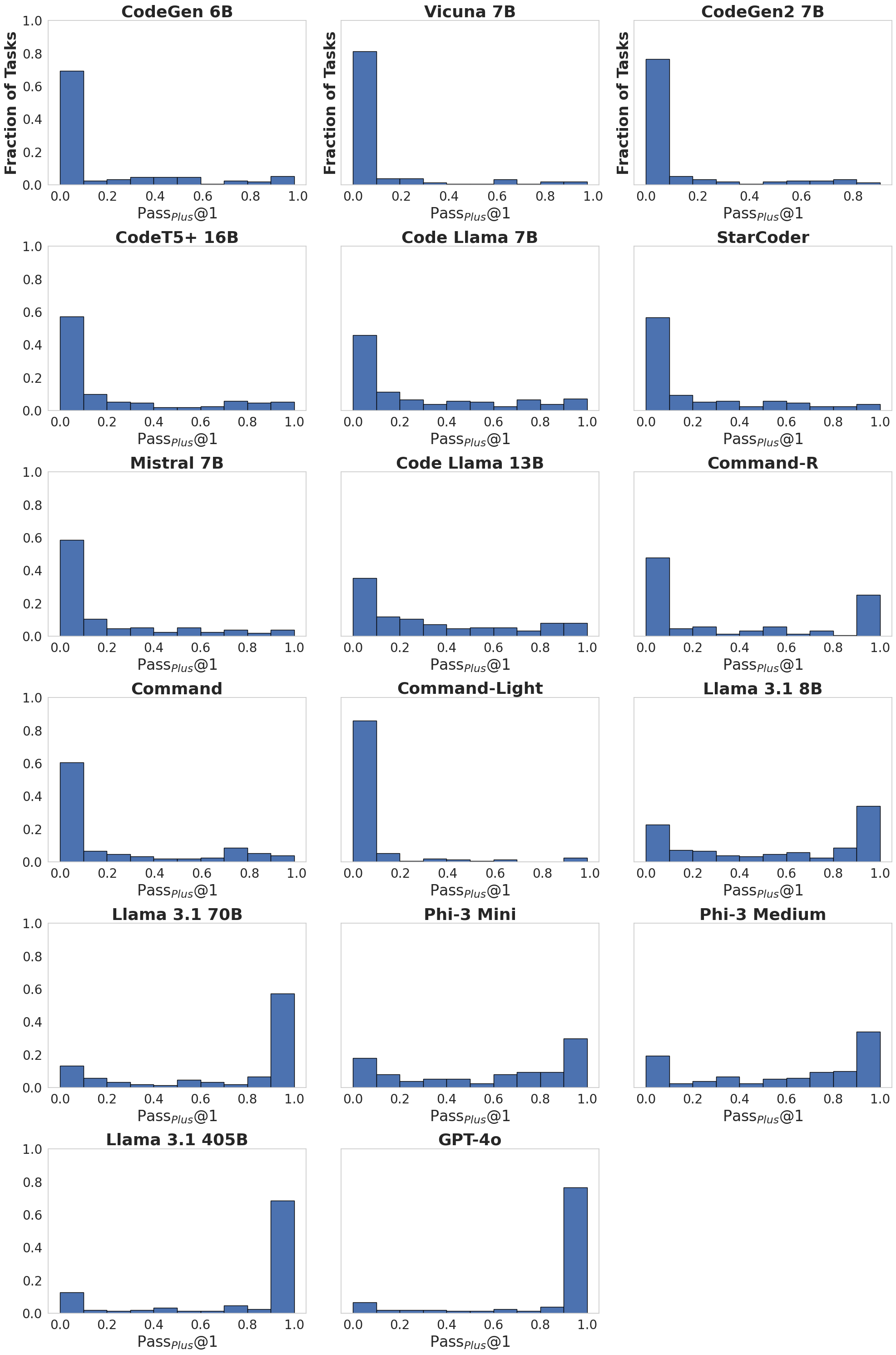}
\caption{Distribution of task difficulties across models on HumanEval+. }
\label{fig:task_diff_humaneval}
\end{figure}

\begin{figure}[H]
\centering
\begin{minipage}[b]{0.24\textwidth}
\centering
\includegraphics[height=3.2cm]{fpr/fpr_llama_3.1_70b.png}
\end{minipage}
\begin{minipage}[b]{0.24\textwidth}
\centering
\includegraphics[height=3.2cm]{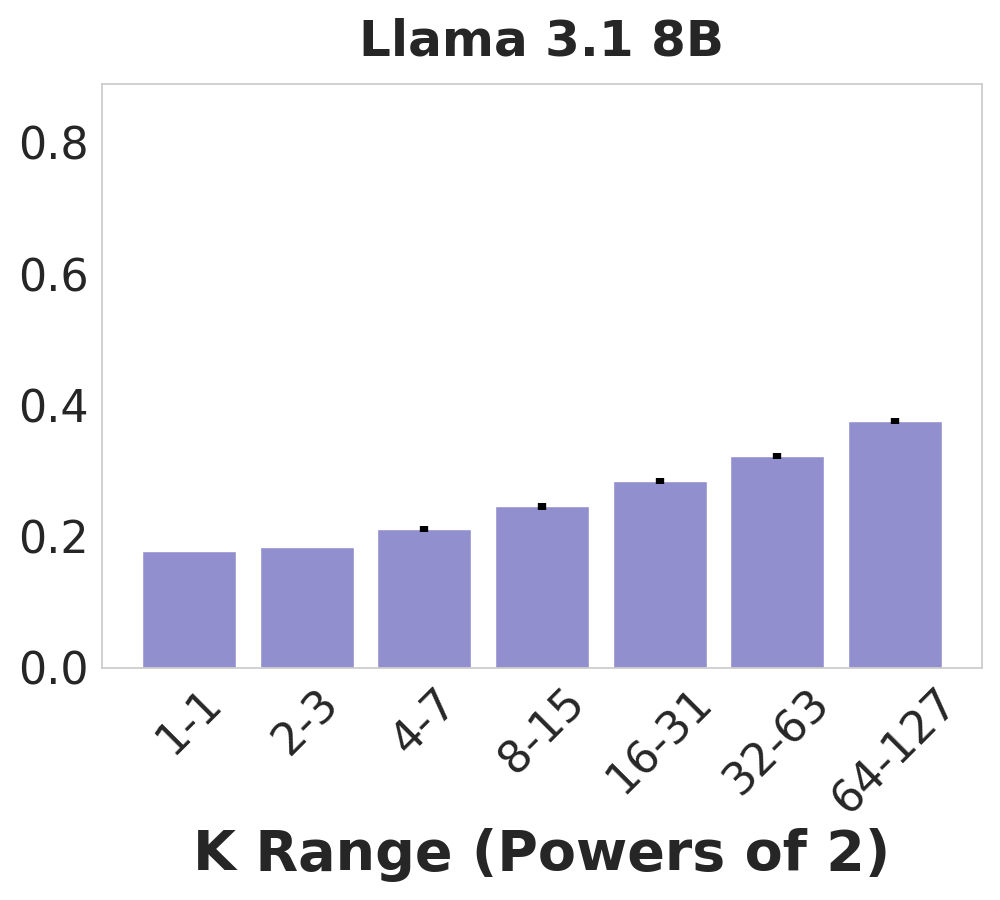}
\end{minipage}
\begin{minipage}[b]{0.24\textwidth}
\centering
\includegraphics[height=3.2cm]{fpr/fpr_code_llama_7b.png}
\end{minipage}
\begin{minipage}[b]{0.24\textwidth}
\centering
\includegraphics[height=3.2cm]{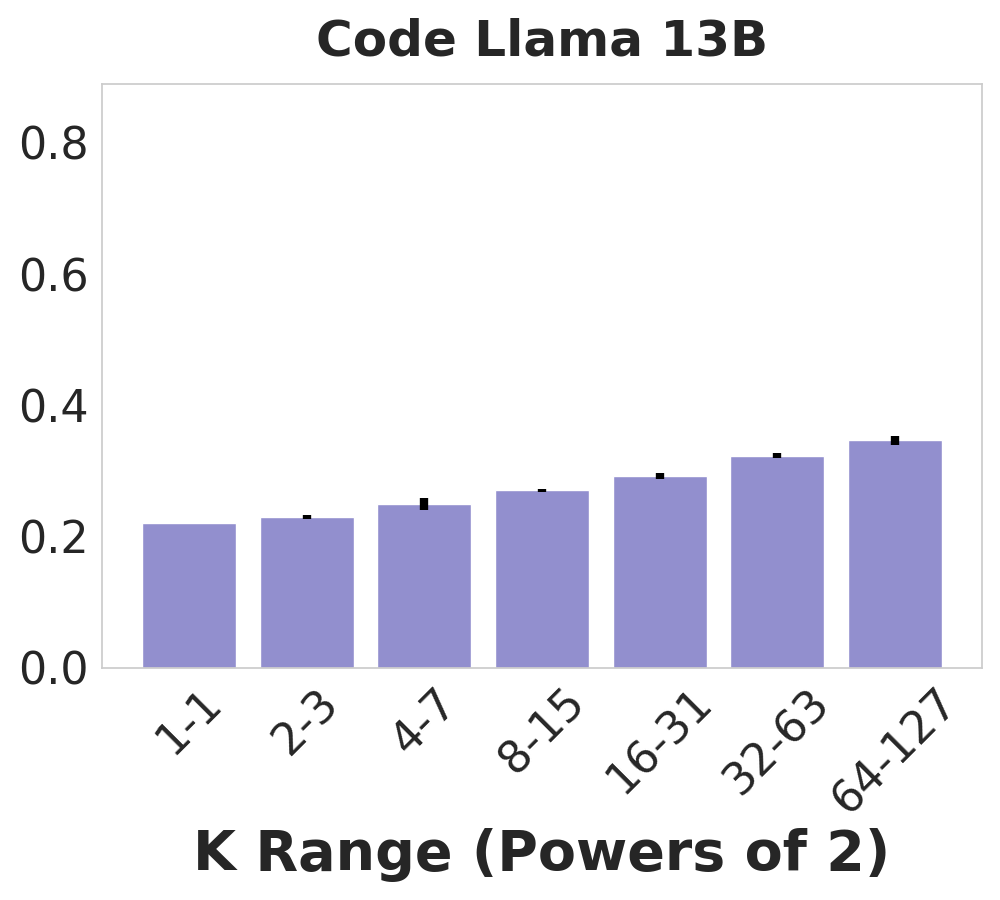}
\end{minipage}
\begin{minipage}[b]{0.24\textwidth}
\centering
\includegraphics[height=3.2cm]{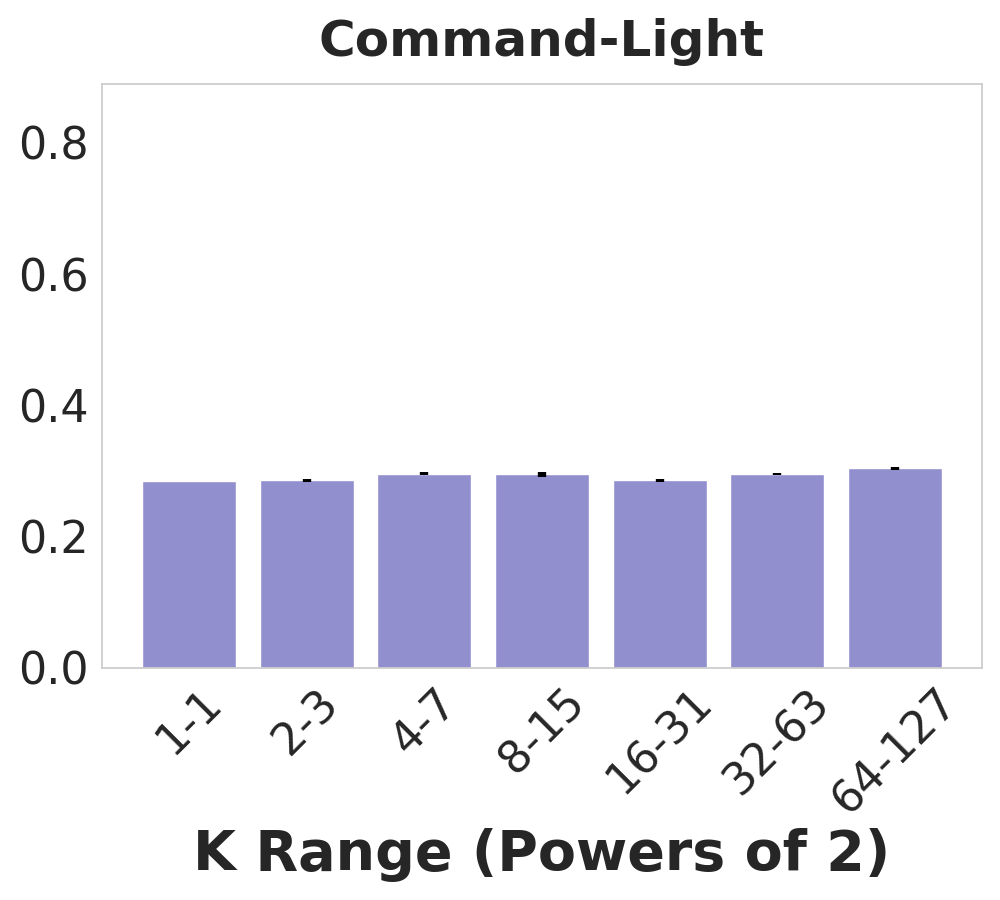}
\end{minipage}
\begin{minipage}[b]{0.24\textwidth}
\centering
\includegraphics[height=3.2cm]{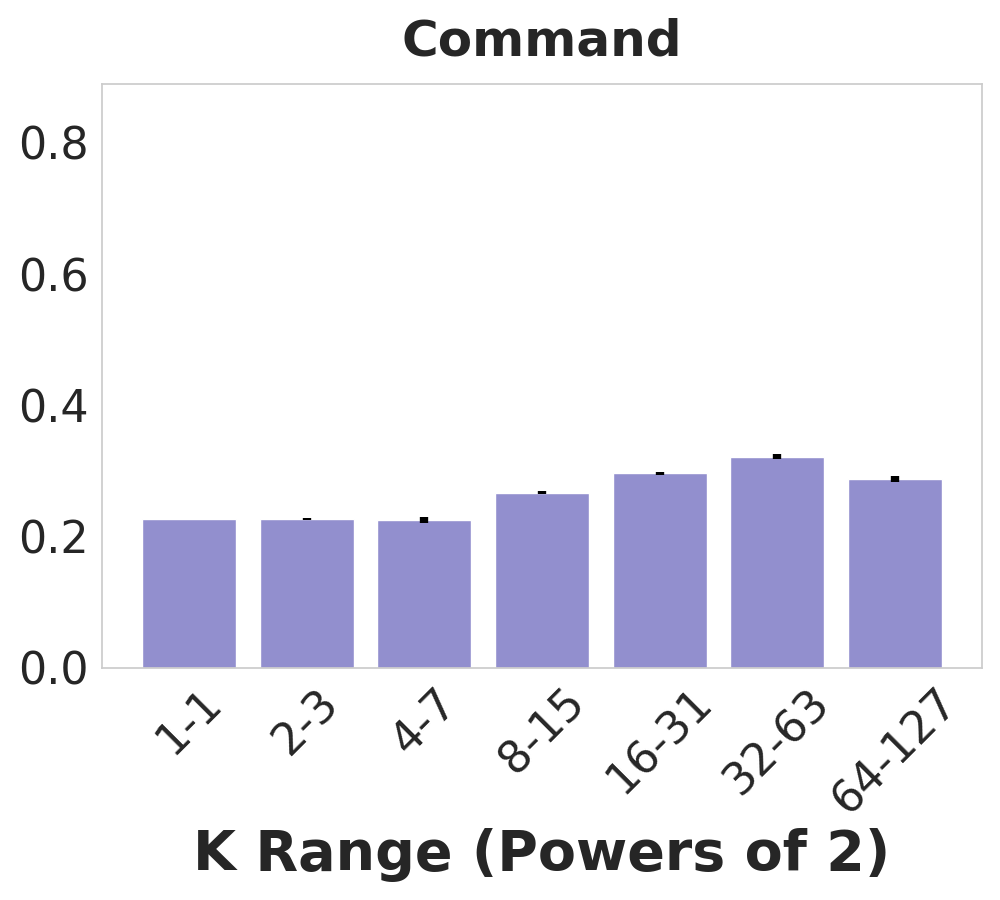}
\end{minipage}
\begin{minipage}[b]{0.24\textwidth}
\centering
\includegraphics[height=3.2cm]{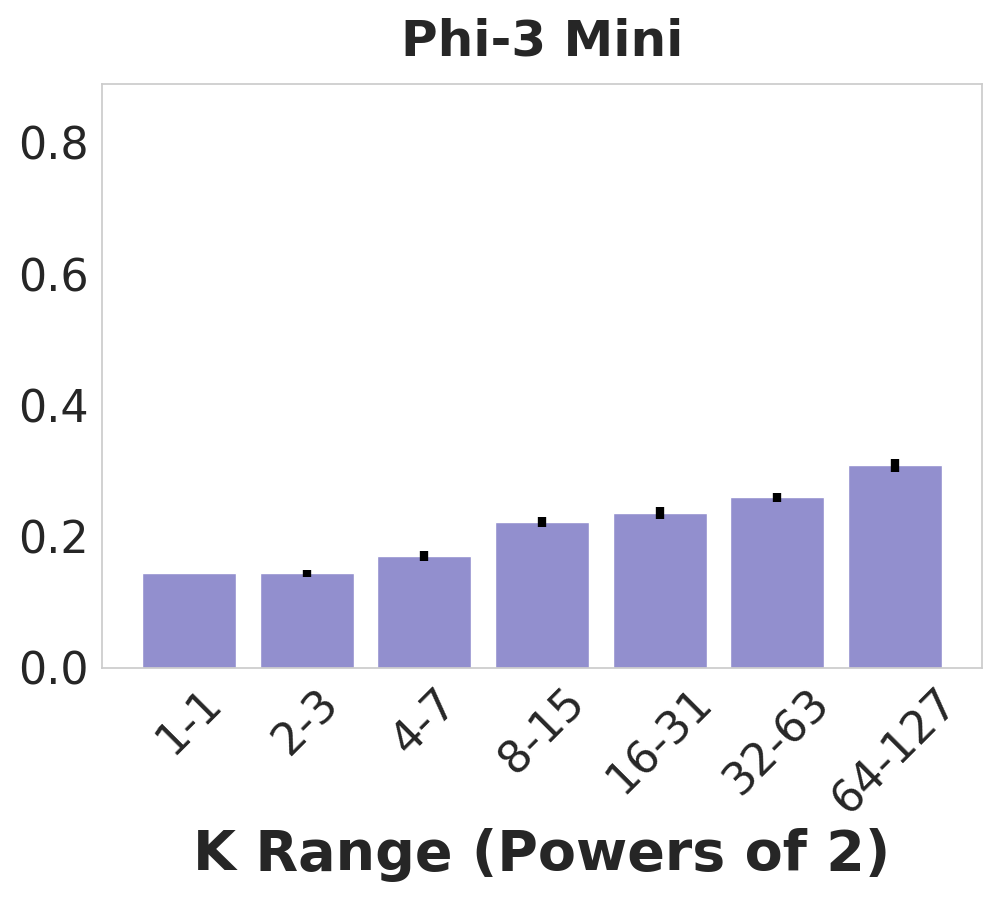}
\end{minipage}
\begin{minipage}[b]{0.24\textwidth}
\centering
\includegraphics[height=3.2cm]{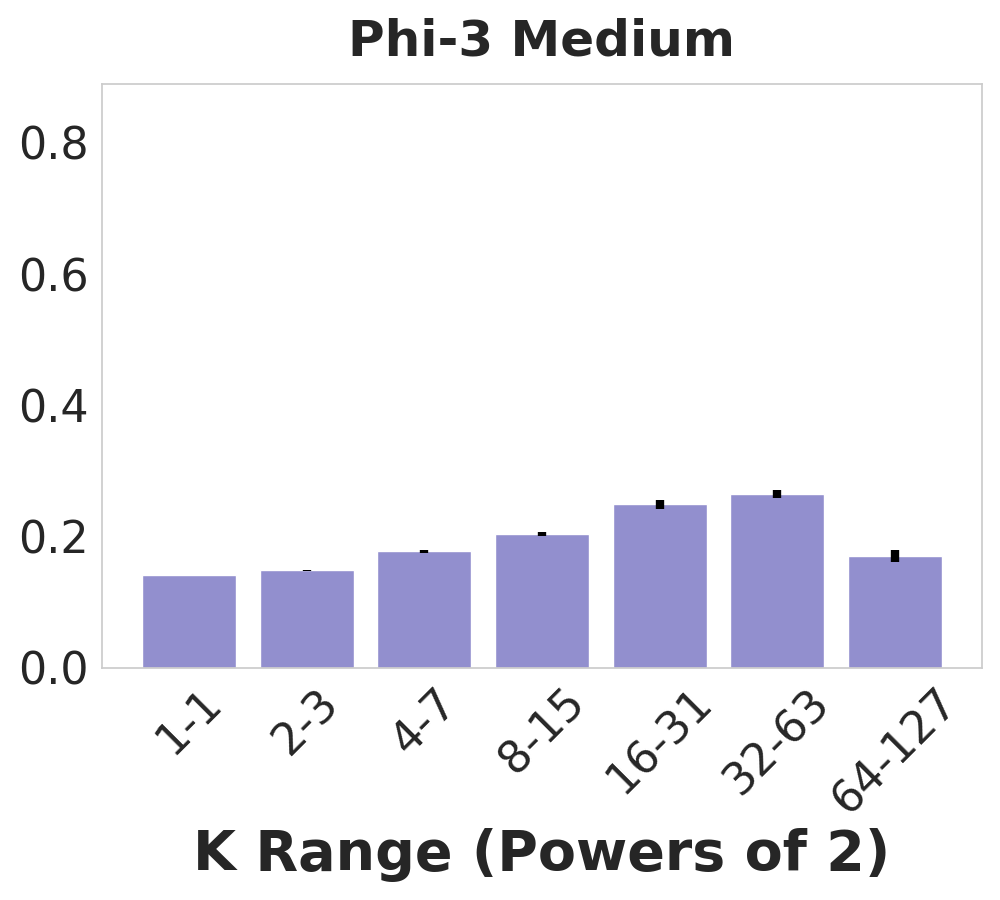}
\end{minipage}
\begin{minipage}[b]{0.24\textwidth}
\centering
\includegraphics[height=3.2cm]{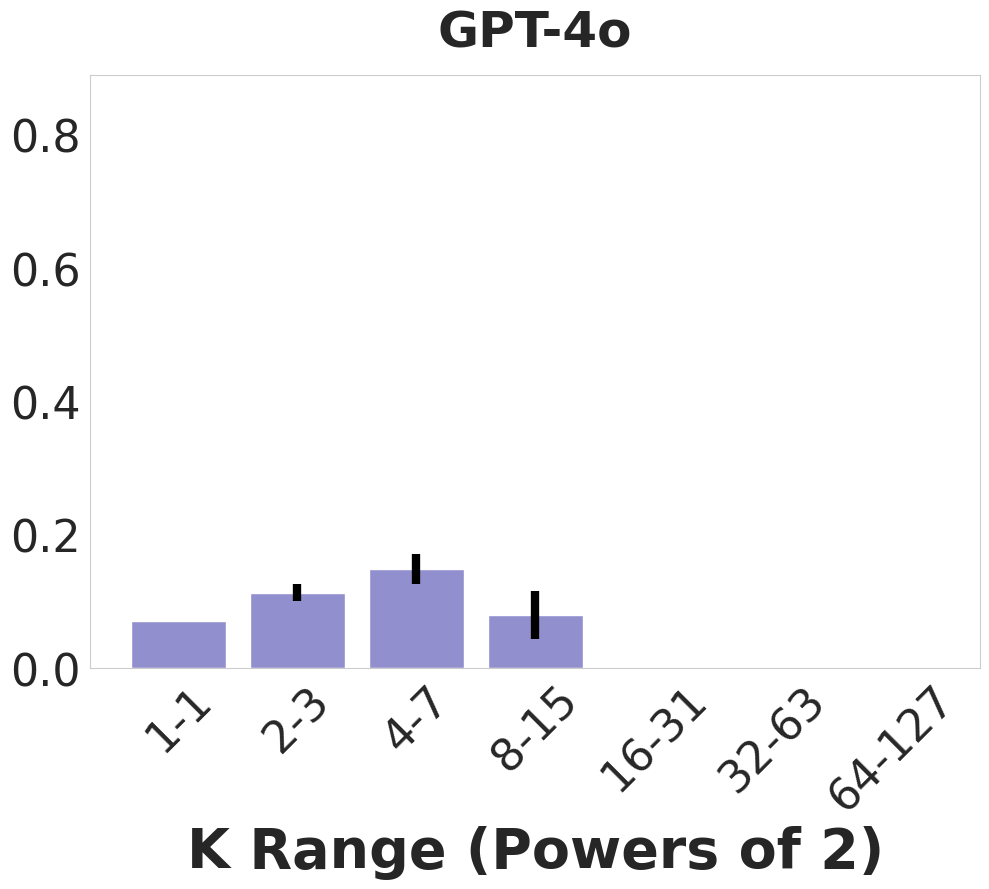}
\end{minipage}
\caption{\textbf{False positive rates as a function of the number of attempts $K$ in HumanEval+.} $K$ ranges are aggregated into bins of increasing size (i.e., powers of 2).}
\label{app:false_positive_rates}
\end{figure}

\begin{figure*}[h]
\centering
\begin{minipage}[b]{0.48\columnwidth}
\centering
\includegraphics[height=4.8cm]{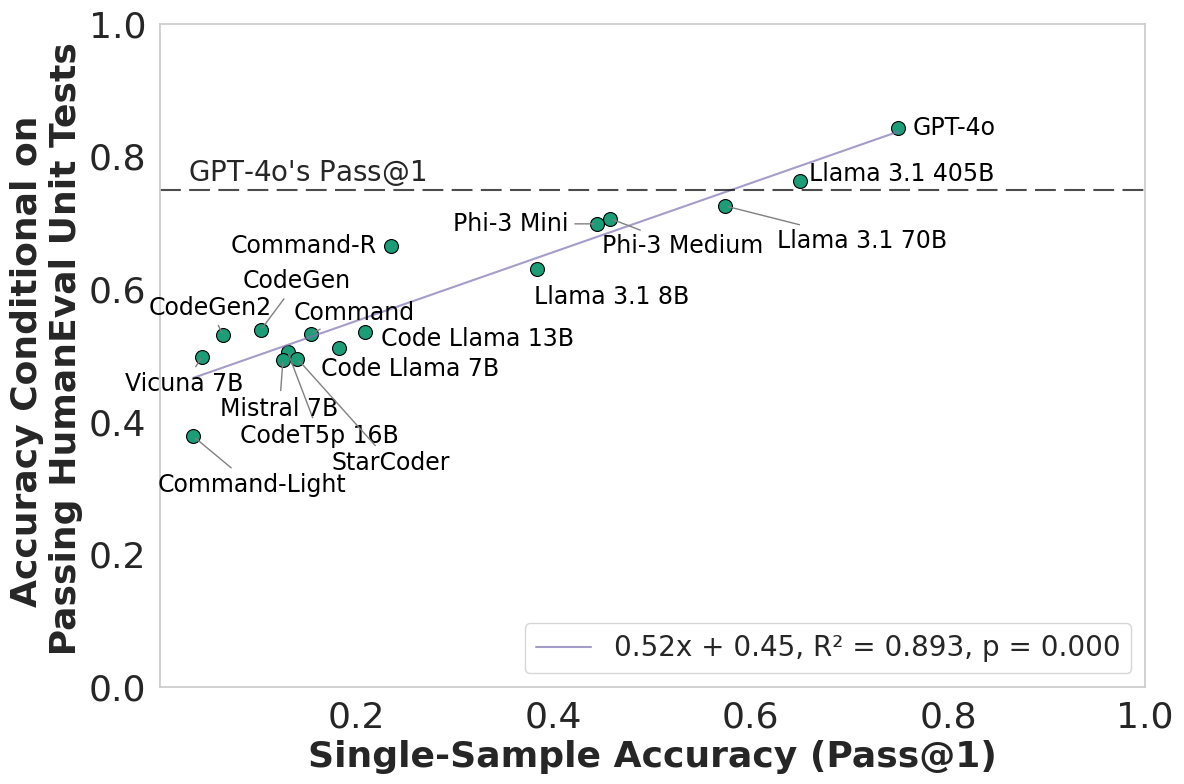}
\caption*{HumanEval+}
\end{minipage}
\hfill
\begin{minipage}[b]{0.48\columnwidth}
\centering
\includegraphics[height=4.8cm]{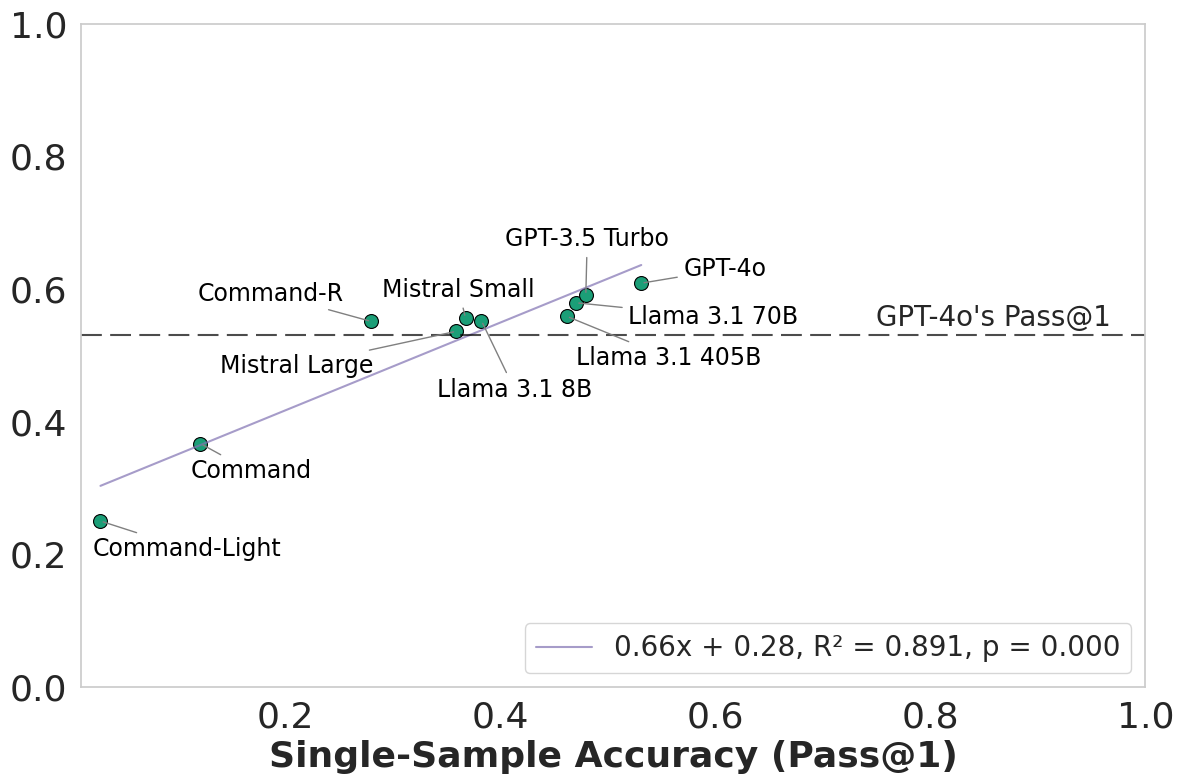}
\caption*{MBPP+}
\end{minipage}
\caption{\textbf{Relationship between the conditional accuracy after passing the standard unit tests and single-sample accuracy for tasks on which the unit tests (i.e. the "verifier") have a precision of less than 90\%.} For both benchmarks, we find a more pronounced relationship between capability and the probability of a false positive than when considering all tasks. Note that, the number of considered tasks with 70/150 and 128/321 is substantial. This plot shows the full y-axis.}
\label{app:90_percent}
\end{figure*}

\begin{figure*}[ht]
\centering
\begin{minipage}[b]{0.48\textwidth}
\centering
\includegraphics[height=4.8cm]{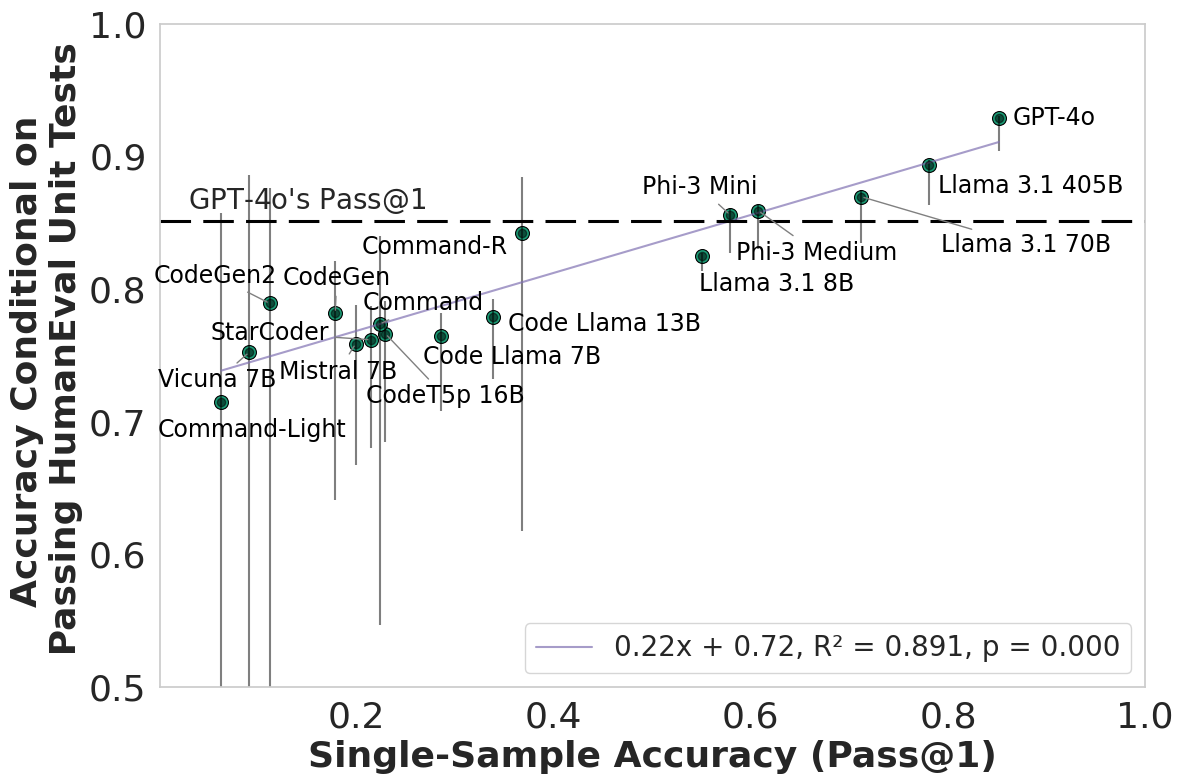}
\caption*{HumanEval+}
\end{minipage}
\hfill
\begin{minipage}[b]{0.48\textwidth}
\centering
\includegraphics[height=4.8cm]{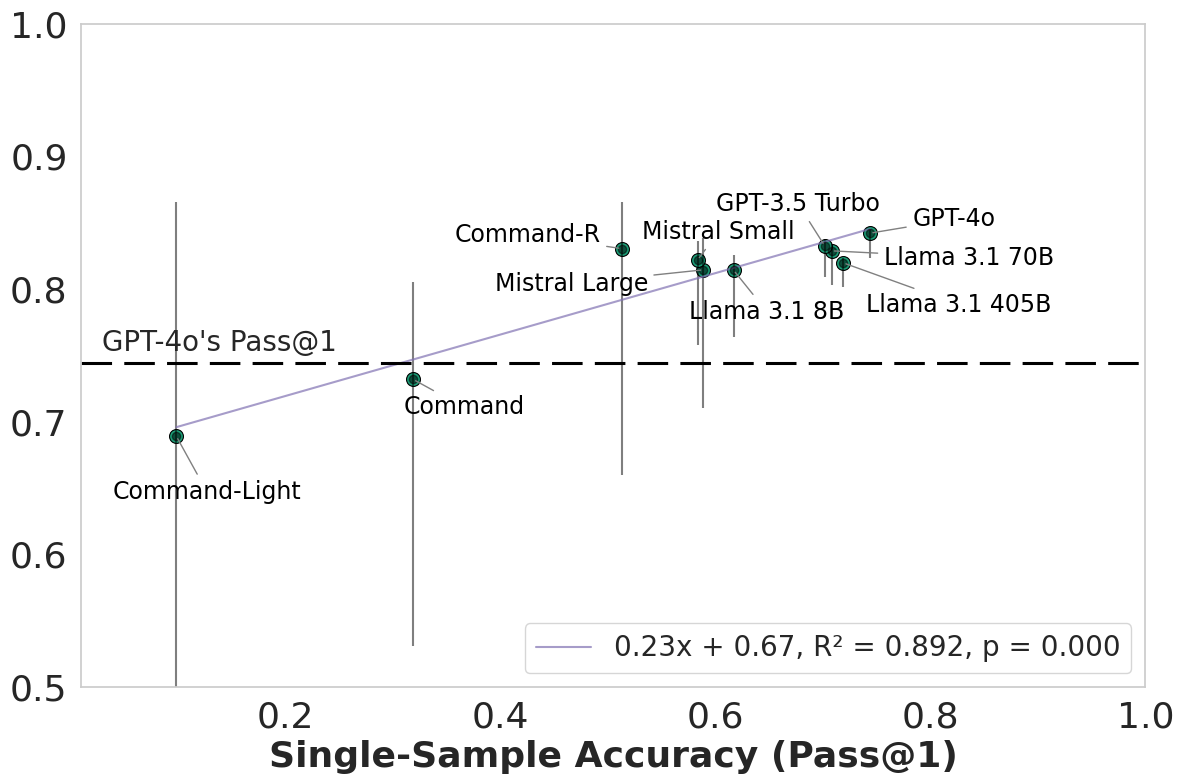}
\caption*{MBPP+}
\end{minipage}
\caption{\textbf{Worst-case upper and lower bounds for our HumanEval+ and MBPP+ analysis.} We show the relationship between the accuracy of individual samples (x-axis) and the achievable accuracy given an infinite compute budget and limited unit tests (y-axis; note that it starts at 0.5). Error bars indicate the bounds on the conditional accuracy when we account for tasks for which we did not observe \textit{any} passing solutions. For the upper (lower) bound, we set the conditional passing rate to 1 (0) when computing the accuracy estimates. Note that for Command Light, even after collecting 1000 samples for each task on HumanEval+, we still observed a substantial fraction of the tasks without a single passing solution. 
}
\label{app:scatter-errbars}
\end{figure*}

\subsection{Details on data and implementation} 
\label{app:dataandimplementation}

In the following, we provide more details on our analysis on HumanEval+ and MBPP+.

\xhdr{Sample Collection} To evaluate the generalization gap between weaker and stronger models, we collected multiple samples per model and task. For both benchmarks and each model, we used samples generated with a temperature setting of 0.8. For sample generation, we use the implementation provided by \citet{liu_is_2023}, and other than the temperature use their default settings.\footnote{See: \url{https://github.com/evalplus/evalplus/tree/937c46858cf8e687b31b5a728b7083d6e5a84971}} We collected a minimum of 50 samples for each model and task. For most models in our experiments Vicuna 7B, Mistral 7B, CodeT5p 16B, CodeGen, CodeGen2, Code Llama 7B, and Code Llama 13B, we used samples made available by \citet{liu_is_2023}. These were collected using the same temperature setting (i.e., 0.8). We had access to 200 samples per model and task for these models. Additionally, on HumanEval+, we collected 200 samples for Llama 3.1, Phi-3, GPT-4o, and the Command family of models. For Command Light, we even collected 1000 samples for each task to reduce the number of tasks without \textit{any} solutions passing the HumanEval unit tests (\cref{app:scatter-errbars}). On MBPP+, we collected 50 samples for each model and task. 

\subsection{Additional details on excluded tasks from MBPP+}
\label{app:excludedtasks-mbpp}

\xhdr{Tasks excluded by original EvalPlus authors (21 tasks)} These exclusions are based on an update to MBPP+, during which the authors removed several broken tasks, reducing the total to 378 tasks. These tasks were excluded because of issues with the oracle implementation leading to unreliable evaluations \footnote{See: \url{https://github.com/evalplus/evalplus/releases/tag/v0.3.1}}.

In addition to the task excluded by the MBPP+ creators, in our evaluations, we excluded a total of 57 tasks from the benchmark for two main reasons:

\xhdr{Tasks excluded due to additional oracle issues (28 tasks)} We identified and excluded an additional 28 tasks where \textit{all} generated solutions that passed the base tests failed the extended test suite and across \textit{all} models. We used this strict criterion to ensure we would not count solutions as false positives that are in fact robust but fail the extended test suite due to bugs in the MBPP+ harness. The primary cause was an implementation issue in the MBPP+ oracle when handling large numerical inputs, where the \texttt{np.allclose()} function used for checking output equivalence would raise exceptions. After these exclusions, our final evaluation set consisted of 350 tasks from the MBPP+ benchmark.

\xhdr{Tasks excluded due to solutions passing plus tests but failing standard unit tests (29 tasks)} An additional 29 tasks were excluded for which the extended unit tests yielded passing solutions that failed the base tests provided with the original benchmark. We intend to report these tasks to the benchmark creators, aiming to include these tasks in a future version of the paper once the issue is resolved. Notably, excluding these tasks did not significantly impact our final results.

\subsection{Additional details on excluded tasks from HumanEval+}
\label{app:excludedtasks-humaneval}

\xhdr{Tasks excluded due to solutions passing plus tests but failing standard unit tests (14 tasks)} Similar to MBPP+, we excluded 14 tasks from HumanEval+ from our analysis because passing solutions on the plus tests failed the standard unit tests. Including those tasks in the analysis did, as for MBPP+, not impact our results in any significant way. We plan to report these tasks to the benchmark creators.
\newpage
\section{Additional details on \cref{sec:theorysection}}
\label{app:theoretical-model}

In addition to the empirical analysis presented in \cref{sec:theorysection}, in this appendix, we provide a theoretical model that formalizes the limitations of inference scaling with imperfect verifiers and generalizes our findings to other benchmarks. We build on the verifier-based judge setup introduced by \citet{davis_networks_2024}. We provide a Python notebook with the implementation of our model\footnote{See supplementary materials, file \texttt{limits\_to\_inference\_scaling\_model.ipynb}}.

\begin{figure}[ht]
\begin{center}
\includegraphics[width=\textwidth]{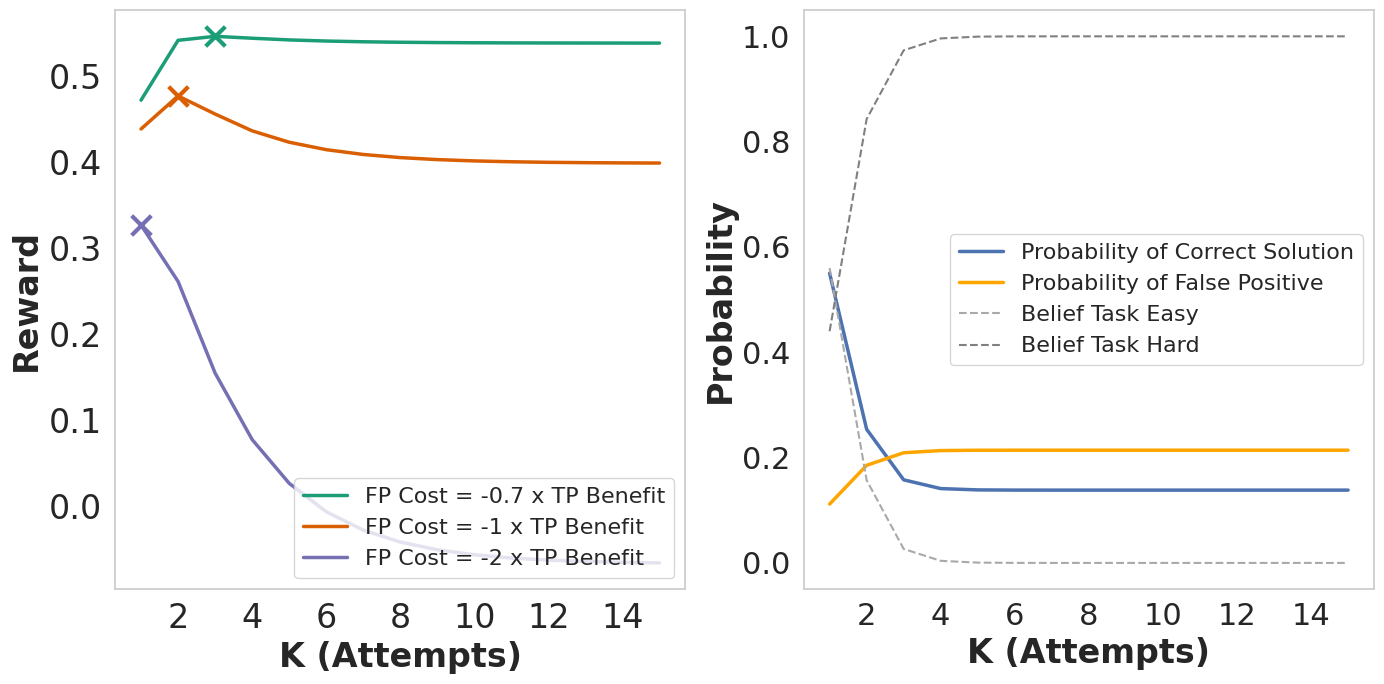}
\end{center}
\caption{\textbf{Even with \textit{zero computational cost}, the optimal number of samples is finite and very low ($K \leq 3$).} For this plot, we set the parameters as empirically observed for Llama 3.1 8B on HumanEval (see \cref{tab:parameter-values} for the exact values). The left plot shows the expected value of generating additional candidate solutions as a function of the number of attempts $K$ for various cost-benefit ratios. For all cost-benefit ratios, the expected value peaks at very low $K$, after which it begins to decline, indicating negative returns from additional sampling. The right depicts the probabilities of generating a correct solution vs. a false positive at each step $K$. As $K$ increases, the likelihood of generating a correct solution decreases, while the probability of generating a false positive increases. There is a trade-off between continued sampling and increasing risk, emphasizing the limitations of scaling inference compute with imperfect verifiers. Note that when setting the cost of a false positive to be 10 times higher than the benefit of a true positive, the optimal number of samples becomes $K = 0$ (\cref{fig:app-simulation}).}
\label{fig:simulation}
\end{figure}
\newpage

\begin{figure}[H]
\begin{center}
\includegraphics[width=\textwidth]{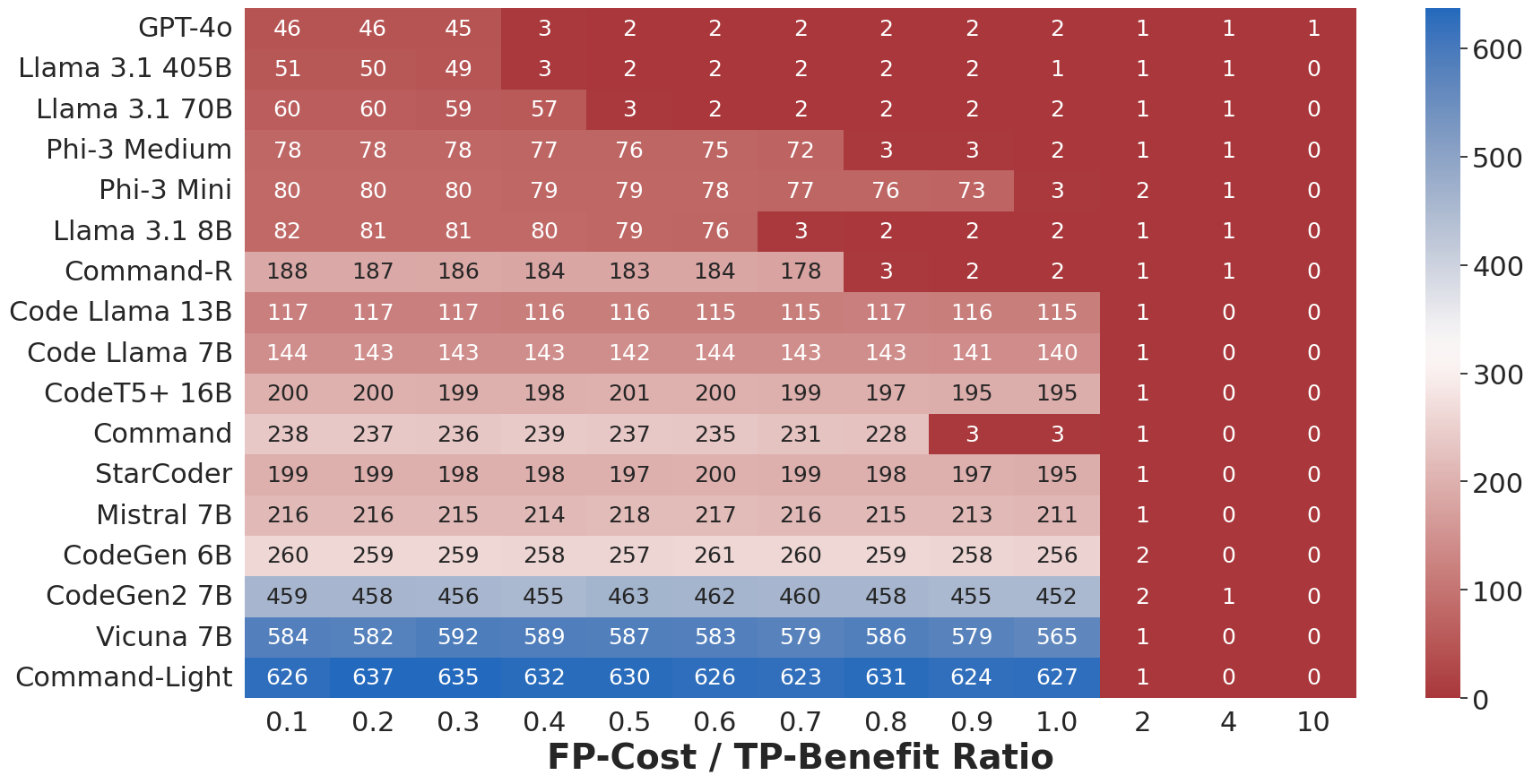}
\end{center}
\caption{Heatmap of optimal number of samples $K$ for various false positive cost vs. true positive benefit ratios in our model with parameters set as observed on HumanEval+. The y-axis shows models sorted by their Pass@1 accuracy. We observe that for a relative cost of 10 (i.e., the cost of returning a false positive is 10 times more costly than the reward of returning a true positive), the optimal number of samples is $K = 0$ for almost all models, effectively making them useless.}
\label{fig:app-simulation}
\end{figure}

\subsection{Model setup}

The underlying model consists of two components:
\begin{itemize}
    \item \textbf{Generator:} Produces candidate solutions to a task, with different success probabilities based on task difficulty.
    \begin{itemize}
        \item Tasks are either \textit{easy} ($T_1$) or \textit{hard} ($T_2$), with prior probabilities $p_1$ and $p_2$ respectively.
        \item The probabilities of generating a correct solution are $r_1$ for easy tasks and $r_2$ for hard tasks, so $r_1 > r_2$.
    \end{itemize}
    \item \textbf{Verifier:} An imperfect verifier checks the correctness of generated solutions.
    \begin{itemize}
        \item \textit{Completeness} ($c$): Conditional probability of accepting a correct solution.
        \item \textit{Soundness} ($s$): Conditional probability of rejecting an incorrect solution.
    \end{itemize}
\end{itemize}

\subsection*{Parameter values used in model underlying \cref{fig:simulation}}

\begin{table}[ht]
\centering
\begin{tabular}{l|c}
\toprule
\textbf{Parameter} & \textbf{Value} \\
\midrule
Probability of correct solution (easy task), $r_1$ & 0.87 \\
Probability of correct solution (hard task), $r_2$ & 0.13 \\
Completeness, $c$ & 1 \\
Soundness, $s$ & 0.75 \\
Prior probability of easy task, $p_1$ & 0.58 \\
Prior probability of hard task, $p_2$ & 0.42 \\
Benefit for correct solution (true positive), $V_{\text{TP}}$ & 1 \\
Cost for false positive, $V_{\text{FP}}$ & [-0.7, -1, -2] \\
Computational cost per attempt, $C_k$ & 0 \\
\bottomrule
\end{tabular}
\caption{Parameter values used in the model setup in \cref{fig:simulation} as observed for the Llama 3.1 8B model evaluated on HumanEval+. These values reflect the empirically observed probabilities and prior settings. Following the observed empirical task difficulty distribution as shown in \cref{fig:diff_dist}, in this setup we assume tasks with Pass@1 $\geq 0.5$ to be easy, and those with Pass@1 $< 0.5$ to be hard.}
\label{tab:parameter-values}
\end{table}

\subsection*{Probability of rejection}

The probability that a sample is being rejected by the verifier, denoted $\beta_i$, is given by:

\begin{equation}
    \beta_i = (1 - c)r_i + s(1 - r_i)
\end{equation}

where $i = 1$ for easy tasks and $i = 2$ for hard tasks. These probabilities ($\beta_1$ and $\beta_2$) determine how likely a generated solution is to be rejected depending on the task type.

\subsection*{Belief updates}

After each rejection, the belief that the task is of type $T_2$ (hard) increases. The posterior probability that the task is of type $T_1$ or $T_2$ after $k-1$ rejections is:

\begin{equation}
    p_{T_i}^{(k)} = \frac{\beta_i^{k-1} p_i}{\beta_1^{k-1} \cdot p_1 + \beta_2^{k-1} \cdot p_2}
\end{equation}

As more rejections occur, it usually becomes more likely that the task is hard ($T_2$). In \cref{fig:simulation}, we see how the belief that the task is easy decreases, while the belief that the task is hard increases as the number of attempts $K$ grows.

\subsection*{Probability of correct and false positive solutions}

For the $k$-th attempt, the probability of generating a correct solution or a false positive depends on the task type. The overall probabilities are weighted by the posterior beliefs \( p_{T_i}^{(k)} \).

The belief-weighted probability of returning a correct or false positive at attempt $k$, conditional on the $k-1$ previous attempts being rejected are:

\begin{equation}
    P_{\text{TP}}^{(k)} = p_{T_1}^{(k)} \cdot P_{\text{TP},T_1} + p_{T_2}^{(k)} \cdot P_{\text{TP},T_2}
\end{equation}

\begin{equation}
    P_{\text{FP}}^{(k)} = p_{T_1}^{(k)} \cdot P_{\text{FP},T_1} + p_{T_2}^{(k)} \cdot P_{\text{FP},T_2}
\end{equation}

where:
\[
P_{\text{TP},T_1} = c \cdot r_1, \quad P_{\text{TP},T_2} = c \cdot r_2
\]
\[
P_{\text{FP},T_1} = (1 - r_1) \cdot (1 - s), \quad P_{\text{FP},T_2} = (1 - r_2) \cdot (1 - s)
\]

In \cref{fig:simulation}, the right plot shows the evolution of \( P_{\text{TP}}^{(k)} \) and \( P_{\text{FP}}^{(k)} \) as the number of attempts $K$ increases. Initially, the probability of generating a correct solution is higher, but for higher $K$, the probability of generating a false positive increases.

\subsection*{Expected value of generating additional solutions}

The expected value of generating a solution at the $k$-th attempt is:

\begin{equation}
    \text{EV}_k = \left[ V_{\text{TP}} \cdot  P_{\text{TP}}^{(k)} + V_{\text{FP}} \cdot P_{\text{FP}}^{(k)}  \right] \cdot \left[ \beta_1^{k-1} \cdot p_{T_1}^{(k)} + \beta_2^{k-1} \cdot p_{T_2}^{(k)} \right] 
\end{equation}

where:
\begin{itemize}
    \item $V_{\text{TP}}$ is the benefit for a correct solution.
    \item $V_{\text{FP}}$ is the cost for a false positive being ``accepted'' as the solution.
\end{itemize}

\subsection*{Optimal number of attempts}

The total expected value after $K$ attempts is:

\begin{equation}
    \text{Reward} = \sum_{k=1}^{K} \text{EV}_k
\end{equation}

The optimal number of attempts, $K_{\text{opt}}$, is the value of $K$ that maximizes the reward, which are shown across models and for various $V_{FP}/V_{TP}$-ratios in \cref{fig:app-simulation}.

\subsection{Inference scaling curve for GPT-4o}

\begin{figure}[H]
    \centering
    \includegraphics[width=0.4\textwidth]{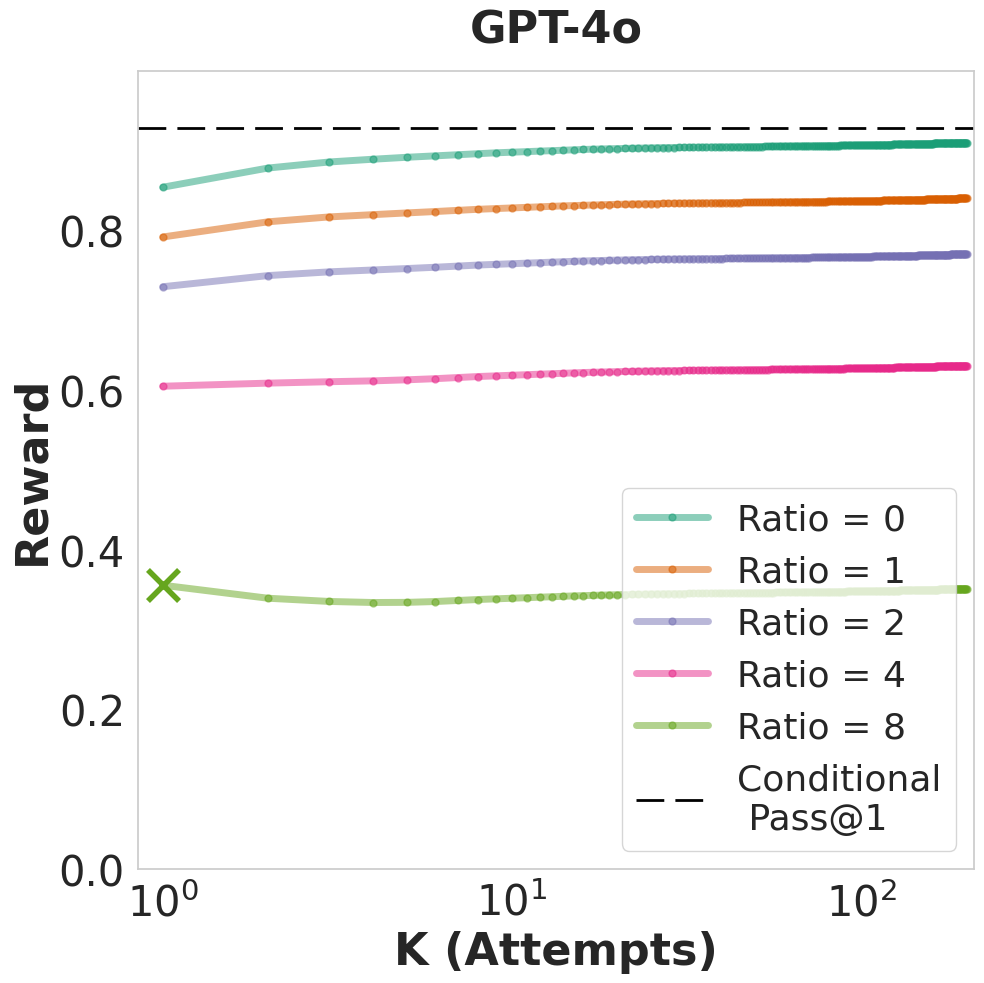}
    \caption{\textbf{Inference scaling curves in the presence of a cost for GPT-4o.} In addition to the models in \cref{fig:empirical_scaling_curves}, we provide the inference scaling curves for GPT-4o as the model with the highest single-sample accuracy on on HumanEval+ in our experiments. We find that the benefits of search are minimal (i.e. curves are flat) in line with what we expect from the empirical task difficulty distribution shown in \cref{fig:task_diff_humaneval}.}
    \label{app:scaling-curve-gpt4o}
\end{figure}

\newpage
\section{Additional details on \cref{sec:beyondcorrectness}}
\label{app:beyondcorrectness}

\begin{figure}[ht]
\begin{center}
\includegraphics[width=\textwidth]{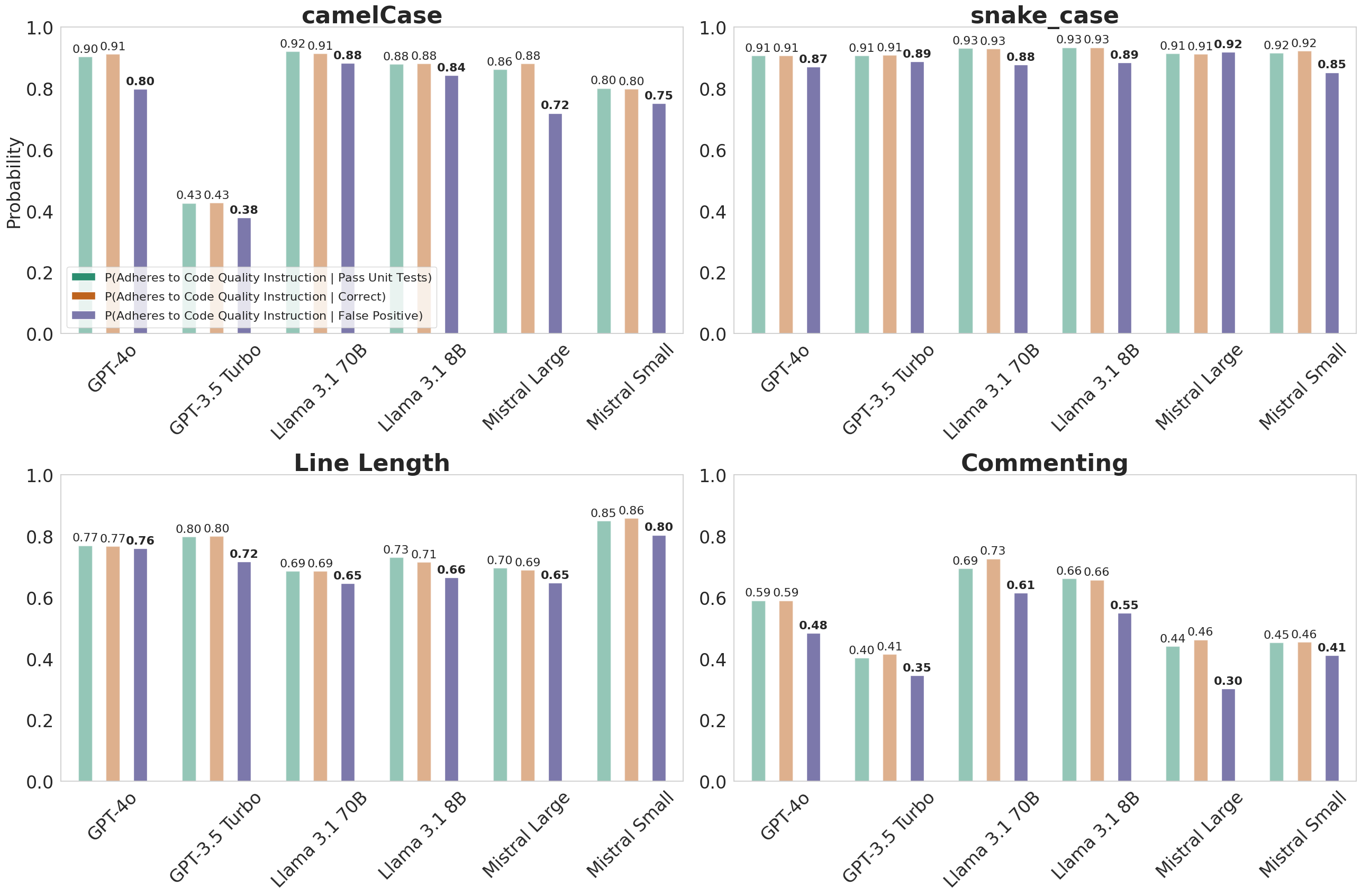}
\end{center}
\caption{\textbf{False positives tend to be lower-quality code compared to correct solutions across all models and code quality metrics.} We evaluated four key code quality metrics: adherence to \texttt{camelCase} and \texttt{snake\_case} naming conventions, line-length compliance, and presence of line-level comments. This trend holds consistently across models and for all four code quality instructions we test.}
\label{app:fig:codequality}
\end{figure}

\subsection{Details on data and implementation}

We used the implementation provided by \citet{zheng_beyond_2024} to collect samples and evaluate the different code readability metrics. Each code quality metric had a separate prompt instructing the model to follow certain guidelines (\cref{app:promptexamples}). For each model and code quality instruction, we generated 50 samples per task on HumanEval+. As for our main experiments, we set the temperature to 0.8. All other parameters were set to their default value as provided with the implementation.\footnote{See: \url{https://github.com/jszheng21/RACE/tree/3b8ee591abd5febd8ae8ec17c7b9907949c5e1d5}}

\subsection{Prompt examples for readability metrics}
\label{app:promptexamples}

\begin{figure}[H]
\centering
\begin{tcolorbox}[width=1\textwidth, fontupper=\small, colback=blue!2, boxrule=0.9pt]

\textbf{1) Naming conventions} \\

Please generate the Python code to solve the following problem, and use \texttt{camelCase} for both function names and variable names.\textbackslash n\textbackslash nProblem:\textbackslash n\textbackslash n\{problem\} \\

Please generate the Python code to solve the following problem, and use \texttt{snake\_case} for both function names and variable names.\textbackslash n\textbackslash nProblem:\textbackslash n\textbackslash n\{problem\} \\

\textbf{2) Code length} \\

Please generate the Python code to solve the following problem, where each line is less than 70 characters long and each function is less than 30 lines long.\textbackslash n\textbackslash nProblem:\textbackslash n\textbackslash n\{problem\} \\

\textbf{3) Commenting guidelines} \\

Please generate the Python code to solve the following problem, and add comments for each line in each function.\textbackslash n\textbackslash nProblem:\textbackslash n\textbackslash n\{problem\} \\
\end{tcolorbox}
\caption{Prompt templates for each metric of code readability we consider in our experiments for \cref{sec:beyondcorrectness} following \citet{zheng_beyond_2024}.}
\label{fig:instrutions_c_r}
\end{figure}

\subsection{Qualitative examples of false positives}
\label{app:beyondcorrectness_qualexamples}

\begin{figure}[h!]
  \centering
  \begin{minipage}{\columnwidth}
    \begin{lstlisting}[language=Python]
def get_positive(l: list):
    return [int(item) for item in l if item > 0]
    \end{lstlisting}
    \caption*{a) Example implementation generated by CodeGen-6B which fails additional unit tests from HumanEval+.}
    \vspace{0.2cm}
  \end{minipage}
    \begin{minipage}{\columnwidth}
    \begin{lstlisting}[language=Python]
# Standard unit tests
assert get_positive([5, 3, -5, 2, -3, 3, 9, 0, 123, 1, -10]) == [5, 3, 2, 3, 9, 123, 1]
assert get_positive([-1, 2, -4, 5, 6]) == [2, 5, 6]

# Example unit test from extended test suite
assert get_positive([0.5, 0, -4, 2.5, 5, -2.2, -8, 7.7, 9.9, -10.5]) == [0.5, 2.5, 5, 7.7, 9.9]
    \end{lstlisting}
    \caption*{b) Standard unit tests for the HumanEval/30 task and one example test from the extended test suite of HumanEval+.}
  \end{minipage}
  \label{fig:flukeexample}
  \caption{Example implementation that passes the standard tests but is incorrect. In HumanEval/30, the model is asked to \textit{Return only positive numbers in the list.} The code was generated by CodeGen-6B \citep{nijkamp_codegen_2023} and passes the tests provided with the original HumanEval benchmark but fails the extended tests. The implementation contains a logical error due to unnecessarily converting positive floats to integers.}
\end{figure}

\begin{figure}[H]
  \centering
  \begin{minipage}[t]{0.9\textwidth}
    \begin{lstlisting}[language=Python]
def string_sequence(n: int) -> str:
    ans = ""
    for i in range(0, n + 1):
        ans = ans + str(i) + " "
    return ans[:-1]
    \end{lstlisting}
  \end{minipage}
  \caption{HumanEval/15 asks to generate a space-delimited sequence of numbers. The implementation generated by CodeGen fails on the extended test suite due to the inefficient for-loop that times out on larger inputs.}
\end{figure}

\begin{figure}[H]
  \centering
  \begin{minipage}[t]{0.9\textwidth}
    \begin{lstlisting}[language=Python]
def sort_numbers(numbers: str) -> str:
    str_num_dict = {
        "zero": 0, "one": 1, "two": 2, "three": 3, "four": 4,
        "five": 5, "six": 6, "seven": 7, "eight": 8, "nine": 9,
    }
    num_list: List[int] = [str_num_dict[word] for word in numbers.split()]
    num_list.sort()
    sorted_str_nums: List[str] = [
        word for word, num in str_num_dict.items() if num in num_list
    ]
    return " ".join(sorted_str_nums)
    \end{lstlisting}
  \end{minipage}
  \caption{HumanEval/19 asks to sort space-delimited number words. The implementation generated by Code Llama 7B fails to maintain duplicates in the output, incorrectly removing repeated numbers during the dictionary lookup process.}
\end{figure}

\begin{figure}[H]
  \centering
  \begin{minipage}[t]{0.35\textwidth}
    \begin{lstlisting}[language=Python]
def find_char_long(string):    
  words = string.split()    
  return [word for word in words if len(word) >= 4]
    \end{lstlisting}
    \caption*{Example implementation generated by GPT-3.5 which fails additional unit tests from MBPP+.}
  \end{minipage}
  \hfill
  \begin{minipage}[t]{0.6\textwidth}
    \begin{lstlisting}[language=Python]
# Standard unit tests
assert set(find_char_long('Please move back to stream')) == set(['Please', 'move', 'back', 'stream'])
assert set(find_char_long('Jing Eco and Tech')) == set(['Jing', 'Tech'])
assert set(find_char_long('Jhingai wulu road Zone 3')) == set(['Jhingai', 'wulu', 'road', 'Zone'])

# Example unit tests from extended test suite
assert set(find_char_long("This is a very long string with no words that are at least 4 characters long.")) == set(['This', 'very', 'long', 'string', 'with', 'words', 'that', 'least', 'characters', 'long'])
    \end{lstlisting}
    \caption*{Standard unit tests for the MBPP/7 task.}
  \end{minipage}
  \caption{Exemplary implementation that passes the standard tests but is incorrect. In MBPP/7, the model is asked to \textit{write a function to find all words that are at least 4 characters long in a string.} The code was generated by gpt-3.5 and passes the tests provided with the original MBPP benchmark but is incorrect and fails to pass the extended tests.}
\end{figure}

\begin{figure}[H]
  \centering
  \begin{minipage}[t]{0.9\textwidth}
    \begin{lstlisting}[language=Python]
def parse_nested_parens(paren_string: str) -> List[int]:
    def parse_paren_group(paren_group: str) -> int:
        counter = 0
        max_counter = 0
        for paren in paren_group:
            if paren == '(':
                counter += 1
                if counter > max_counter:
                    max_counter = counter
            elif paren == ')':
                counter -= 1
        return max_counter
    return [parse_paren_group(paren_group) for paren_group in paren_string.split(' ')]
    \end{lstlisting}
  \end{minipage}
  \caption{HumanEval/6 requires parsing nested parentheses groups. The implementation generated by Code Llama 7B incorrectly processes empty input strings.}
\end{figure}

\begin{figure}[H]
  \centering
  \begin{minipage}[t]{0.9\textwidth}
    \begin{lstlisting}[language=Python]
def sorted_list_sum(lst):
    assert lst, "List can not be empty."
    if isinstance(lst[0], str):
        new_lst = [word for word in lst if len(word) % 2 == 0]
        new_lst.sort()
        return sorted(new_lst, key=len)
    else:
        return sorted(lst)
    \end{lstlisting}
  \end{minipage}
  \caption{HumanEval/149 asks to process a list of strings, removing odd-length strings and sorting the result. The implementation generated by CodeT5p 16B incorrectly handles empty input by raising an assertion error instead of returning an empty list.}
\end{figure}

\begin{figure}[H]
  \centering
    \begin{lstlisting}[language=Python]
def countNumbersWithOne(n):
    if n == 1:
        # For single-digit numbers, we only have 1
        return 1

    count = 0

    # Calculate the range for n-digit numbers
    start = 10**(n-1)
    end = 10**n
    
    for num in range(start, end):
        str_num = str(num)
        if str_num[0] == '1' or str_num[-1] == '1':
            count += 1

    return count
    \end{lstlisting}
    \caption{Example candidate implementation failing to adhere to \texttt{camelCase} naming convention for HumanEval/83 generated by GPT-4o due to variable \textit{str\_num} in line 13 not complying.}
\end{figure}

\begin{figure}[H]
  \centering
  \begin{minipage}[t]{0.9\textwidth}
    \begin{lstlisting}[language=Python]
def get_positive(l: list):
    return [int(item) for item in l if item > 0]
    \end{lstlisting}
  \end{minipage}
  \caption{HumanEval/30 asks to return positive numbers from a list. The implementation generated by CodeGen fails on the extended test suite because it converts float values to integers before returning the list.}
\end{figure}

\begin{figure}[H]
  \centering
  \begin{minipage}[t]{0.9\textwidth}
    \begin{lstlisting}[language=Python]
def number_of_substrings(s):
    char_count = [0] * 128
    result = 0
    for char in s:
        char_count[ord(char)] += 1
    for i in range(len(s)):
        result += char_count[ord(s[i])] * (len(s) - i)
    return result
    \end{lstlisting}
  \end{minipage}
  \caption{MBPP/238 asks to count non-empty substrings in a string. The implementation generated by Command uses an incorrect algorithm that produces wrong results.}
\end{figure}

\begin{figure}[H]
  \centering
  \begin{minipage}[t]{0.9\textwidth}
    \begin{lstlisting}[language=Python]
def split(string):
    return [char for char in string if char.isalpha()]
    \end{lstlisting}
  \end{minipage}
  \caption{MBPP/565 asks to split a string into characters. The implementation generated by Command Light incorrectly filters for alphabetic characters only and fails to handle empty strings correctly.}
\end{figure}

\begin{figure}[H]
  \centering
  \begin{minipage}[t]{0.9\textwidth}
    \begin{lstlisting}[language=Python]
def is_samepatterns(words, patterns):
    if len(words) != len(patterns):
        return False
    word_to_pattern = {}
    pattern_to_word = {}
    for word, pattern in zip(words, patterns):
        if word in word_to_pattern and word_to_pattern[word] != pattern:
            return False
        if pattern in pattern_to_word and pattern_to_word[pattern] != word:
            return False
        word_to_pattern[word] = pattern
        pattern_to_word[pattern] = word
    return True
    \end{lstlisting}
  \end{minipage}
  \caption{MBPP/74 asks to check if words follow given patterns. The implementation generated by GPT-3.5 fails on the extended test suite when handling repeated elements, incorrectly returning False.}
\end{figure}

\begin{figure}[H]
  \centering
  \begin{minipage}[t]{0.9\textwidth}
    \begin{lstlisting}[language=Python]
def find_even_pair(nums):
    even_count = 0
    xor_pairs = set()
    for i in range(len(nums)):
        for j in range(i+1, len(nums)):
            if (nums[i] ^ nums[j]) % 2 == 0:
                pair = tuple(sorted((nums[i], nums[j])))
                xor_pairs.add(pair)
    return len(xor_pairs)
    \end{lstlisting}
  \end{minipage}
  \caption{MBPP/166 asks to count pairs of integers with even XOR. The implementation generated by Llama 3.1 8B incorrectly counts only unique pairs using a set.}
\end{figure}

\begin{figure}[H]
  \centering
  \begin{minipage}[t]{0.9\textwidth}
    \begin{lstlisting}[language=Python]
def diff_even_odd(numbers):
    even = None
    odd = None
    for num in numbers:
        if num % 2 == 0 and even is None:
            even = num
        elif num % 2 == 1 and odd is None:
            odd = num
        if even is not None and odd is not None:
            break
    return abs(even - odd)
    \end{lstlisting}
  \end{minipage}
  \caption{MBPP/594 asks to write a function to find the difference between the first even and the first odd number of a given list. The implementation generated by GPT-3.5 incorrectly returns the \textit{absolute} value of the difference.}
\end{figure}

\section{Declaration of LLM usage and compute resources}
\label{llm_compute_declaration}

In this work, LLMs are the main subject of study and used for sampling solutions for coding benchmarks. For all our experiments using OpenAI models, we utilized the endpoints provided by OpenAI, either directly or through the Azure OpenAI Service. For the analysis on HotPotQA using Llama models, we relied on the endpoints provided by Together.ai. As our work primarily relied on external APIs, we did not use any GPUs for inference and our experiments did not require training of LLMs.

\section{Impact Statement}
\label{impact}

Our work aims to enhance the technical understanding of the limitations of inference scaling methods. In particular, we contribute new findings on resampling in combination with imperfect verifiers and how this approach can fail to lift the accuracy of weaker models to match the performance of stronger models. While these findings have implications for the training of reasoning models and the deployment of compound AI systems—including code generation tools—their broader ethical and societal implications mirror those already familiar in the development and use of large-scale language models. We do not identify any additional, domain-specific concerns that arise uniquely from our study. Instead, our results reinforce the importance of reliable evaluation metrics and thorough verification methods, which in turn support safer and more trustworthy applications in coding as well as in other areas where compound AI systems are increasingly adopted.

\end{document}

%% file: math_commands.tex
\usepackage{amsmath,amsfonts,bm}

\def\eqref#1{equation~\ref{#1}}

\def\1{\bm{1}}

\DeclareMathAlphabet{\mathsfit}{\encodingdefault}{\sfdefault}{m}{sl}
\SetMathAlphabet{\mathsfit}{bold}{\encodingdefault}{\sfdefault}{bx}{n}



%% file: add-ref.bib
@misc{sky_t1_2025,
  author       = {{NovaSky Team}},
  title        = {Sky-T1: Train your own O1 preview model within $450},
  howpublished = {https://novasky-ai.github.io/posts/sky-t1},
  note         = {Accessed: 2025-01-09},
  year         = {2025}
}

@misc{bespoke_stratos,  
    author = {{Bespoke Labs}},  
    title = {Bespoke-Stratos: The unreasonable effectiveness of reasoning distillation},  
    howpublished = {www.bespokelabs.ai/blog/bespoke-stratos-the-unreasonable-effectiveness-of-reasoning-distillation},  
    note = {Accessed: 2025-01-22},  
    year = {2025}
}

@misc{gao2022scalinglawsrewardmodel,
      title={Scaling Laws for Reward Model Overoptimization}, 
      author={Leo Gao and John Schulman and Jacob Hilton},
      year={2022},
      eprint={2210.10760},
      archivePrefix={arXiv},
      primaryClass={cs.LG},
      url={https://arxiv.org/abs/2210.10760}, 
}


%% file: iclr2026/iclr2026_conference.bib
@misc{deepseek-ai_deepseek-r1_2025,
	title = {{DeepSeek}-{R1}: {Incentivizing} {Reasoning} {Capability} in {LLMs} via {Reinforcement} {Learning}},
	shorttitle = {{DeepSeek}-{R1}},
	url = {http://arxiv.org/abs/2501.12948},
	doi = {10.48550/arXiv.2501.12948},
	urldate = {2025-01-29},
	publisher = {arXiv},
	author = {DeepSeek-AI and Guo, Daya and Yang, Dejian and Zhang, Haowei and Song, Junxiao and Zhang, Ruoyu and Xu, Runxin and Zhu, Qihao and Ma, Shirong and Wang, Peiyi and Bi, Xiao and Zhang, Xiaokang and Yu, Xingkai and Wu, Yu and Wu, Z. F. and Gou, Zhibin and Shao, Zhihong and Li, Zhuoshu and Gao, Ziyi and Liu, Aixin and Xue, Bing and Wang, Bingxuan and Wu, Bochao and Feng, Bei and Lu, Chengda and Zhao, Chenggang and Deng, Chengqi and Zhang, Chenyu and Ruan, Chong and Dai, Damai and Chen, Deli and Ji, Dongjie and Li, Erhang and Lin, Fangyun and Dai, Fucong and Luo, Fuli and Hao, Guangbo and Chen, Guanting and Li, Guowei and Zhang, H. and Bao, Han and Xu, Hanwei and Wang, Haocheng and Ding, Honghui and Xin, Huajian and Gao, Huazuo and Qu, Hui and Li, Hui and Guo, Jianzhong and Li, Jiashi and Wang, Jiawei and Chen, Jingchang and Yuan, Jingyang and Qiu, Junjie and Li, Junlong and Cai, J. L. and Ni, Jiaqi and Liang, Jian and Chen, Jin and Dong, Kai and Hu, Kai and Gao, Kaige and Guan, Kang and Huang, Kexin and Yu, Kuai and Wang, Lean and Zhang, Lecong and Zhao, Liang and Wang, Litong and Zhang, Liyue and Xu, Lei and Xia, Leyi and Zhang, Mingchuan and Zhang, Minghua and Tang, Minghui and Li, Meng and Wang, Miaojun and Li, Mingming and Tian, Ning and Huang, Panpan and Zhang, Peng and Wang, Qiancheng and Chen, Qinyu and Du, Qiushi and Ge, Ruiqi and Zhang, Ruisong and Pan, Ruizhe and Wang, Runji and Chen, R. J. and Jin, R. L. and Chen, Ruyi and Lu, Shanghao and Zhou, Shangyan and Chen, Shanhuang and Ye, Shengfeng and Wang, Shiyu and Yu, Shuiping and Zhou, Shunfeng and Pan, Shuting and Li, S. S. and Zhou, Shuang and Wu, Shaoqing and Ye, Shengfeng and Yun, Tao and Pei, Tian and Sun, Tianyu and Wang, T. and Zeng, Wangding and Zhao, Wanjia and Liu, Wen and Liang, Wenfeng and Gao, Wenjun and Yu, Wenqin and Zhang, Wentao and Xiao, W. L. and An, Wei and Liu, Xiaodong and Wang, Xiaohan and Chen, Xiaokang and Nie, Xiaotao and Cheng, Xin and Liu, Xin and Xie, Xin and Liu, Xingchao and Yang, Xinyu and Li, Xinyuan and Su, Xuecheng and Lin, Xuheng and Li, X. Q. and Jin, Xiangyue and Shen, Xiaojin and Chen, Xiaosha and Sun, Xiaowen and Wang, Xiaoxiang and Song, Xinnan and Zhou, Xinyi and Wang, Xianzu and Shan, Xinxia and Li, Y. K. and Wang, Y. Q. and Wei, Y. X. and Zhang, Yang and Xu, Yanhong and Li, Yao and Zhao, Yao and Sun, Yaofeng and Wang, Yaohui and Yu, Yi and Zhang, Yichao and Shi, Yifan and Xiong, Yiliang and He, Ying and Piao, Yishi and Wang, Yisong and Tan, Yixuan and Ma, Yiyang and Liu, Yiyuan and Guo, Yongqiang and Ou, Yuan and Wang, Yuduan and Gong, Yue and Zou, Yuheng and He, Yujia and Xiong, Yunfan and Luo, Yuxiang and You, Yuxiang and Liu, Yuxuan and Zhou, Yuyang and Zhu, Y. X. and Xu, Yanhong and Huang, Yanping and Li, Yaohui and Zheng, Yi and Zhu, Yuchen and Ma, Yunxian and Tang, Ying and Zha, Yukun and Yan, Yuting and Ren, Z. Z. and Ren, Zehui and Sha, Zhangli and Fu, Zhe and Xu, Zhean and Xie, Zhenda and Zhang, Zhengyan and Hao, Zhewen and Ma, Zhicheng and Yan, Zhigang and Wu, Zhiyu and Gu, Zihui and Zhu, Zijia and Liu, Zijun and Li, Zilin and Xie, Ziwei and Song, Ziyang and Pan, Zizheng and Huang, Zhen and Xu, Zhipeng and Zhang, Zhongyu and Zhang, Zhen},
	month = jan,
	year = {2025},
	note = {arXiv:2501.12948 [cs]},
}

@misc{liu_tinygsm_2023,
	title = {{TinyGSM}: achieving {\textgreater}80\% on {GSM8k} with small language models},
	shorttitle = {{TinyGSM}},
	url = {http://arxiv.org/abs/2312.09241},
	doi = {10.48550/arXiv.2312.09241},
	urldate = {2024-10-10},
	publisher = {arXiv},
	author = {Liu, Bingbin and Bubeck, Sebastien and Eldan, Ronen and Kulkarni, Janardhan and Li, Yuanzhi and Nguyen, Anh and Ward, Rachel and Zhang, Yi},
	month = dec,
	year = {2023},
	note = {arXiv:2312.09241},
}

@misc{qin_large_2024,
	title = {Large {Language} {Models} are {Effective} {Text} {Rankers} with {Pairwise} {Ranking} {Prompting}},
	url = {http://arxiv.org/abs/2306.17563},
	doi = {10.48550/arXiv.2306.17563},
	urldate = {2024-10-10},
	publisher = {arXiv},
	author = {Qin, Zhen and Jagerman, Rolf and Hui, Kai and Zhuang, Honglei and Wu, Junru and Yan, Le and Shen, Jiaming and Liu, Tianqi and Liu, Jialu and Metzler, Donald and Wang, Xuanhui and Bendersky, Michael},
	month = mar,
	year = {2024},
	note = {arXiv:2306.17563},
}

@misc{song_good_2024,
	title = {The {Good}, {The} {Bad}, and {The} {Greedy}: {Evaluation} of {LLMs} {Should} {Not} {Ignore} {Non}-{Determinism}},
	shorttitle = {The {Good}, {The} {Bad}, and {The} {Greedy}},
	url = {http://arxiv.org/abs/2407.10457},
	doi = {10.48550/arXiv.2407.10457},
	urldate = {2024-10-09},
	publisher = {arXiv},
	author = {Song, Yifan and Wang, Guoyin and Li, Sujian and Lin, Bill Yuchen},
	month = jul,
	year = {2024},
	note = {arXiv:2407.10457},
}

@misc{kapoor_ai_2024,
	title = {{AI} {Agents} {That} {Matter}},
	url = {http://arxiv.org/abs/2407.01502},
	doi = {10.48550/arXiv.2407.01502},
	urldate = {2024-08-31},
	publisher = {arXiv},
	author = {Kapoor, Sayash and Stroebl, Benedikt and Siegel, Zachary S. and Nadgir, Nitya and Narayanan, Arvind},
	month = jul,
	year = {2024},
	note = {arXiv:2407.01502 [cs]},
}

@misc{hassid_larger_2024,
	title = {The {Larger} the {Better}? {Improved} {LLM} {Code}-{Generation} via {Budget} {Reallocation}},
	shorttitle = {The {Larger} the {Better}?},
	url = {http://arxiv.org/abs/2404.00725},
	doi = {10.48550/arXiv.2404.00725},
	urldate = {2024-08-31},
	publisher = {arXiv},
	author = {Hassid, Michael and Remez, Tal and Gehring, Jonas and Schwartz, Roy and Adi, Yossi},
	month = jul,
	year = {2024},
	note = {arXiv:2404.00725 [cs]},
}

@misc{brown_large_2024,
	title = {Large {Language} {Monkeys}: {Scaling} {Inference} {Compute} with {Repeated} {Sampling}},
	shorttitle = {Large {Language} {Monkeys}},
	url = {http://arxiv.org/abs/2407.21787},
	doi = {10.48550/arXiv.2407.21787},
	urldate = {2024-08-31},
	publisher = {arXiv},
	author = {Brown, Bradley and Juravsky, Jordan and Ehrlich, Ryan and Clark, Ronald and Le, Quoc V. and Ré, Christopher and Mirhoseini, Azalia},
	month = jul,
	year = {2024},
	note = {arXiv:2407.21787 [cs]},
}

@misc{snell_scaling_2024,
	title = {Scaling {LLM} {Test}-{Time} {Compute} {Optimally} can be {More} {Effective} than {Scaling} {Model} {Parameters}},
	url = {http://arxiv.org/abs/2408.03314},
	language = {en},
	urldate = {2024-08-31},
	publisher = {arXiv},
	author = {Snell, Charlie and Lee, Jaehoon and Xu, Kelvin and Kumar, Aviral},
	month = aug,
	year = {2024},
	note = {arXiv:2408.03314 [cs]},
}

@misc{welleck_decoding_2024,
	title = {From {Decoding} to {Meta}-{Generation}: {Inference}-time {Algorithms} for {Large} {Language} {Models}},
	shorttitle = {From {Decoding} to {Meta}-{Generation}},
	url = {http://arxiv.org/abs/2406.16838},
	doi = {10.48550/arXiv.2406.16838},
	urldate = {2024-06-28},
	publisher = {arXiv},
	author = {Welleck, Sean and Bertsch, Amanda and Finlayson, Matthew and Schoelkopf, Hailey and Xie, Alex and Neubig, Graham and Kulikov, Ilia and Harchaoui, Zaid},
	month = jun,
	year = {2024},
	note = {arXiv:2406.16838 [cs]},
}

@misc{hendrycks_measuring_2021,
	title = {Measuring {Massive} {Multitask} {Language} {Understanding}},
	url = {http://arxiv.org/abs/2009.03300},
	doi = {10.48550/arXiv.2009.03300},
	urldate = {2024-06-26},
	publisher = {arXiv},
	author = {Hendrycks, Dan and Burns, Collin and Basart, Steven and Zou, Andy and Mazeika, Mantas and Song, Dawn and Steinhardt, Jacob},
	month = jan,
	year = {2021},
	note = {arXiv:2009.03300 [cs]},
}

@misc{yao_tau-bench_2024,
	title = {\${\textbackslash}tau\$-bench: {A} {Benchmark} for {Tool}-{Agent}-{User} {Interaction} in {Real}-{World} {Domains}},
	shorttitle = {\${\textbackslash}tau\$-bench},
	url = {http://arxiv.org/abs/2406.12045},
	language = {en},
	urldate = {2024-06-25},
	publisher = {arXiv},
	author = {Yao, Shunyu and Shinn, Noah and Razavi, Pedram and Narasimhan, Karthik},
	month = jun,
	year = {2024},
	note = {arXiv:2406.12045 [cs]},
}

@misc{kambhampati_llms_2024,
	title = {{LLMs} {Can}'t {Plan}, {But} {Can} {Help} {Planning} in {LLM}-{Modulo} {Frameworks}},
	url = {http://arxiv.org/abs/2402.01817},
	language = {en},
	urldate = {2024-05-31},
	publisher = {arXiv},
	author = {Kambhampati, Subbarao and Valmeekam, Karthik and Guan, Lin and Stechly, Kaya and Verma, Mudit and Bhambri, Siddhant and Saldyt, Lucas and Murthy, Anil},
	month = feb,
	year = {2024},
	note = {arXiv:2402.01817 [cs]},
}

@misc{chen_evaluating_2021,
	title = {Evaluating {Large} {Language} {Models} {Trained} on {Code}},
	url = {http://arxiv.org/abs/2107.03374},
	doi = {10.48550/arXiv.2107.03374},
	urldate = {2024-05-25},
	publisher = {arXiv},
	author = {Chen, Mark and Tworek, Jerry and Jun, Heewoo and Yuan, Qiming and Pinto, Henrique Ponde de Oliveira and Kaplan, Jared and Edwards, Harri and Burda, Yuri and Joseph, Nicholas and Brockman, Greg and Ray, Alex and Puri, Raul and Krueger, Gretchen and Petrov, Michael and Khlaaf, Heidy and Sastry, Girish and Mishkin, Pamela and Chan, Brooke and Gray, Scott and Ryder, Nick and Pavlov, Mikhail and Power, Alethea and Kaiser, Lukasz and Bavarian, Mohammad and Winter, Clemens and Tillet, Philippe and Such, Felipe Petroski and Cummings, Dave and Plappert, Matthias and Chantzis, Fotios and Barnes, Elizabeth and Herbert-Voss, Ariel and Guss, William Hebgen and Nichol, Alex and Paino, Alex and Tezak, Nikolas and Tang, Jie and Babuschkin, Igor and Balaji, Suchir and Jain, Shantanu and Saunders, William and Hesse, Christopher and Carr, Andrew N. and Leike, Jan and Achiam, Josh and Misra, Vedant and Morikawa, Evan and Radford, Alec and Knight, Matthew and Brundage, Miles and Murati, Mira and Mayer, Katie and Welinder, Peter and McGrew, Bob and Amodei, Dario and McCandlish, Sam and Sutskever, Ilya and Zaremba, Wojciech},
	month = jul,
	year = {2021},
	note = {arXiv:2107.03374 [cs]},
}

@misc{biderman_lessons_2024,
	title = {Lessons from the {Trenches} on {Reproducible} {Evaluation} of {Language} {Models}},
	url = {http://arxiv.org/abs/2405.14782},
	language = {en},
	urldate = {2024-05-25},
	publisher = {arXiv},
	author = {Biderman, Stella and Schoelkopf, Hailey and Sutawika, Lintang and Gao, Leo and Tow, Jonathan and Abbasi, Baber and Aji, Alham Fikri and Ammanamanchi, Pawan Sasanka and Black, Sidney and Clive, Jordan and DiPofi, Anthony and Etxaniz, Julen and Fattori, Benjamin and Forde, Jessica Zosa and Foster, Charles and Jaiswal, Mimansa and Lee, Wilson Y. and Li, Haonan and Lovering, Charles and Muennighoff, Niklas and Pavlick, Ellie and Phang, Jason and Skowron, Aviya and Tan, Samson and Tang, Xiangru and Wang, Kevin A. and Winata, Genta Indra and Yvon, François and Zou, Andy},
	month = may,
	year = {2024},
	note = {arXiv:2405.14782 [cs]},
}

@misc{liang_holistic_2023,
	title = {Holistic {Evaluation} of {Language} {Models}},
	url = {http://arxiv.org/abs/2211.09110},
	doi = {10.48550/arXiv.2211.09110},
	urldate = {2024-05-24},
	publisher = {arXiv},
	author = {Liang, Percy and Bommasani, Rishi and Lee, Tony and Tsipras, Dimitris and Soylu, Dilara and Yasunaga, Michihiro and Zhang, Yian and Narayanan, Deepak and Wu, Yuhuai and Kumar, Ananya and Newman, Benjamin and Yuan, Binhang and Yan, Bobby and Zhang, Ce and Cosgrove, Christian and Manning, Christopher D. and Ré, Christopher and Acosta-Navas, Diana and Hudson, Drew A. and Zelikman, Eric and Durmus, Esin and Ladhak, Faisal and Rong, Frieda and Ren, Hongyu and Yao, Huaxiu and Wang, Jue and Santhanam, Keshav and Orr, Laurel and Zheng, Lucia and Yuksekgonul, Mert and Suzgun, Mirac and Kim, Nathan and Guha, Neel and Chatterji, Niladri and Khattab, Omar and Henderson, Peter and Huang, Qian and Chi, Ryan and Xie, Sang Michael and Santurkar, Shibani and Ganguli, Surya and Hashimoto, Tatsunori and Icard, Thomas and Zhang, Tianyi and Chaudhary, Vishrav and Wang, William and Li, Xuechen and Mai, Yifan and Zhang, Yuhui and Koreeda, Yuta},
	month = oct,
	year = {2023},
	note = {arXiv:2211.09110 [cs]},
}

@misc{shinn_reflexion_2023,
	title = {Reflexion: {Language} {Agents} with {Verbal} {Reinforcement} {Learning}},
	shorttitle = {Reflexion},
	url = {http://arxiv.org/abs/2303.11366},
	language = {en},
	urldate = {2024-05-24},
	publisher = {arXiv},
	author = {Shinn, Noah and Cassano, Federico and Berman, Edward and Gopinath, Ashwin and Narasimhan, Karthik and Yao, Shunyu},
	month = oct,
	year = {2023},
	note = {arXiv:2303.11366 [cs]},
}

@misc{li_more_2024,
	title = {More {Agents} {Is} {All} {You} {Need}},
	url = {http://arxiv.org/abs/2402.05120},
	doi = {10.48550/arXiv.2402.05120},
	urldate = {2024-05-24},
	publisher = {arXiv},
	author = {Li, Junyou and Zhang, Qin and Yu, Yangbin and Fu, Qiang and Ye, Deheng},
	month = feb,
	year = {2024},
	note = {arXiv:2402.05120 [cs]},
}

@misc{chen_are_2024,
	title = {Are {More} {LLM} {Calls} {All} {You} {Need}? {Towards} {Scaling} {Laws} of {Compound} {Inference} {Systems}},
	shorttitle = {Are {More} {LLM} {Calls} {All} {You} {Need}?},
	url = {http://arxiv.org/abs/2403.02419},
	language = {en},
	urldate = {2024-05-24},
	publisher = {arXiv},
	author = {Chen, Lingjiao and Davis, Jared Quincy and Hanin, Boris and Bailis, Peter and Stoica, Ion and Zaharia, Matei and Zou, James},
	month = mar,
	year = {2024},
	note = {arXiv:2403.02419 [cs, eess]},
}

@misc{gundawar_robust_2024,
	title = {Robust {Planning} with {Compound} {LLM} {Architectures}: {An} {LLM}-{Modulo} {Approach}},
	shorttitle = {Robust {Planning} with {Compound} {LLM} {Architectures}},
	url = {http://arxiv.org/abs/2411.14484},
	doi = {10.48550/arXiv.2411.14484},
	urldate = {2024-11-26},
	publisher = {arXiv},
	author = {Gundawar, Atharva and Valmeekam, Karthik and Verma, Mudit and Kambhampati, Subbarao},
	month = nov,
	year = {2024},
	note = {arXiv:2411.14484},
}

@misc{tufano_unit_2021,
	title = {Unit {Test} {Case} {Generation} with {Transformers} and {Focal} {Context}},
	url = {http://arxiv.org/abs/2009.05617},
	doi = {10.48550/arXiv.2009.05617},
	urldate = {2024-11-18},
	publisher = {arXiv},
	author = {Tufano, Michele and Drain, Dawn and Svyatkovskiy, Alexey and Deng, Shao Kun and Sundaresan, Neel},
	month = may,
	year = {2021},
	note = {arXiv:2009.05617},
}

@inproceedings{lukasczyk_pynguin_2022,
	address = {New York, NY, USA},
	series = {{ICSE} '22},
	title = {Pynguin: automated unit test generation for {Python}},
	isbn = {978-1-4503-9223-5},
	shorttitle = {Pynguin},
	url = {https://dl.acm.org/doi/10.1145/3510454.3516829},
	doi = {10.1145/3510454.3516829},
	urldate = {2024-11-18},
	booktitle = {Proceedings of the {ACM}/{IEEE} 44th {International} {Conference} on {Software} {Engineering}: {Companion} {Proceedings}},
	publisher = {Association for Computing Machinery},
	author = {Lukasczyk, Stephan and Fraser, Gordon},
	month = oct,
	year = {2022},
	pages = {168--172},
}

@inproceedings{chen_chatunitest_2024,
	address = {New York, NY, USA},
	series = {{FSE} 2024},
	title = {{ChatUniTest}: {A} {Framework} for {LLM}-{Based} {Test} {Generation}},
	isbn = {9798400706585},
	shorttitle = {{ChatUniTest}},
	url = {https://dl.acm.org/doi/10.1145/3663529.3663801},
	doi = {10.1145/3663529.3663801},
	urldate = {2024-11-15},
	booktitle = {Companion {Proceedings} of the 32nd {ACM} {International} {Conference} on the {Foundations} of {Software} {Engineering}},
	publisher = {Association for Computing Machinery},
	author = {Chen, Yinghao and Hu, Zehao and Zhi, Chen and Han, Junxiao and Deng, Shuiguang and Yin, Jianwei},
	month = jul,
	year = {2024},
	pages = {572--576},
}

@misc{chen_when_2024,
	title = {When is {Tree} {Search} {Useful} for {LLM} {Planning}? {It} {Depends} on the {Discriminator}},
	shorttitle = {When is {Tree} {Search} {Useful} for {LLM} {Planning}?},
	url = {http://arxiv.org/abs/2402.10890},
	urldate = {2024-11-15},
	publisher = {arXiv},
	author = {Chen, Ziru and White, Michael and Mooney, Raymond and Payani, Ali and Su, Yu and Sun, Huan},
	month = jun,
	year = {2024},
	note = {arXiv:2402.10890},
}

@misc{cook_ticking_2024,
	title = {{TICKing} {All} the {Boxes}: {Generated} {Checklists} {Improve} {LLM} {Evaluation} and {Generation}},
	shorttitle = {{TICKing} {All} the {Boxes}},
	url = {http://arxiv.org/abs/2410.03608},
	urldate = {2024-11-13},
	publisher = {arXiv},
	author = {Cook, Jonathan and Rocktäschel, Tim and Foerster, Jakob and Aumiller, Dennis and Wang, Alex},
	month = oct,
	year = {2024},
	note = {arXiv:2410.03608},
}

@misc{dubey_llama_2024,
	title = {The {Llama} 3 {Herd} of {Models}},
	url = {http://arxiv.org/abs/2407.21783},
	doi = {10.48550/arXiv.2407.21783},
	urldate = {2024-11-11},
	publisher = {arXiv},
	author = {Dubey, Abhimanyu and Jauhri, Abhinav and Pandey, Abhinav and Kadian, Abhishek and Al-Dahle, Ahmad and Letman, Aiesha and Mathur, Akhil and Schelten, Alan and Yang, Amy and Fan, Angela and Goyal, Anirudh and Hartshorn, Anthony and Yang, Aobo and Mitra, Archi and Sravankumar, Archie and Korenev, Artem and Hinsvark, Arthur and Rao, Arun and Zhang, Aston and Rodriguez, Aurelien and Gregerson, Austen and Spataru, Ava and Roziere, Baptiste and Biron, Bethany and Tang, Binh and Chern, Bobbie and Caucheteux, Charlotte and Nayak, Chaya and Bi, Chloe and Marra, Chris and McConnell, Chris and Keller, Christian and Touret, Christophe and Wu, Chunyang and Wong, Corinne and Ferrer, Cristian Canton and Nikolaidis, Cyrus and Allonsius, Damien and Song, Daniel and Pintz, Danielle and Livshits, Danny and Esiobu, David and Choudhary, Dhruv and Mahajan, Dhruv and Garcia-Olano, Diego and Perino, Diego and Hupkes, Dieuwke and Lakomkin, Egor and AlBadawy, Ehab and Lobanova, Elina and Dinan, Emily and Smith, Eric Michael and Radenovic, Filip and Zhang, Frank and Synnaeve, Gabriel and Lee, Gabrielle and Anderson, Georgia Lewis and Nail, Graeme and Mialon, Gregoire and Pang, Guan and Cucurell, Guillem and Nguyen, Hailey and Korevaar, Hannah and Xu, Hu and Touvron, Hugo and Zarov, Iliyan and Ibarra, Imanol Arrieta and Kloumann, Isabel and Misra, Ishan and Evtimov, Ivan and Copet, Jade and Lee, Jaewon and Geffert, Jan and Vranes, Jana and Park, Jason and Mahadeokar, Jay and Shah, Jeet and Linde, Jelmer van der and Billock, Jennifer and Hong, Jenny and Lee, Jenya and Fu, Jeremy and Chi, Jianfeng and Huang, Jianyu and Liu, Jiawen and Wang, Jie and Yu, Jiecao and Bitton, Joanna and Spisak, Joe and Park, Jongsoo and Rocca, Joseph and Johnstun, Joshua and Saxe, Joshua and Jia, Junteng and Alwala, Kalyan Vasuden and Upasani, Kartikeya and Plawiak, Kate and Li, Ke and Heafield, Kenneth and Stone, Kevin and El-Arini, Khalid and Iyer, Krithika and Malik, Kshitiz and Chiu, Kuenley and Bhalla, Kunal and Rantala-Yeary, Lauren and Maaten, Laurens van der and Chen, Lawrence and Tan, Liang and Jenkins, Liz and Martin, Louis and Madaan, Lovish and Malo, Lubo and Blecher, Lukas and Landzaat, Lukas and Oliveira, Luke de and Muzzi, Madeline and Pasupuleti, Mahesh and Singh, Mannat and Paluri, Manohar and Kardas, Marcin and Oldham, Mathew and Rita, Mathieu and Pavlova, Maya and Kambadur, Melanie and Lewis, Mike and Si, Min and Singh, Mitesh Kumar and Hassan, Mona and Goyal, Naman and Torabi, Narjes and Bashlykov, Nikolay and Bogoychev, Nikolay and Chatterji, Niladri and Duchenne, Olivier and Çelebi, Onur and Alrassy, Patrick and Zhang, Pengchuan and Li, Pengwei and Vasic, Petar and Weng, Peter and Bhargava, Prajjwal and Dubal, Pratik and Krishnan, Praveen and Koura, Punit Singh and Xu, Puxin and He, Qing and Dong, Qingxiao and Srinivasan, Ragavan and Ganapathy, Raj and Calderer, Ramon and Cabral, Ricardo Silveira and Stojnic, Robert and Raileanu, Roberta and Girdhar, Rohit and Patel, Rohit and Sauvestre, Romain and Polidoro, Ronnie and Sumbaly, Roshan and Taylor, Ross and Silva, Ruan and Hou, Rui and Wang, Rui and Hosseini, Saghar and Chennabasappa, Sahana and Singh, Sanjay and Bell, Sean and Kim, Seohyun Sonia and Edunov, Sergey and Nie, Shaoliang and Narang, Sharan and Raparthy, Sharath and Shen, Sheng and Wan, Shengye and Bhosale, Shruti and Zhang, Shun and Vandenhende, Simon and Batra, Soumya and Whitman, Spencer and Sootla, Sten and Collot, Stephane and Gururangan, Suchin and Borodinsky, Sydney and Herman, Tamar and Fowler, Tara and Sheasha, Tarek and Georgiou, Thomas and Scialom, Thomas and Speckbacher, Tobias and Mihaylov, Todor and Xiao, Tong and Karn, Ujjwal and Goswami, Vedanuj and Gupta, Vibhor and Ramanathan, Vignesh and Kerkez, Viktor and Gonguet, Vincent and Do, Virginie and Vogeti, Vish and Petrovic, Vladan and Chu, Weiwei and Xiong, Wenhan and Fu, Wenyin and Meers, Whitney and Martinet, Xavier and Wang, Xiaodong and Tan, Xiaoqing Ellen and Xie, Xinfeng and Jia, Xuchao and Wang, Xuewei and Goldschlag, Yaelle and Gaur, Yashesh and Babaei, Yasmine and Wen, Yi and Song, Yiwen and Zhang, Yuchen and Li, Yue and Mao, Yuning and Coudert, Zacharie Delpierre and Yan, Zheng and Chen, Zhengxing and Papakipos, Zoe and Singh, Aaditya and Grattafiori, Aaron and Jain, Abha and Kelsey, Adam and Shajnfeld, Adam and Gangidi, Adithya and Victoria, Adolfo and Goldstand, Ahuva and Menon, Ajay and Sharma, Ajay and Boesenberg, Alex and Vaughan, Alex and Baevski, Alexei and Feinstein, Allie and Kallet, Amanda and Sangani, Amit and Yunus, Anam and Lupu, Andrei and Alvarado, Andres and Caples, Andrew and Gu, Andrew and Ho, Andrew and Poulton, Andrew and Ryan, Andrew and Ramchandani, Ankit and Franco, Annie and Saraf, Aparajita and Chowdhury, Arkabandhu and Gabriel, Ashley and Bharambe, Ashwin and Eisenman, Assaf and Yazdan, Azadeh and James, Beau and Maurer, Ben and Leonhardi, Benjamin and Huang, Bernie and Loyd, Beth and Paola, Beto De and Paranjape, Bhargavi and Liu, Bing and Wu, Bo and Ni, Boyu and Hancock, Braden and Wasti, Bram and Spence, Brandon and Stojkovic, Brani and Gamido, Brian and Montalvo, Britt and Parker, Carl and Burton, Carly and Mejia, Catalina and Wang, Changhan and Kim, Changkyu and Zhou, Chao and Hu, Chester and Chu, Ching-Hsiang and Cai, Chris and Tindal, Chris and Feichtenhofer, Christoph and Civin, Damon and Beaty, Dana and Kreymer, Daniel and Li, Daniel and Wyatt, Danny and Adkins, David and Xu, David and Testuggine, Davide and David, Delia and Parikh, Devi and Liskovich, Diana and Foss, Didem and Wang, Dingkang and Le, Duc and Holland, Dustin and Dowling, Edward and Jamil, Eissa and Montgomery, Elaine and Presani, Eleonora and Hahn, Emily and Wood, Emily and Brinkman, Erik and Arcaute, Esteban and Dunbar, Evan and Smothers, Evan and Sun, Fei and Kreuk, Felix and Tian, Feng and Ozgenel, Firat and Caggioni, Francesco and Guzmán, Francisco and Kanayet, Frank and Seide, Frank and Florez, Gabriela Medina and Schwarz, Gabriella and Badeer, Gada and Swee, Georgia and Halpern, Gil and Thattai, Govind and Herman, Grant and Sizov, Grigory and Guangyi and Zhang and Lakshminarayanan, Guna and Shojanazeri, Hamid and Zou, Han and Wang, Hannah and Zha, Hanwen and Habeeb, Haroun and Rudolph, Harrison and Suk, Helen and Aspegren, Henry and Goldman, Hunter and Damlaj, Ibrahim and Molybog, Igor and Tufanov, Igor and Veliche, Irina-Elena and Gat, Itai and Weissman, Jake and Geboski, James and Kohli, James and Asher, Japhet and Gaya, Jean-Baptiste and Marcus, Jeff and Tang, Jeff and Chan, Jennifer and Zhen, Jenny and Reizenstein, Jeremy and Teboul, Jeremy and Zhong, Jessica and Jin, Jian and Yang, Jingyi and Cummings, Joe and Carvill, Jon and Shepard, Jon and McPhie, Jonathan and Torres, Jonathan and Ginsburg, Josh and Wang, Junjie and Wu, Kai and U, Kam Hou and Saxena, Karan and Prasad, Karthik and Khandelwal, Kartikay and Zand, Katayoun and Matosich, Kathy and Veeraraghavan, Kaushik and Michelena, Kelly and Li, Keqian and Huang, Kun and Chawla, Kunal and Lakhotia, Kushal and Huang, Kyle and Chen, Lailin and Garg, Lakshya and A, Lavender and Silva, Leandro and Bell, Lee and Zhang, Lei and Guo, Liangpeng and Yu, Licheng and Moshkovich, Liron and Wehrstedt, Luca and Khabsa, Madian and Avalani, Manav and Bhatt, Manish and Tsimpoukelli, Maria and Mankus, Martynas and Hasson, Matan and Lennie, Matthew and Reso, Matthias and Groshev, Maxim and Naumov, Maxim and Lathi, Maya and Keneally, Meghan and Seltzer, Michael L. and Valko, Michal and Restrepo, Michelle and Patel, Mihir and Vyatskov, Mik and Samvelyan, Mikayel and Clark, Mike and Macey, Mike and Wang, Mike and Hermoso, Miquel Jubert and Metanat, Mo and Rastegari, Mohammad and Bansal, Munish and Santhanam, Nandhini and Parks, Natascha and White, Natasha and Bawa, Navyata and Singhal, Nayan and Egebo, Nick and Usunier, Nicolas and Laptev, Nikolay Pavlovich and Dong, Ning and Zhang, Ning and Cheng, Norman and Chernoguz, Oleg and Hart, Olivia and Salpekar, Omkar and Kalinli, Ozlem and Kent, Parkin and Parekh, Parth and Saab, Paul and Balaji, Pavan and Rittner, Pedro and Bontrager, Philip and Roux, Pierre and Dollar, Piotr and Zvyagina, Polina and Ratanchandani, Prashant and Yuvraj, Pritish and Liang, Qian and Alao, Rachad and Rodriguez, Rachel and Ayub, Rafi and Murthy, Raghotham and Nayani, Raghu and Mitra, Rahul and Li, Raymond and Hogan, Rebekkah and Battey, Robin and Wang, Rocky and Maheswari, Rohan and Howes, Russ and Rinott, Ruty and Bondu, Sai Jayesh and Datta, Samyak and Chugh, Sara and Hunt, Sara and Dhillon, Sargun and Sidorov, Sasha and Pan, Satadru and Verma, Saurabh and Yamamoto, Seiji and Ramaswamy, Sharadh and Lindsay, Shaun and Lindsay, Shaun and Feng, Sheng and Lin, Shenghao and Zha, Shengxin Cindy and Shankar, Shiva and Zhang, Shuqiang and Zhang, Shuqiang and Wang, Sinong and Agarwal, Sneha and Sajuyigbe, Soji and Chintala, Soumith and Max, Stephanie and Chen, Stephen and Kehoe, Steve and Satterfield, Steve and Govindaprasad, Sudarshan and Gupta, Sumit and Cho, Sungmin and Virk, Sunny and Subramanian, Suraj and Choudhury, Sy and Goldman, Sydney and Remez, Tal and Glaser, Tamar and Best, Tamara and Kohler, Thilo and Robinson, Thomas and Li, Tianhe and Zhang, Tianjun and Matthews, Tim and Chou, Timothy and Shaked, Tzook and Vontimitta, Varun and Ajayi, Victoria and Montanez, Victoria and Mohan, Vijai and Kumar, Vinay Satish and Mangla, Vishal and Albiero, Vítor and Ionescu, Vlad and Poenaru, Vlad and Mihailescu, Vlad Tiberiu and Ivanov, Vladimir and Li, Wei and Wang, Wenchen and Jiang, Wenwen and Bouaziz, Wes and Constable, Will and Tang, Xiaocheng and Wang, Xiaofang and Wu, Xiaojian and Wang, Xiaolan and Xia, Xide and Wu, Xilun and Gao, Xinbo and Chen, Yanjun and Hu, Ye and Jia, Ye and Qi, Ye and Li, Yenda and Zhang, Yilin and Zhang, Ying and Adi, Yossi and Nam, Youngjin and Yu and Wang and Hao, Yuchen and Qian, Yundi and He, Yuzi and Rait, Zach and DeVito, Zachary and Rosnbrick, Zef and Wen, Zhaoduo and Yang, Zhenyu and Zhao, Zhiwei},
	month = aug,
	year = {2024},
	note = {arXiv:2407.21783},
}

@misc{roziere_code_2024,
	title = {Code {Llama}: {Open} {Foundation} {Models} for {Code}},
	shorttitle = {Code {Llama}},
	url = {http://arxiv.org/abs/2308.12950},
	doi = {10.48550/arXiv.2308.12950},
	urldate = {2024-11-11},
	publisher = {arXiv},
	author = {Rozière, Baptiste and Gehring, Jonas and Gloeckle, Fabian and Sootla, Sten and Gat, Itai and Tan, Xiaoqing Ellen and Adi, Yossi and Liu, Jingyu and Sauvestre, Romain and Remez, Tal and Rapin, Jérémy and Kozhevnikov, Artyom and Evtimov, Ivan and Bitton, Joanna and Bhatt, Manish and Ferrer, Cristian Canton and Grattafiori, Aaron and Xiong, Wenhan and Défossez, Alexandre and Copet, Jade and Azhar, Faisal and Touvron, Hugo and Martin, Louis and Usunier, Nicolas and Scialom, Thomas and Synnaeve, Gabriel},
	month = jan,
	year = {2024},
	note = {arXiv:2308.12950},
}

@misc{zhang_naturalcodebench_2024,
	title = {{NaturalCodeBench}: {Examining} {Coding} {Performance} {Mismatch} on {HumanEval} and {Natural} {User} {Prompts}},
	shorttitle = {{NaturalCodeBench}},
	url = {http://arxiv.org/abs/2405.04520},
	urldate = {2024-11-07},
	publisher = {arXiv},
	author = {Zhang, Shudan and Zhao, Hanlin and Liu, Xiao and Zheng, Qinkai and Qi, Zehan and Gu, Xiaotao and Zhang, Xiaohan and Dong, Yuxiao and Tang, Jie},
	month = may,
	year = {2024},
	note = {arXiv:2405.04520},
}

@misc{chen_codet_2022,
	title = {{CodeT}: {Code} {Generation} with {Generated} {Tests}},
	shorttitle = {{CodeT}},
	url = {http://arxiv.org/abs/2207.10397},
	urldate = {2024-11-07},
	publisher = {arXiv},
	author = {Chen, Bei and Zhang, Fengji and Nguyen, Anh and Zan, Daoguang and Lin, Zeqi and Lou, Jian-Guang and Chen, Weizhu},
	month = nov,
	year = {2022},
	note = {arXiv:2207.10397},
}

@misc{liang_improving_2024,
	title = {Improving {LLM} {Reasoning} through {Scaling} {Inference} {Computation} with {Collaborative} {Verification}},
	url = {http://arxiv.org/abs/2410.05318},
	urldate = {2024-11-06},
	publisher = {arXiv},
	author = {Liang, Zhenwen and Liu, Ye and Niu, Tong and Zhang, Xiangliang and Zhou, Yingbo and Yavuz, Semih},
	month = oct,
	year = {2024},
	note = {arXiv:2410.05318},
}

@misc{madaan_self-refine_2023,
	title = {Self-{Refine}: {Iterative} {Refinement} with {Self}-{Feedback}},
	shorttitle = {Self-{Refine}},
	url = {http://arxiv.org/abs/2303.17651},
	urldate = {2024-11-05},
	publisher = {arXiv},
	author = {Madaan, Aman and Tandon, Niket and Gupta, Prakhar and Hallinan, Skyler and Gao, Luyu and Wiegreffe, Sarah and Alon, Uri and Dziri, Nouha and Prabhumoye, Shrimai and Yang, Yiming and Gupta, Shashank and Majumder, Bodhisattwa Prasad and Hermann, Katherine and Welleck, Sean and Yazdanbakhsh, Amir and Clark, Peter},
	month = may,
	year = {2023},
	note = {arXiv:2303.17651},
}

@misc{wang_soft_2024,
	title = {Soft {Self}-{Consistency} {Improves} {Language} {Model} {Agents}},
	url = {http://arxiv.org/abs/2402.13212},
	urldate = {2024-11-05},
	publisher = {arXiv},
	author = {Wang, Han and Prasad, Archiki and Stengel-Eskin, Elias and Bansal, Mohit},
	month = jun,
	year = {2024},
	note = {arXiv:2402.13212},
}

@misc{saad-falcon_archon_2024,
	title = {Archon: {An} {Architecture} {Search} {Framework} for {Inference}-{Time} {Techniques}},
	shorttitle = {Archon},
	url = {http://arxiv.org/abs/2409.15254},
	urldate = {2024-11-03},
	publisher = {arXiv},
	author = {Saad-Falcon, Jon and Lafuente, Adrian Gamarra and Natarajan, Shlok and Maru, Nahum and Todorov, Hristo and Guha, Etash and Buchanan, E. Kelly and Chen, Mayee and Guha, Neel and Ré, Christopher and Mirhoseini, Azalia},
	month = oct,
	year = {2024},
	note = {arXiv:2409.15254},
}

@misc{he_webvoyager_2024,
	title = {{WebVoyager}: {Building} an {End}-to-{End} {Web} {Agent} with {Large} {Multimodal} {Models}},
	shorttitle = {{WebVoyager}},
	url = {http://arxiv.org/abs/2401.13919},
	doi = {10.48550/arXiv.2401.13919},
	urldate = {2024-10-28},
	publisher = {arXiv},
	author = {He, Hongliang and Yao, Wenlin and Ma, Kaixin and Yu, Wenhao and Dai, Yong and Zhang, Hongming and Lan, Zhenzhong and Yu, Dong},
	month = jun,
	year = {2024},
	note = {arXiv:2401.13919},
}

@misc{bai_digirl_2024,
	title = {{DigiRL}: {Training} {In}-{The}-{Wild} {Device}-{Control} {Agents} with {Autonomous} {Reinforcement} {Learning}},
	shorttitle = {{DigiRL}},
	url = {http://arxiv.org/abs/2406.11896},
	doi = {10.48550/arXiv.2406.11896},
	urldate = {2024-10-28},
	publisher = {arXiv},
	author = {Bai, Hao and Zhou, Yifei and Cemri, Mert and Pan, Jiayi and Suhr, Alane and Levine, Sergey and Kumar, Aviral},
	month = jun,
	year = {2024},
	note = {arXiv:2406.11896},
}

@misc{wang_self-consistency_2023,
	title = {Self-{Consistency} {Improves} {Chain} of {Thought} {Reasoning} in {Language} {Models}},
	url = {http://arxiv.org/abs/2203.11171},
	doi = {10.48550/arXiv.2203.11171},
	urldate = {2024-10-28},
	publisher = {arXiv},
	author = {Wang, Xuezhi and Wei, Jason and Schuurmans, Dale and Le, Quoc and Chi, Ed and Narang, Sharan and Chowdhery, Aakanksha and Zhou, Denny},
	month = mar,
	year = {2023},
	note = {arXiv:2203.11171},
}

@misc{yang_leandojo_2023,
	title = {{LeanDojo}: {Theorem} {Proving} with {Retrieval}-{Augmented} {Language} {Models}},
	shorttitle = {{LeanDojo}},
	url = {http://arxiv.org/abs/2306.15626},
	doi = {10.48550/arXiv.2306.15626},
	urldate = {2024-10-28},
	publisher = {arXiv},
	author = {Yang, Kaiyu and Swope, Aidan M. and Gu, Alex and Chalamala, Rahul and Song, Peiyang and Yu, Shixing and Godil, Saad and Prenger, Ryan and Anandkumar, Anima},
	month = oct,
	year = {2023},
	note = {arXiv:2306.15626},
}

@misc{huang_mustard_2024,
	title = {{MUSTARD}: {Mastering} {Uniform} {Synthesis} of {Theorem} and {Proof} {Data}},
	shorttitle = {{MUSTARD}},
	url = {http://arxiv.org/abs/2402.08957},
	doi = {10.48550/arXiv.2402.08957},
	urldate = {2024-10-28},
	publisher = {arXiv},
	author = {Huang, Yinya and Lin, Xiaohan and Liu, Zhengying and Cao, Qingxing and Xin, Huajian and Wang, Haiming and Li, Zhenguo and Song, Linqi and Liang, Xiaodan},
	month = may,
	year = {2024},
	note = {arXiv:2402.08957},
}

@misc{azerbayev_llemma_2024,
	title = {Llemma: {An} {Open} {Language} {Model} {For} {Mathematics}},
	shorttitle = {Llemma},
	url = {http://arxiv.org/abs/2310.10631},
	doi = {10.48550/arXiv.2310.10631},
	urldate = {2024-10-28},
	publisher = {arXiv},
	author = {Azerbayev, Zhangir and Schoelkopf, Hailey and Paster, Keiran and Santos, Marco Dos and McAleer, Stephen and Jiang, Albert Q. and Deng, Jia and Biderman, Stella and Welleck, Sean},
	month = mar,
	year = {2024},
	note = {arXiv:2310.10631},
}

@misc{thakur_-context_2024,
	title = {An {In}-{Context} {Learning} {Agent} for {Formal} {Theorem}-{Proving}},
	url = {http://arxiv.org/abs/2310.04353},
	doi = {10.48550/arXiv.2310.04353},
	urldate = {2024-10-28},
	publisher = {arXiv},
	author = {Thakur, Amitayush and Tsoukalas, George and Wen, Yeming and Xin, Jimmy and Chaudhuri, Swarat},
	month = aug,
	year = {2024},
	note = {arXiv:2310.04353},
}

@misc{first_baldur_2023,
	title = {Baldur: {Whole}-{Proof} {Generation} and {Repair} with {Large} {Language} {Models}},
	shorttitle = {Baldur},
	url = {http://arxiv.org/abs/2303.04910},
	doi = {10.48550/arXiv.2303.04910},
	urldate = {2024-10-28},
	publisher = {arXiv},
	author = {First, Emily and Rabe, Markus N. and Ringer, Talia and Brun, Yuriy},
	month = mar,
	year = {2023},
	note = {arXiv:2303.04910},
}

@misc{wang_lego-prover_2023,
	title = {{LEGO}-{Prover}: {Neural} {Theorem} {Proving} with {Growing} {Libraries}},
	shorttitle = {{LEGO}-{Prover}},
	url = {http://arxiv.org/abs/2310.00656},
	doi = {10.48550/arXiv.2310.00656},
	urldate = {2024-10-28},
	publisher = {arXiv},
	author = {Wang, Haiming and Xin, Huajian and Zheng, Chuanyang and Li, Lin and Liu, Zhengying and Cao, Qingxing and Huang, Yinya and Xiong, Jing and Shi, Han and Xie, Enze and Yin, Jian and Li, Zhenguo and Liao, Heng and Liang, Xiaodan},
	month = oct,
	year = {2023},
	note = {arXiv:2310.00656},
}

@misc{xin_deepseek-prover_2024,
	title = {{DeepSeek}-{Prover}: {Advancing} {Theorem} {Proving} in {LLMs} through {Large}-{Scale} {Synthetic} {Data}},
	shorttitle = {{DeepSeek}-{Prover}},
	url = {http://arxiv.org/abs/2405.14333},
	urldate = {2024-10-28},
	publisher = {arXiv},
	author = {Xin, Huajian and Guo, Daya and Shao, Zhihong and Ren, Zhizhou and Zhu, Qihao and Liu, Bo and Ruan, Chong and Li, Wenda and Liang, Xiaodan},
	month = may,
	year = {2024},
	note = {arXiv:2405.14333},
}

@misc{li_common_2024,
	title = {Common {7B} {Language} {Models} {Already} {Possess} {Strong} {Math} {Capabilities}},
	url = {http://arxiv.org/abs/2403.04706},
	doi = {10.48550/arXiv.2403.04706},
	urldate = {2024-10-28},
	publisher = {arXiv},
	author = {Li, Chen and Wang, Weiqi and Hu, Jingcheng and Wei, Yixuan and Zheng, Nanning and Hu, Han and Zhang, Zheng and Peng, Houwen},
	month = mar,
	year = {2024},
	note = {arXiv:2403.04706},
}

@misc{liu_improving_2023,
	title = {Improving {Large} {Language} {Model} {Fine}-tuning for {Solving} {Math} {Problems}},
	url = {http://arxiv.org/abs/2310.10047},
	doi = {10.48550/arXiv.2310.10047},
	urldate = {2024-10-28},
	publisher = {arXiv},
	author = {Liu, Yixin and Singh, Avi and Freeman, C. Daniel and Co-Reyes, John D. and Liu, Peter J.},
	month = oct,
	year = {2023},
	note = {arXiv:2310.10047},
}

@misc{lightman_lets_2023,
	title = {Let's {Verify} {Step} by {Step}},
	url = {http://arxiv.org/abs/2305.20050},
	urldate = {2024-10-27},
	publisher = {arXiv},
	author = {Lightman, Hunter and Kosaraju, Vineet and Burda, Yura and Edwards, Harri and Baker, Bowen and Lee, Teddy and Leike, Jan and Schulman, John and Sutskever, Ilya and Cobbe, Karl},
	month = may,
	year = {2023},
	note = {arXiv:2305.20050},
}

@misc{zhuge_agent-as--judge_2024,
	title = {Agent-as-a-{Judge}: {Evaluate} {Agents} with {Agents}},
	shorttitle = {Agent-as-a-{Judge}},
	url = {http://arxiv.org/abs/2410.10934},
	urldate = {2024-10-24},
	publisher = {arXiv},
	author = {Zhuge, Mingchen and Zhao, Changsheng and Ashley, Dylan and Wang, Wenyi and Khizbullin, Dmitrii and Xiong, Yunyang and Liu, Zechun and Chang, Ernie and Krishnamoorthi, Raghuraman and Tian, Yuandong and Shi, Yangyang and Chandra, Vikas and Schmidhuber, Jürgen},
	month = oct,
	year = {2024},
	note = {arXiv:2410.10934},
}

@misc{amodei_concrete_2016,
	title = {Concrete {Problems} in {AI} {Safety}},
	url = {http://arxiv.org/abs/1606.06565},
	doi = {10.48550/arXiv.1606.06565},
	urldate = {2024-10-17},
	publisher = {arXiv},
	author = {Amodei, Dario and Olah, Chris and Steinhardt, Jacob and Christiano, Paul and Schulman, John and Mané, Dan},
	month = jul,
	year = {2016},
	note = {arXiv:1606.06565},
}

@misc{yao_tree_2023,
	title = {Tree of {Thoughts}: {Deliberate} {Problem} {Solving} with {Large} {Language} {Models}},
	shorttitle = {Tree of {Thoughts}},
	url = {http://arxiv.org/abs/2305.10601},
	doi = {10.48550/arXiv.2305.10601},
	urldate = {2024-10-23},
	publisher = {arXiv},
	author = {Yao, Shunyu and Yu, Dian and Zhao, Jeffrey and Shafran, Izhak and Griffiths, Thomas L. and Cao, Yuan and Narasimhan, Karthik},
	month = dec,
	year = {2023},
	note = {arXiv:2305.10601},
}

@misc{dhuliawala_chain--verification_2023,
	title = {Chain-of-{Verification} {Reduces} {Hallucination} in {Large} {Language} {Models}},
	url = {http://arxiv.org/abs/2309.11495},
	doi = {10.48550/arXiv.2309.11495},
	urldate = {2024-10-21},
	publisher = {arXiv},
	author = {Dhuliawala, Shehzaad and Komeili, Mojtaba and Xu, Jing and Raileanu, Roberta and Li, Xian and Celikyilmaz, Asli and Weston, Jason},
	month = sep,
	year = {2023},
	note = {arXiv:2309.11495},
}

@misc{nijkamp_codegen_2023,
	title = {{CodeGen}: {An} {Open} {Large} {Language} {Model} for {Code} with {Multi}-{Turn} {Program} {Synthesis}},
	shorttitle = {{CodeGen}},
	url = {http://arxiv.org/abs/2203.13474},
	doi = {10.48550/arXiv.2203.13474},
	urldate = {2024-10-18},
	publisher = {arXiv},
	author = {Nijkamp, Erik and Pang, Bo and Hayashi, Hiroaki and Tu, Lifu and Wang, Huan and Zhou, Yingbo and Savarese, Silvio and Xiong, Caiming},
	month = feb,
	year = {2023},
	note = {arXiv:2203.13474},
}

@misc{setlur_rewarding_2024,
	title = {Rewarding {Progress}: {Scaling} {Automated} {Process} {Verifiers} for {LLM} {Reasoning}},
	shorttitle = {Rewarding {Progress}},
	url = {http://arxiv.org/abs/2410.08146},
	doi = {10.48550/arXiv.2410.08146},
	urldate = {2024-10-17},
	publisher = {arXiv},
	author = {Setlur, Amrith and Nagpal, Chirag and Fisch, Adam and Geng, Xinyang and Eisenstein, Jacob and Agarwal, Rishabh and Agarwal, Alekh and Berant, Jonathan and Kumar, Aviral},
	month = oct,
	year = {2024},
	note = {arXiv:2410.08146},
}

@misc{wu_thinking_2024,
	title = {Thinking {LLMs}: {General} {Instruction} {Following} with {Thought} {Generation}},
	shorttitle = {Thinking {LLMs}},
	url = {http://arxiv.org/abs/2410.10630},
	doi = {10.48550/arXiv.2410.10630},
	urldate = {2024-10-17},
	publisher = {arXiv},
	author = {Wu, Tianhao and Lan, Janice and Yuan, Weizhe and Jiao, Jiantao and Weston, Jason and Sukhbaatar, Sainbayar},
	month = oct,
	year = {2024},
	note = {arXiv:2410.10630},
}

@misc{zhong_debug_2024,
	title = {Debug like a {Human}: {A} {Large} {Language} {Model} {Debugger} via {Verifying} {Runtime} {Execution} {Step}-by-step},
	shorttitle = {Debug like a {Human}},
	url = {http://arxiv.org/abs/2402.16906},
	doi = {10.48550/arXiv.2402.16906},
	urldate = {2024-10-17},
	publisher = {arXiv},
	author = {Zhong, Li and Wang, Zilong and Shang, Jingbo},
	month = jun,
	year = {2024},
	note = {arXiv:2402.16906},
}

@misc{wei_chain--thought_2023,
	title = {Chain-of-{Thought} {Prompting} {Elicits} {Reasoning} in {Large} {Language} {Models}},
	url = {http://arxiv.org/abs/2201.11903},
	doi = {10.48550/arXiv.2201.11903},
	urldate = {2024-10-17},
	publisher = {arXiv},
	author = {Wei, Jason and Wang, Xuezhi and Schuurmans, Dale and Bosma, Maarten and Ichter, Brian and Xia, Fei and Chi, Ed and Le, Quoc and Zhou, Denny},
	month = jan,
	year = {2023},
	note = {arXiv:2201.11903},
}

@misc{jain_llm-assisted_2023,
	title = {{LLM}-{Assisted} {Code} {Cleaning} {For} {Training} {Accurate} {Code} {Generators}},
	url = {http://arxiv.org/abs/2311.14904},
	urldate = {2024-10-17},
	publisher = {arXiv},
	author = {Jain, Naman and Zhang, Tianjun and Chiang, Wei-Lin and Gonzalez, Joseph E. and Sen, Koushik and Stoica, Ion},
	month = nov,
	year = {2023},
	note = {arXiv:2311.14904},
}

@article{borstler_developers_2023,
	title = {Developers talking about code quality},
	volume = {28},
	issn = {1573-7616},
	url = {https://doi.org/10.1007/s10664-023-10381-0},
	doi = {10.1007/s10664-023-10381-0},
	language = {en},
	number = {6},
	urldate = {2024-10-14},
	journal = {Empirical Software Engineering},
	author = {Börstler, Jürgen and Bennin, Kwabena E. and Hooshangi, Sara and Jeuring, Johan and Keuning, Hieke and Kleiner, Carsten and MacKellar, Bonnie and Duran, Rodrigo and Störrle, Harald and Toll, Daniel and van Assema, Jelle},
	month = sep,
	year = {2023},
	pages = {128},
}

@misc{siddiq_using_2024,
	title = {Using {Large} {Language} {Models} to {Generate} {JUnit} {Tests}: {An} {Empirical} {Study}},
	shorttitle = {Using {Large} {Language} {Models} to {Generate} {JUnit} {Tests}},
	url = {http://arxiv.org/abs/2305.00418},
	urldate = {2024-10-14},
	publisher = {arXiv},
	author = {Siddiq, Mohammed Latif and Santos, Joanna C. S. and Tanvir, Ridwanul Hasan and Ulfat, Noshin and Rifat, Fahmid Al and Lopes, Vinicius Carvalho},
	month = mar,
	year = {2024},
	note = {arXiv:2305.00418},
}

@inproceedings{liu_is_2023,
	title = {Is {Your} {Code} {Generated} by {ChatGPT} {Really} {Correct}? {Rigorous} {Evaluation} of {Large} {Language} {Models} for {Code} {Generation}},
	shorttitle = {Is {Your} {Code} {Generated} by {ChatGPT} {Really} {Correct}?},
	url = {https://openreview.net/forum?id=1qvx610Cu7},
	language = {en},
	urldate = {2024-10-14},
	author = {Liu, Jiawei and Xia, Chunqiu Steven and Wang, Yuyao and Zhang, Lingming},
	month = nov,
	year = {2023},
}

@misc{austin_program_2021,
	title = {Program {Synthesis} with {Large} {Language} {Models}},
	url = {http://arxiv.org/abs/2108.07732},
	urldate = {2024-10-14},
	publisher = {arXiv},
	author = {Austin, Jacob and Odena, Augustus and Nye, Maxwell and Bosma, Maarten and Michalewski, Henryk and Dohan, David and Jiang, Ellen and Cai, Carrie and Terry, Michael and Le, Quoc and Sutton, Charles},
	month = aug,
	year = {2021},
	note = {arXiv:2108.07732 [cs]},
}

@misc{zhang_generative_2024,
	title = {Generative {Verifiers}: {Reward} {Modeling} as {Next}-{Token} {Prediction}},
	shorttitle = {Generative {Verifiers}},
	url = {http://arxiv.org/abs/2408.15240},
	urldate = {2024-10-12},
	publisher = {arXiv},
	author = {Zhang, Lunjun and Hosseini, Arian and Bansal, Hritik and Kazemi, Mehran and Kumar, Aviral and Agarwal, Rishabh},
	month = aug,
	year = {2024},
	note = {arXiv:2408.15240 [cs]},
}

@misc{zheng_beyond_2024,
	title = {Beyond {Correctness}: {Benchmarking} {Multi}-dimensional {Code} {Generation} for {Large} {Language} {Models}},
	shorttitle = {Beyond {Correctness}},
	url = {http://arxiv.org/abs/2407.11470},
	doi = {10.48550/arXiv.2407.11470},
	urldate = {2024-10-13},
	publisher = {arXiv},
	author = {Zheng, Jiasheng and Cao, Boxi and Ma, Zhengzhao and Pan, Ruotong and Lin, Hongyu and Lu, Yaojie and Han, Xianpei and Sun, Le},
	month = oct,
	year = {2024},
	note = {arXiv:2407.11470},
}

@article{gulwani_program_2017,
	title = {Program {Synthesis}},
	volume = {4},
	issn = {2325-1107, 2325-1131},
	url = {https://www.nowpublishers.com/article/Details/PGL-010},
	doi = {10.1561/2500000010},
	language = {English},
	number = {1-2},
	urldate = {2024-10-13},
	journal = {Foundations and Trends® in Programming Languages},
	author = {Gulwani, Sumit and Polozov, Oleksandr and Singh, Rishabh},
	month = jul,
	year = {2017},
	note = {Publisher: Now Publishers, Inc.},
	pages = {1--119},
}

@misc{cobbe_training_2021,
	title = {Training {Verifiers} to {Solve} {Math} {Word} {Problems}},
	url = {http://arxiv.org/abs/2110.14168},
	urldate = {2024-10-12},
	publisher = {arXiv},
	author = {Cobbe, Karl and Kosaraju, Vineet and Bavarian, Mohammad and Chen, Mark and Jun, Heewoo and Kaiser, Lukasz and Plappert, Matthias and Tworek, Jerry and Hilton, Jacob and Nakano, Reiichiro and Hesse, Christopher and Schulman, John},
	month = nov,
	year = {2021},
	note = {arXiv:2110.14168 [cs]},
}

@misc{kirchner_prover-verifier_2024,
	title = {Prover-{Verifier} {Games} improve legibility of {LLM} outputs},
	url = {http://arxiv.org/abs/2407.13692},
	urldate = {2024-10-12},
	publisher = {arXiv},
	author = {Kirchner, Jan Hendrik and Chen, Yining and Edwards, Harri and Leike, Jan and McAleese, Nat and Burda, Yuri},
	month = aug,
	year = {2024},
	note = {arXiv:2407.13692 [cs]},
}

@misc{vacareanu_general_2024,
	title = {General {Purpose} {Verification} for {Chain} of {Thought} {Prompting}},
	url = {http://arxiv.org/abs/2405.00204},
	urldate = {2024-10-12},
	publisher = {arXiv},
	author = {Vacareanu, Robert and Pratik, Anurag and Spiliopoulou, Evangelia and Qi, Zheng and Paolini, Giovanni and John, Neha Anna and Ma, Jie and Benajiba, Yassine and Ballesteros, Miguel},
	month = apr,
	year = {2024},
	note = {arXiv:2405.00204 [cs]},
}

@misc{wang_planning_2024,
	title = {Planning {In} {Natural} {Language} {Improves} {LLM} {Search} {For} {Code} {Generation}},
	url = {http://arxiv.org/abs/2409.03733},
	urldate = {2024-10-12},
	publisher = {arXiv},
	author = {Wang, Evan and Cassano, Federico and Wu, Catherine and Bai, Yunfeng and Song, Will and Nath, Vaskar and Han, Ziwen and Hendryx, Sean and Yue, Summer and Zhang, Hugh},
	month = sep,
	year = {2024},
	note = {arXiv:2409.03733 [cs]},
}

@misc{krakovna_avoiding_2020,
	title = {Avoiding {Side} {Effects} {By} {Considering} {Future} {Tasks}},
	url = {http://arxiv.org/abs/2010.07877},
	doi = {10.48550/arXiv.2010.07877},
	urldate = {2024-10-13},
	publisher = {arXiv},
	author = {Krakovna, Victoria and Orseau, Laurent and Ngo, Richard and Martic, Miljan and Legg, Shane},
	month = oct,
	year = {2020},
	note = {arXiv:2010.07877},
}

@misc{davis_networks_2024,
	title = {Networks of {Networks}: {Complexity} {Class} {Principles} {Applied} to {Compound} {AI} {Systems} {Design}},
	shorttitle = {Networks of {Networks}},
	url = {http://arxiv.org/abs/2407.16831},
	urldate = {2024-10-12},
	publisher = {arXiv},
	author = {Davis, Jared Quincy and Hanin, Boris and Chen, Lingjiao and Bailis, Peter and Stoica, Ion and Zaharia, Matei},
	month = jul,
	year = {2024},
	note = {arXiv:2407.16831 [cs]},
}

@misc{hosseini_not_2024,
	title = {Not {All} {LLM} {Reasoners} {Are} {Created} {Equal}},
	url = {http://arxiv.org/abs/2410.01748},
	urldate = {2024-10-12},
	publisher = {arXiv},
	author = {Hosseini, Arian and Sordoni, Alessandro and Toyama, Daniel and Courville, Aaron and Agarwal, Rishabh},
	month = oct,
	year = {2024},
	note = {arXiv:2410.01748 [cs]},
}

@misc{hosseini_v-star_2024,
	title = {V-{STaR}: {Training} {Verifiers} for {Self}-{Taught} {Reasoners}},
	shorttitle = {V-{STaR}},
	url = {http://arxiv.org/abs/2402.06457},
	urldate = {2024-10-12},
	publisher = {arXiv},
	author = {Hosseini, Arian and Yuan, Xingdi and Malkin, Nikolay and Courville, Aaron and Sordoni, Alessandro and Agarwal, Rishabh},
	month = aug,
	year = {2024},
	note = {arXiv:2402.06457 [cs]},
}
